\documentclass[letterpaper]{article} % DO NOT CHANGE THIS
\usepackage{aaai2026}  % DO NOT CHANGE THIS
\usepackage{times}  % DO NOT CHANGE THIS
\usepackage{helvet}  % DO NOT CHANGE THIS
\usepackage{courier}  % DO NOT CHANGE THIS
\usepackage[hyphens]{url}  % DO NOT CHANGE THIS
\usepackage{graphicx} % DO NOT CHANGE THIS
\usepackage{natbib}  % DO NOT CHANGE THIS AND DO NOT ADD ANY OPTIONS TO IT
\usepackage{caption} % DO NOT CHANGE THIS AND DO NOT ADD ANY OPTIONS TO IT
\usepackage{algorithm}
\usepackage{algorithmic}

\usepackage{newfloat}
\usepackage{listings}
\DeclareCaptionStyle{ruled}{labelfont=normalfont,labelsep=colon,strut=off} % DO NOT CHANGE THIS
\floatstyle{ruled}
\newfloat{listing}{tb}{lst}{}
\floatname{listing}{Listing}
\usepackage{amssymb}
\usepackage{amsmath}
\usepackage{multirow}
\usepackage{bbding}
\usepackage{booktabs}
\title{Frequency-Aware Vision-Language Multimodality Generalization Network for Remote Sensing Image Classification}
\author{
    Junjie Zhang\textsuperscript{\rm 1}\thanks{A brief version of this work has been accepted by AAAI 2026. This extended version is considered as a more detailed description.}, 
    Feng Zhao\textsuperscript{\rm 1}\thanks{Corresponding author.}, 
    Hanqiang Liu\textsuperscript{\rm 2}, 
    Jun Yu\textsuperscript{\rm 3} 
}
\affiliations{
    \textsuperscript{\rm 1}School of Communications and Information Engineering, Xi’an University of Posts and Telecommunications\\
    \textsuperscript{\rm 2}School of Computer Science, Shaanxi Normal University\\
    \textsuperscript{\rm 3}Department of Automation, University of Science and Technology of China\\
    junjiezhang@stu.xupt.edu.cn, zhaofeng201@xupt.edu.cn, liuhq@snnu.edu.cn, harryjun@ustc.edu.cn
}

\begin{document}

\maketitle

\begin{abstract}
The booming remote sensing (RS) technology is giving rise to a novel multimodality generalization task, which requires the model to overcome data heterogeneity while possessing powerful cross-scene generalization ability. Moreover, most vision-language models (VLMs) usually describe surface materials in RS images using universal texts, lacking proprietary linguistic prior knowledge specific to different RS vision modalities. In this work, we formalize RS multimodality generalization (RSMG) as a learning paradigm, and propose a frequency-aware vision-language multimodality generalization network (FVMGN) for RS image classification. Specifically, a diffusion-based training-test-time augmentation (DTAug) strategy is designed to reconstruct multimodal land-cover distributions, enriching input information for FVMGN. Following that, to overcome multimodal heterogeneity, a multimodal wavelet disentanglement (MWDis) module is developed to learn cross-domain invariant features by resampling low and high frequency components in the frequency domain. Considering the characteristics of RS vision modalities, shared and proprietary class texts is designed as linguistic inputs for the transformer-based text encoder to extract diverse text features. For multimodal vision inputs, a spatial-frequency-aware image encoder (SFIE) is constructed to realize local-global feature reconstruction and representation. Finally, a multiscale spatial-frequency feature alignment (MSFFA) module is suggested to construct a unified semantic space, ensuring refined multiscale alignment of different text and vision features in spatial and frequency domains. Extensive experiments show that FVMGN has the excellent multimodality generalization ability compared with state-of-the-art (SOTA) methods.
\end{abstract}

% Uncomment the following to link to your code, datasets, an extended version or similar.
% You must keep this block between (not within) the abstract and the main body of the paper.
\begin{links}
    \link{Code}{https://github.com/ZJier/FVMGN}
%    \link{Datasets}{https://aaai.org/example/datasets}
%    \link{Extended version}{https://aaai.org/example/extended-version}
\end{links}

\section{Introduction}
Remote sensing image classification (RSIC) is an important means for perceiving geomorphic features and landscape architectures~\cite{RSIC1}. Deep learning, as a catalyst for the rapid development of RSIC, has elevated the efficiency and accuracy to a new degree for land-cover classification task, promoting innovative research and practical applications within the RS community~\cite{Appl1,Appl2,Appl3,Appl4}. 

Recently, with the development of sensor and satellite technologies, types of RS images are becoming increasingly diverse, such as hyperspectral (HS) image, light detection and ranging (LiDAR) image, and synthetic aperture radar (SAR) image, which usually have distinct data characteristics ~\cite{Review1}. 
%In view of this, researchers have developed classification methods for different RS modalities. 
In view of this, researchers have developed numerous advanced \textbf{single-modal RSIC} methods for different modal RS data. For HS images, existing methods usually extract feature information from the perspective of jointly spectral and spatial modeling, so as to more clearly describe the land-cover distribution~\cite{MHSI0}. For SAR images, most networks distinguish the objects with structure and texture differences by learning the unique scattering property and polarization information~\cite{SAR1}. For LiDAR images, most algorithms capture the elevation information contained in images to reflect the height differences among different objects~\cite{LiDAR1}.

Real-world tasks usually require models to adapt to unseen RS scenes, while single-modal RSIC methods face limitations in domain generalization (DG)~\cite{MMRS1,MMRS2}. Based on this, \textbf{single-modal RSDG} methods have sprouted up everywhere. It refers to training a model using RS data from a scene, hoping that the model achieves good generalization performance on new and unseen scenes. This task requires the model to learn robust and domain-invariant features that can adapt to data distribution characteristics from the target domain~\cite{SSDG1,SSDG2}. Common single-modal RSDG methods include meta-learning, adversarial training, feature disentanglement~\cite{Feadis1}, and others. In addition, modality types in a RS scene are becoming increasingly diverse, which reflects geomorphic structures comprehensively while also posing challenges to single-modal RSIC methods. Given this, \textbf{multimodal RSIC} methods have developed rapidly. It refers to performing classification tasks using RS data obtained from different types of sensors, namely various modality combinations, such as HS, LiDAR, and others. This task requires that the model understand the complementarity and similarity between different modalities, realizing multimodal interaction while overcoming the data heterogeneity~\cite{MRSIC2}. Common multimodal RSIC methods include feature-level fusion classification, decision-level fusion classification, and others. 

However, multimodal RSIC-based methods are usually required to have the capability to effectively adapt to unseen new scenes in practical applications, where heterogeneity among modalities and distribution differences cross RS scenes are urgent problems that need to be addressed. Therefore, a novel and interesting task, \textbf{RS multimodality generalization (RSMG)} for image classification, comes into being, which requires the designed algorithm to consider multimodality heterogeneity and generalization simultaneously. RSMG requires that a well-trained model on the source domain can show excellent generalization ability on the target domain under the premise where multimodal training and test data are mutually unseen. RSMG aims to more effectively reflect the geographical environment and urbanization process in different regions, which can provide crucial support for multi-perspective and cross-regional disaster monitoring, resource management and land planning. 

In this paper, we propose a frequency-aware vision-language multimodality generalization network (FVMGN) for RSIC, which can achieve the joint representation for land-cover distribution in spatial and frequency domains. In the pre-processing phase, we design a diffusion-based training-test-time augmentation (DTAug) strategy to achieve multimodal land-cover distribution reconstructions, thereby enhancing the diversity of input information. In the feature interaction phase, we develop a multimodal wavelet disentanglement (MWDis) module, which can learn cross-domain invariant features while achieving multimodal feature interaction. In the feature extraction phase, we  design a spatial-frequency-aware image encoder (SFIE) to achieve spatial and multi-frequency analysis by organically integrating wavelet transform into CNN and ViT. In the feature alignment phase, we develop a multiscale spatial-frequency feature alignment (MSFFA) module to realize multi-level supervision between text and vision features, thereby further improving multimodality generalization ability. To summarize, contributions of this work are as follows.

\begin{itemize}
	\item We formalize RSMG as a learning paradigm, and propose a frequency-aware vision-language multimodality generalization network (FVMGN).
	\item Shared and proprietary class texts are designed based on large language models, which can reflect general characteristics and modality-specific geospatial attributes.
	\item A MWDis module is designed to realize the multimodal information interaction, helping the network learn domain-invariant generalization features.
	\item A SFIE is developed by integrating wavelet transform, convolution operation, and self-attention mechanism to realize local-global spatial-frequency feature extraction.
	\item A MSFFA module is proposed to achieve fine multiscale vision-vision and vision-text feature alignments in spatial and frequency domains.
\end{itemize}

\section{Related Works}
\subsection{Vision-Language Model}
Vision-language model (VLM) aims to understand the complex relationships between images and text, enabling more accurate decision-making and reasoning~\cite{CLIP,VLMs}. Currently, diverse VLM-based methods have emerged in various fields, such as visual question answering~\cite{VQA}, image captioning~\cite{IMGCAP}, text-to-image retrieval~\cite{TTIR}, and more. In view of this, researchers have explored how VLMs perform on the RSIC task~\cite{EHSNet}. \citet{LDGNet} described coarse-grained and fine-grained texts as linguistic prior knowledge based on class names for HS image classification, serving as effective text-based soft connections for different targets. \citet{TMCFN} refined the class name descriptions based on intrinsic attributes and interclass relationships, and utilized the attention mechanism to fuse text and vision features, thereby effectively achieving joint HS and LiDAR data classification.

\subsection{Denoising Diffusion Probabilistic Model}
Denoising diffusion probabilistic model (DDPM) is a generative model that reconstructs data distribution by adding noise in the forward process and removing noise in the reverse process~\cite{DDPM}. DDPM has been widely applied in various fields, such as image generation~\cite{DDPM1}, image super-resolution~\cite{DDPM2}, and image restoration~\cite{DDPM3}. In recent years, DDPM with powerful data generation capability has shone in RSIC field~\cite{DDPM4,MHSI1}. \citet{DDPM4} reconstructed the spectral-spatial land-cover distribution through the noise addition and denoising processes in DDPM, confirming its effectiveness in HS image classification tasks. \citet{MHSI1} considered the land-cover distribution reconstructed by DDPM as an unsupervised generative knowledge, which is combined with HS data to achieve competitive classification results.

\begin{figure}[t]
	\begin{center}
		\includegraphics[width=\linewidth]{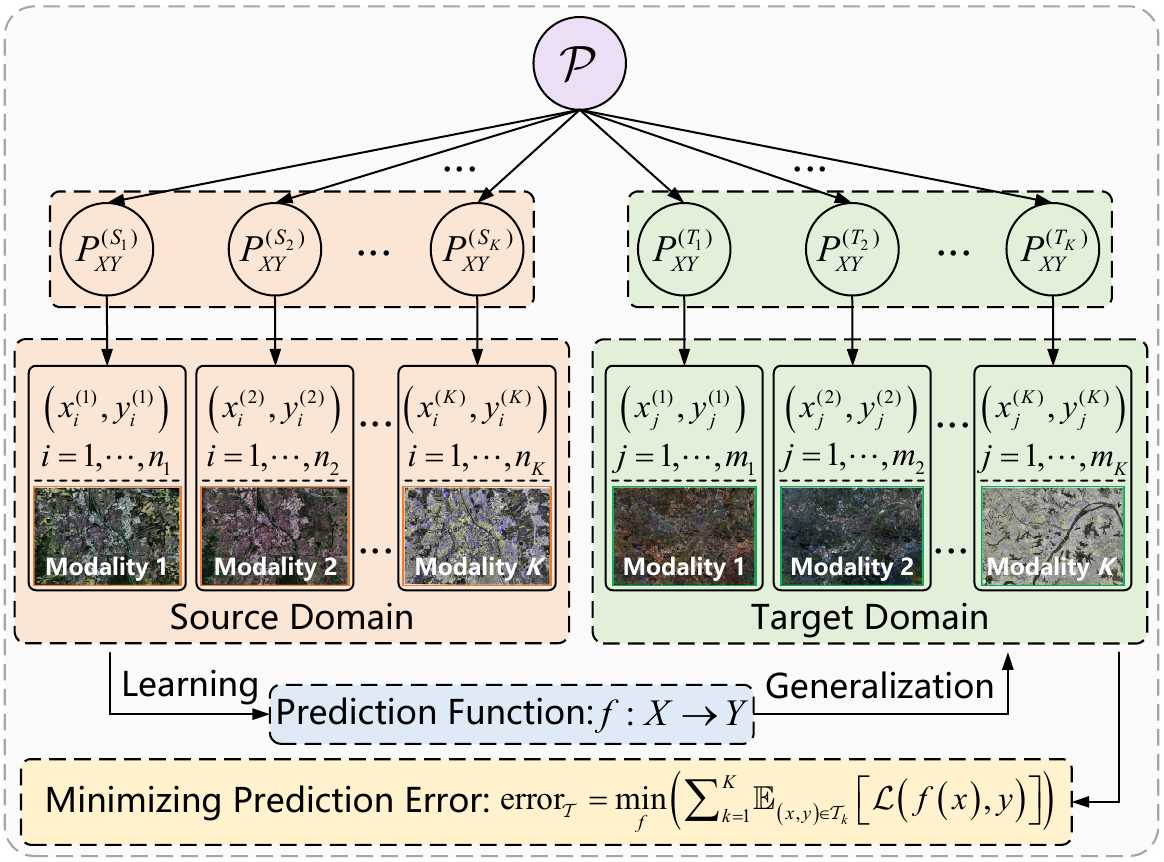}
	\end{center}
	\caption{Problem definition for RSMG.}
	\label{Problem}
\end{figure}

\begin{figure*}[t]
	\begin{center}\vspace{-2mm}
		\includegraphics[width=\linewidth]{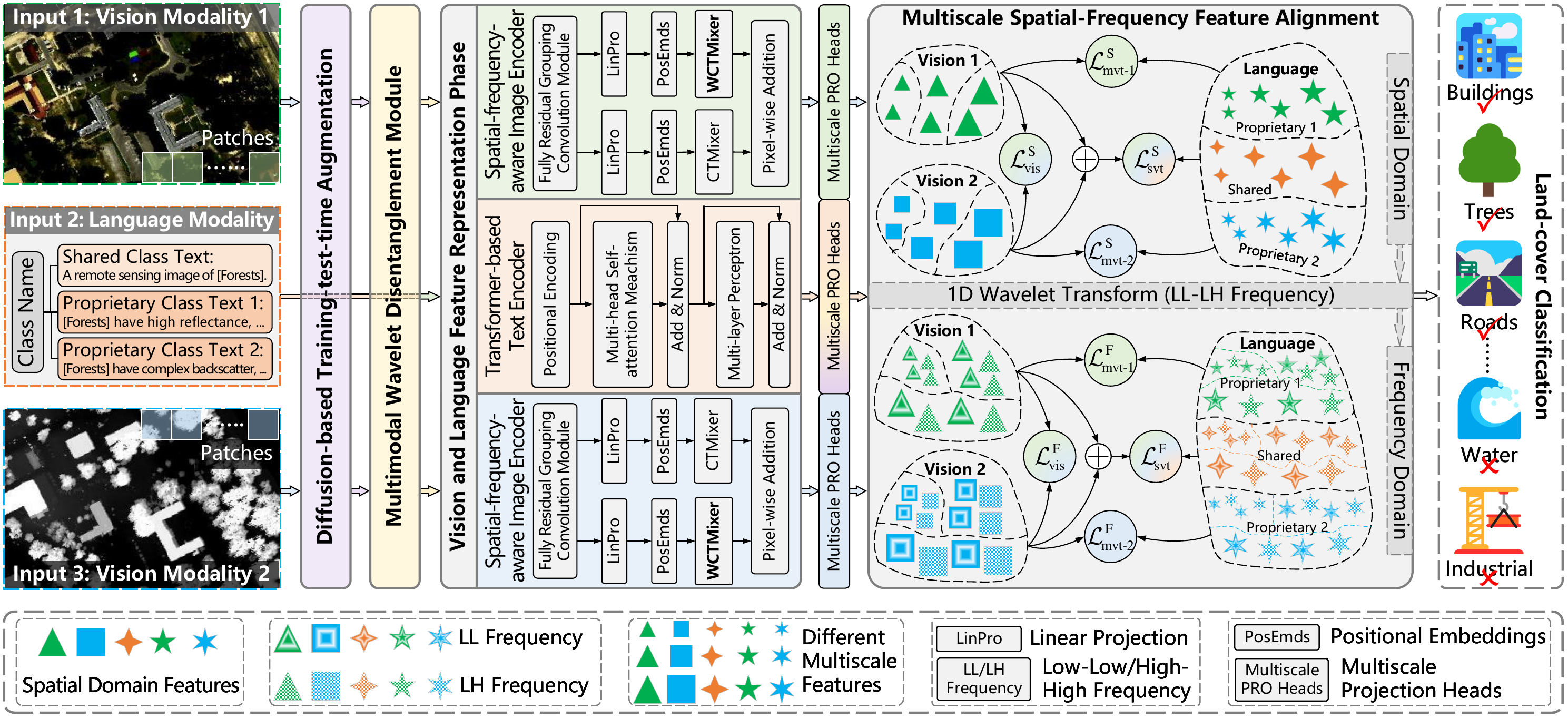}
	\end{center}\vspace{-2mm}
	\caption{Overview of the proposed FVMGN. It consists of five components: DTAug, MWDis, SFIE, TTE, and MSFFA.}
	\label{FVMGN}\vspace{-4mm}
\end{figure*}

\subsection{Wavelet Transform}
Recently, the synthesis of the wavelet transform and neural network has made a remarkable success in the computer vision field~\cite{WTConv3}. Wavelet transform can decompose image data in the spatial domain into multiple frequency signals in the frequency domain, realizing multi-frequency analysis and feature reconstruction for image information~\cite{WTConv1,WTConv4}. \citet{WTConv} proposed wavelet convolution by combining wavelet transform and convolution operation, which proved to have a larger receptive field than vanilla convolution.  Inspired by the above works, researchers have begun to explore the application value of the wavelet transform on RSIC tasks. \citet{WTConv2} applied wavelet transform for reversible down-sampling to achieve lossless compression for feature maps while enhancing structure and shape information in spatial and channel dimensions. \citet{WTConv1} treated the wavelet function as a learnable activation function to achieve nonlinear mapping, which allows the Kolmogorov-Arnold network to mine multiscale spatial-spectral patterns.

\section{Methodology}
\textbf{Problem definition}. As shown in Fig. \ref{Problem}. Let $\mathcal{X}$ be the input space, $\mathcal{Y}$ be the output (label) space, and domain is represented by the joint distribution $P_{XY}$ over the input samples $X$ and labels $Y$ on $\mathcal{X}\times \mathcal{Y}$. Given $K$ RS modalities 
$\mathcal{S}=\{S_{k}=\{(x^{(k)}_{n_{k}},y^{(k)}_{n_{k}})\}\}^{K}_{k=1}\sim P^{(S_k)}_{XY}$, where $n_{k}$ denotes the number of samples in the $k$-th modality $S_{k}$, and $P^{(S_k)}_{XY}\neq P^{(S_{k'})}_{XY}, k\neq k'; k,k'\in\{1,\dots,K\}$. RSMG aims to learn a prediction model $f: \mathcal{X}\rightarrow\mathcal{Y}$ using only the multimodal data from the source domain, such that the model $f$ minimizes the prediction error $\rm{error}_{\mathcal{T}}$ on $K$ unseen target modalities $\mathcal{T}=\{T_{k}=\{(x^{(k)}_{m_{k}},y^{(k)}_{m_{k}})\}\}^{K}_{k=1}\sim P^{(S_k)}_{XY},\forall k\in\{1,\dots,K\}$. Here, $m_{k}$ denotes the number of samples in the $k$-th target modality $T_{k}$, and $P^{(T_{k})}_{XY}\neq P^{(T_{k'})}_{XY}\neq P^{(S_{k})}_{XY}\neq P^{(S_{k'})}_{XY}, k\neq k'; k,k'\in\{1,\dots,K\}$. This means that prediction model $f$ trained on the source domain can generalize well even without access to the target domain during the training phase. The $\rm{error}_{\mathcal{T}}$ is defined as: 
\begin{equation}{\label{rsmgfunc}}
	{\rm{error}}_{\mathcal{T}}=\min_{\underset{}{f}}\left(\sum_{k=1}^{K}\mathbb{E}_{\left(x,y\right)\in T_{k}}\left[\mathcal{L}\left(f\left(x\right),y\right)\right]\right),
\end{equation}
where $\mathbb{E}$ denotes the expectation, and $\mathcal{L}$ is the loss function.

\noindent{\textbf{Overview}}. As shown in Fig. \ref{FVMGN}, FVMGN is mainly composed of the DTAug strategy, MWDis module, SFIE, transformer-based text encoder (TTE), and MSFFA module. These sub-parts form a powerful whole, each with different tasks. 1) DTAug utilizes the DDPM to reconstruct multimodal land-cover distributions, enriching model inputs. 2) MWDis performs feature disentanglement in the frequency domain, enhancing the ability to learn cross-domain-invariant representation. 3) SFIE organically integrates wavelet transform, convolution operations, and attention mechanism, which can realize multi-frequency analysis in the frequency domain while extracting local-global features in the spatial domain. 4) MSFFA performs multiscale vision-vision feature alignments in wavelet and spatial domains, thereby achieving more refined positive sample pair matching. Finally, the classification result can be obtained by the maximum score strategy with two linear classifiers.
\subsection{DTAug}
Motivated by DDPM~\cite{DDPM}, we explore its effectiveness in data augmentation, and design a novel multimodal RS data augmentation strategy, called DTAug, which enriches the input information of FVMGN by reconstructing multimodal land-cover distribution, thereby enhancing the cross-scene generalization ability (refer to \textit{Supp. Fig.} 1). Firstly, we feed the RS data from the two modalities into the DDPM based on 3D UNet~\cite{MHSI1} to generate the reconstructed and unsupervised land-cover distributions. Secondly, we perform principal component analysis (PCA: $30$) on the original and diffusion-enhanced RS data, and concatenate the reduced data on the channel dimension, thereby obtaining obtain new multimodal input data. In addition, the obtained RS data is cropped into patches with the same size, and geometry augmentation based on patches is performed to further increase the diversity, such as flip augmentation and radiation augmentation. Notably, the entire land-cover distribution generation process is unsupervised in DTAug, and the multimodal RS data from source and target domains are mutually unseen in training and test phase.
\subsection{MWDis}
\begin{figure}[t]
	\begin{center}
		\includegraphics[width=0.99\linewidth]{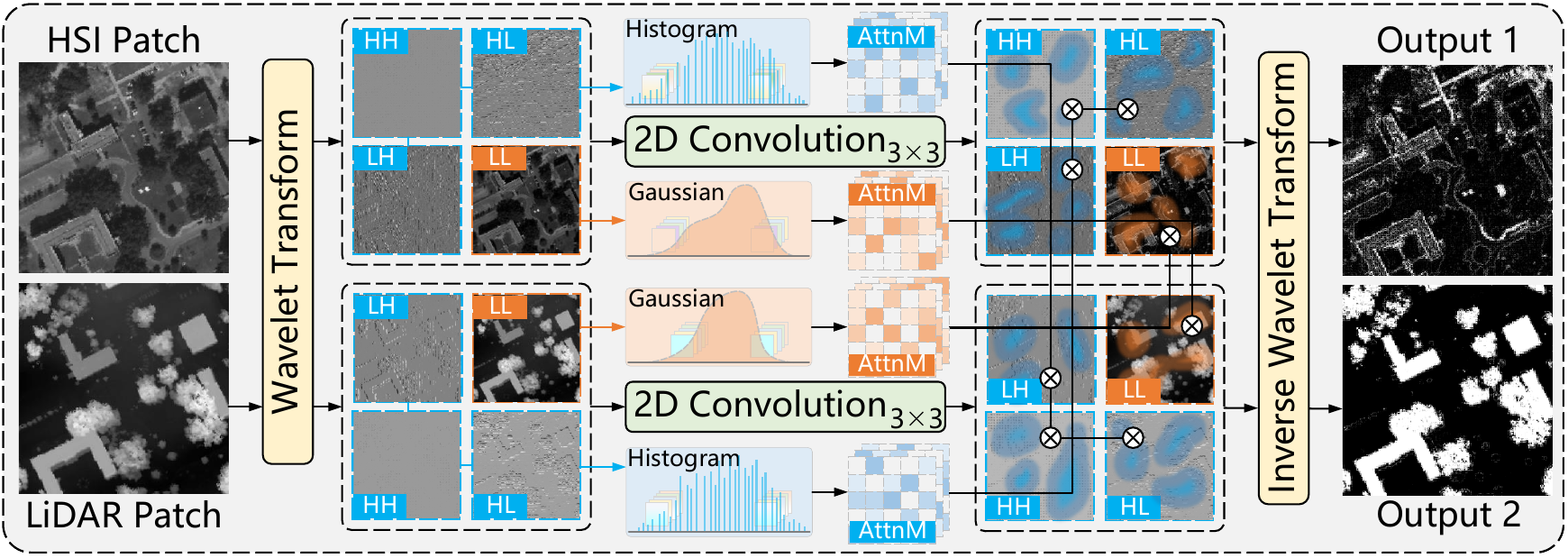}
	\end{center}\vspace{-2mm}
	\caption{Structure of MWDis module.}
	\label{MWDis}
\end{figure}

For various types of RS data, the interaction and fusion between modalities are key to achieve accurate land-cover classification. As shown in Fig. \ref{MWDis}, we utilize wavelet decomposition to achieve image disentanglement for multimodal data, which allows for learning cross-domain invariant features while overcoming modality heterogeneity. Taking HS and LiDAR modalities as examples, they are decomposed into a low-frequency component $\mathbf{F}^{ll}$ (low-low (LL) frequency) and a set of high-frequency components $\mathbf{F}^{h}$ (high-low/low-high/high-high (HL/LH/HH) frequencies). LL frequency usually contains higher energy, and resampling it can change the source sample style~\cite{LLFREQ}. Therefore, to enhance sample diversity, we model LL frequency as the multivariate Gaussian distribution, as follows: 
\begin{equation}{\label{gauss}}
	\begin{aligned}
		\mathbf{\hat{F}}^{ll}&=\mathbf{F}^{ll}+\rho\sigma\left[\mathbf{F}^{ll}\right], \\ 
		\sigma^{2}\left[\mathbf{F}^{ll}\right]&=\frac{1}{N}\sum_{n=1}^{N}\left[\mathbf{F}_{n}^{ll}-\mu\left[\mathbf{F}_{n}^{ll}\right]\right]^{2},
	\end{aligned}
\end{equation}
where $\{\rho\backsim\mathcal{N}(0,\alpha),\alpha\in[0,1]\}$ denotes the strength of domain shifts, $\mathbf{\hat{F}}^{ll}$ is the resampled LL frequency feature, and $N$ is batch size. $\mu[\cdot]$ and $\sigma[\cdot]$ represent the mean and standard deviation, respectively. In addition, the resamped LL frequency features $\mathbf{\hat{F}}^{ll}_{{\rm{v}}1}$ and $\mathbf{\hat{F}}^{ll}_{{\rm{v}}2}$ are calculated as spatial attention (SpatAttn)~\cite{CBAM} matrix, which is acted on the convolution (Conv) features of another modality. The subscripts ${\rm{v}}1$ and ${\rm{v}}2$ represent two vision modalities, respectively. This process can be expressed as follows: 
\begin{equation}{\label{spatattn}}
	\mathbf{F}^{ll}_{{\rm{I}}1}={\rm{SpatAttn}}(\mathbf{\hat{F}}^{ll}_{{\rm{v}}2})\otimes{\rm{Conv}}\left(\mathbf{F}^{ll}_{{\rm{v}}1}\right),
\end{equation}
where $\mathbf{F}^{ll}_{{\rm{I}}1}$ is the weighted low-frequency feature of vision modality 1, and $\otimes$ denotes element-wise multiplication operation. Similarly, we can obtain the weighted LL frequency feature $\mathbf{F}^{ll}_{{\rm{I}}2}$ of vision modality 2.

HL/LH/HH frequencies are usually sharp semantics, with the most noticeable intensity changes occurring at the boundaries between adjacent ground objects~\cite{Gradient}. Given this, we perform gradient-map-based resampling on high-frequency components, which can better capture relative intensity variations as the modality generalization representation. Gradient-map resampling $G(\cdot)$ is formulated by: 
\begin{equation}{\label{gradientmap}}
	G\left(\mathbf{F}^{h}\right)=HE\left(
	\sqrt{\left(\mathbf{F}^{hl}\right)^{2}+\left(\mathbf{F}^{lh}\right)^{2}+\left(\mathbf{F}^{hh}\right)^{2}}
	\right),
\end{equation}
where $HE(\cdot)$ represents the histogram equalization. Similarly, we can obtain the weighted features of HL/LH/HH frequencies through Eq.(\ref{spatattn}). After the frequency interaction mentioned above, we perform an inverse wavelet transform on the resampling frequency components to achieve the effective feature reconstruction.

\begin{figure}[t]
	\begin{center}
		\includegraphics[width=0.99\linewidth]{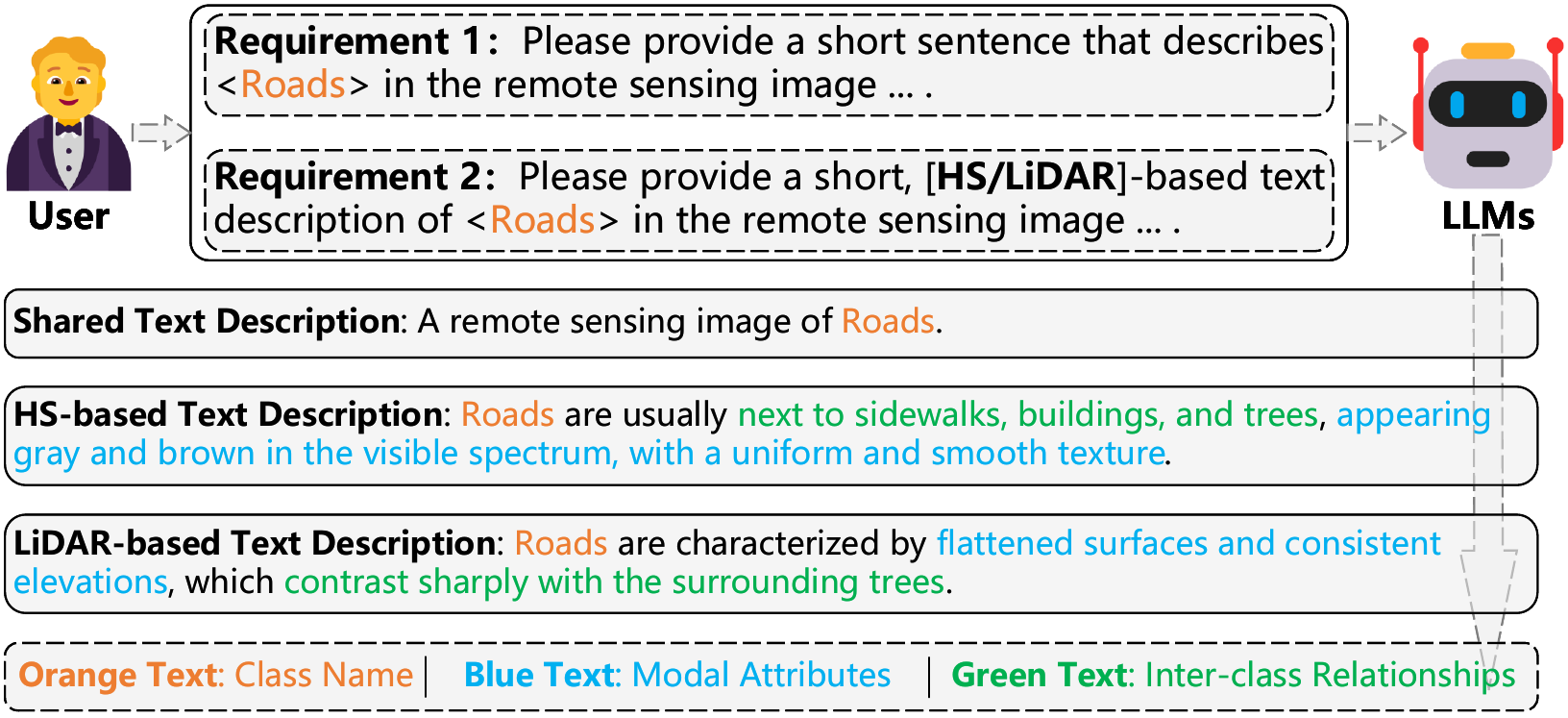}
	\end{center}\vspace{-2mm}
	\caption{Different types of class text descriptions (Class: [\textit{Roads}] and [\textit{Forests}] as examples).}
	\label{Texts}
\end{figure}

\subsection{Vision \& Language Feature Representation}
The vision and language feature representation phase contains a TTE and two SFIEs, as shown in Fig. \ref{FVMGN}.

\noindent{\textbf{TTE}}. As shown in Fig. \ref{Texts}, considering the uniqueness and commonality of multimodal RS data, we design the shared and modality-specific text descriptions for different class names through the large language model and manual fine-tuning. The shared text description provides general linguistic information. Modality-specific text descriptions provide linguistic prior knowledge for vision information representation in terms of modal attributes and inter-class relationships, such as Class [\textit{Roads}]: 1) Inter-class relationships [\textit{Roads are usually next to sidewalks, buildings, and trees}] and 2) HS modal attributes [\textit{Roads appear gray and brown in the visible spectrum, with a uniform and smooth texture}]. As shown in Fig. \ref{FVMGN}, TTE is a language model-based transformer without pre-training~\cite{Trans,Trans1}. Similar to CLIP, TTE utilizes a lower-cased byte pair encoding (BPE) for feature representation of the different text descriptions, where the vocabulary size and maximum sequence length are 49,152 and 76~\cite{BPE}, respectively.

\noindent{\textbf{SFIE}}. To achieve full-domain spatial-frequency feature extraction, we design the SFIE with a fully residual grouping convolution module (FRGCM) and a spatial-frequency dual-branch backbone. For FRGCM, we apply $3\times3$ Conv layers between two $1\times1$ Conv (channel transformation) layers to facilitate inter-group feature interaction and fusion, enhancing feature diversity (refer to \textit{Supp. Fig.} 2). 

Spatial-frequency dual-branch backbone mainly consists of a wavelet convolution transformer mixer (WCTMixer) and a convolution transformer mixer (CTMixer), where CTMixer is applied to extract local-global spatial feature according to~\cite{MVAHN}, as shown in Fig. \ref{FVMGN}. For WCTMixer, we design a multi-head wavelet self-attention (MHWSA) mechanism using wavelet transforms. Specifically, the input feature is decomposed into different frequency components using wavelet transform, which are considered as different parallel heads according to their properties for self-attention mechanism. Here, we take LL frequency as an example to illustrate the above process: 
\begin{equation}{\label{wsa}}
	\begin{aligned}
		&\mathbf{Z}^{ll}_{\rm{out}}={\rm{CPool}}\left({\rm{WSA}}\left(\mathbf{Q}^{ll},\mathbf{K}^{ll},\mathbf{V}^{ll},\mathbf{C}^{ll}\right)\right), \\
		&\mathbf{Q}^{ll},\mathbf{K}^{ll},\mathbf{V}^{ll},\mathbf{C}^{ll}={\rm{Re}}\left({\rm{Chunk}}\left({\rm{Conv}}\left(\mathbf{Z}^{ll}\right)\right)\right), \\
		&\mathbf{Z}^{ll},\mathbf{Z}^{hl},\mathbf{Z}^{lh},\mathbf{Z}^{hh}={\rm{WT}}\left(\mathbf{Z}_{\rm{in}}\right),
	\end{aligned}
\end{equation}
where $\mathbf{Z}_{\rm{in}}$ represents the input of the MHWSA mechanism. $\mathbf{Z}^{ll}$, $\mathbf{Z}^{hl}$, $\mathbf{Z}^{lh}$ and $\mathbf{Z}^{hh}$ are different frequency components. ${\rm{WT}}(\cdot)$, ${\rm{Re}}(\cdot)$, ${\rm{Chunk}}(\cdot)$, ${\rm{CPool}}(\cdot)$ and ${\rm{WSA}}(\cdot)$ represent wavelet transform, reshape, chunking, channel pooling, and wavelet self-attention. $\mathbf{Q}^{ll}$, $\mathbf{K}^{ll}$, $\mathbf{V}^{ll}$, and $\mathbf{C}^{ll}$ are intermediate features. $\mathbf{Z}^{ll}_{\rm{out}}$ is the output of the channel pooling layer. Similarly, we apply Eq.(\ref{wsa}) to obtain the output features ($\mathbf{Z}^{hl}_{\rm{out}}$, $\mathbf{Z}^{lh}_{\rm{out}}$, $\mathbf{Z}^{hh}_{\rm{out}}$) for the other frequency components. Finally, output feature $\mathbf{Z}_{\rm{out}}$ of the MHWSA mechanism is acquired through the inverse WT (IWT), as follows: 
\begin{equation}{\label{wsa1}}
	\mathbf{Z}_{\rm{out}}={\rm{IWT}}\left(\mathbf{Z}^{ll}_{\rm{out}},\mathbf{Z}^{hl}_{\rm{out}},\mathbf{Z}^{lh}_{\rm{out}},\mathbf{Z}^{hh}_{\rm{out}}\right),
\end{equation}

In summary, SFIE first utilizes the designed FRGCM to extract grouping-interactive feature information, and then applies WCTMixer and CTMixer to capture discriminative local-global vision features from the spatial and frequency domains, enhancing the model representation capability.

\begin{figure}[t]
	\begin{center}
		\includegraphics[width=0.98\linewidth]{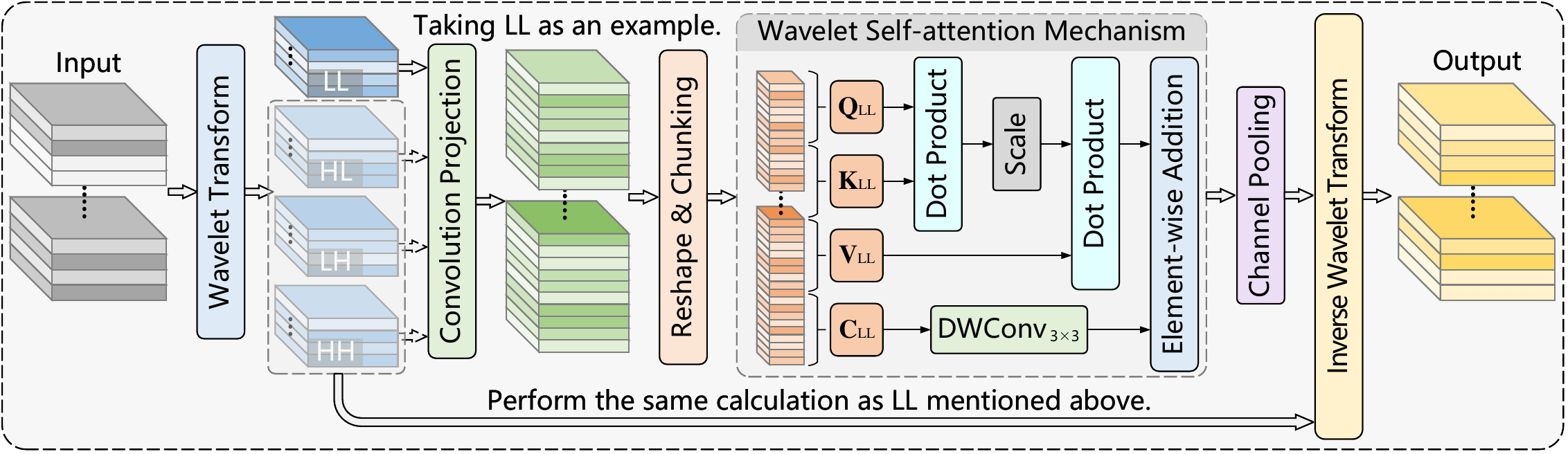}
	\end{center}
	\caption{Structure of MHWSA mechanism.}
	\label{WCTMixer}
\end{figure}

\subsection{MSFFA}
In VLMs, the contrastive learning-based vision-text feature alignment allows matching image-text pairs to be closer in the same embedding space, thereby understanding the relationships between image and text~\cite{CLIP}. 

\begin{figure}[t]
	\begin{center}
		\includegraphics[width=0.98\linewidth]{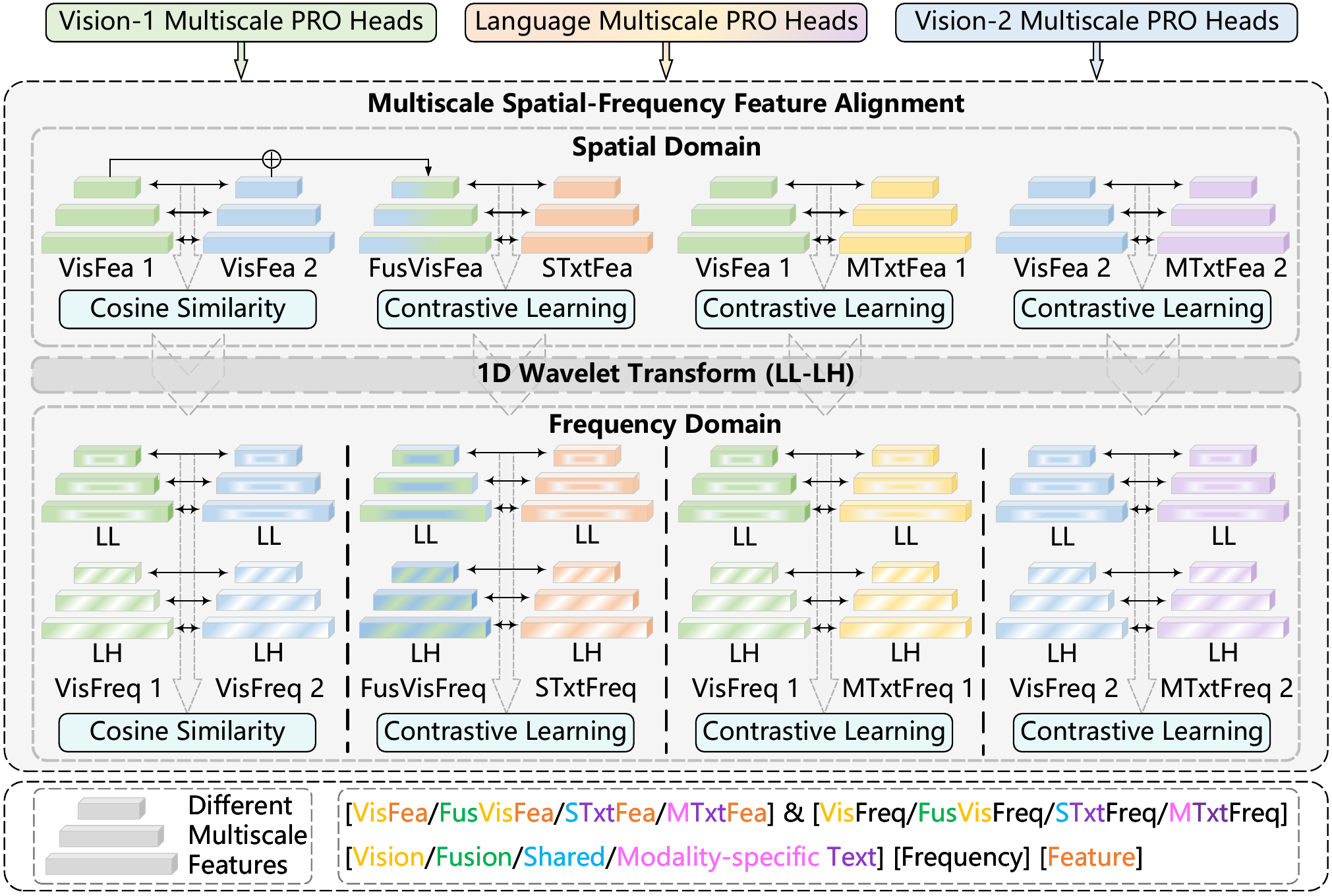}
	\end{center}
	\caption{Structure of MSFFA module.}
	\label{MSFFA}
\end{figure}

As shown in Fig. \ref{MSFFA}, compared with traditional RS feature alignments~\cite{LDGNet,EHSNet,TMCFN}, we consider the following three aspects in MSFFA. \textbf{1) Multiscale characteristics}: MSFFA aligns the multiscale vision and text features obtained from the multiscale projection heads in a unified semantic space, which enables a more refined understanding for vision-text pairs, thereby enhancing overall generalization on RSMG tasks. \textbf{2) Joint spatial-frequency domain}: MSFFA performs multiscale feature alignment in the spatial domain while utilizing wavelet transform to decouple spatial features into the LL and LH frequencies, thereby achieving multiscale and multi-frequency feature alignment in the frequency domain. \textbf{3) Multimodal vision-vision feature alignment}: Most vision-language models focus on feature alignment for vision-text pairs, while neglecting the complementarity betweeen vision features, especially for RS multimodal tasks. In view of this, MSFFA applies cosine similarity to design vision-vision feature alignment loss, enhancing multimodal representation capability. In addition, the vision-text feature alignment losses at various scales and domains refer to contrastive learning~\cite{CLIP} in implementation, and the final loss contains multiscale spatial-wavelet vision-vision and vision-text alignment losses.

\section{Experiments}
\subsection{Experimental Setup}\label{datasetdes}
\noindent{\textbf{Datasets}}. 
We select three publicly available multimodality datasets for the related experiments, namely the HS-LiDAR MUUFL~\cite{MUUFL}, Trento~\cite{Trento}, and Houston 2013 (HU2013)~\cite{HU2013} datasets. For the above three datasets, we select three common classes for generalization analysis, namely, [\textit{Trees}], [\textit{Roads}], and [\textit{Buildings}], which show significant differences in land-cover distribution across different datasets. In addition, to accommodate various types of real-world tasks, we perform mutual multimodality generalization among three datasets, meaning each dataset can serve as the source domain (SD) and the target domain (TD), namely SD$\leftrightarrow$TD: MUUFL$\leftrightarrow$Trento$\leftrightarrow$HU2013 (MU$\leftrightarrow$TR$\leftrightarrow$HU). Dataset details are provided in \textit{Supp. Table} 1 \& \textit{Fig}. 3.

\begin{table*}[t]\centering
	\caption{Classification results of different methods on several multimodality generalization dataset combinations. \label{Coms}}\vspace{-2mm}
	\resizebox{0.99\linewidth}{26.5mm}{
		\begin{tabular}{ccccccccccccccc}
			\toprule
			SD$\rightarrow$TD & Metric (\%) & MFT & MsFE-IFN & CMFAEN & DKDMN & SDENet & LLURNet & FDGNet & TFTNet & ADNet & ISDGS & EHSNet & LDGNet & FVMGN \\ \midrule
			& OA & $77.59_{\pm1.01}$ & $\underline{83.98}_{\pm1.86}$ & $82.82_{\pm1.06}$ & $78.95_{\pm2.38}$ & $78.14_{\pm1.40}$ & $78.23_{\pm1.42}$ & $80.11_{\pm0.26}$ & $76.32_{\pm1.67}$ & $80.36_{\pm0.16}$ & $80.27_\pm0.67$ & $77.58_{\pm4.17}$ & $83.17_{\pm1.63}$ & $\mathbf{92.44}_{\pm1.24}$ \\
			& AA & $61.46_{\pm1.64}$ & $\underline{73.60}_{\pm3.61}$ & $70.75_{\pm1.46}$ & $64.48_{\pm3.77}$ & $63.75_{\pm2.17}$ & $63.17_{\pm2.20}$ & $65.62_{\pm0.37}$ & $60.44_{\pm2.72}$ & $66.03_{\pm0.19}$ & $66.51_{\pm1.23}$ & $63.09_{\pm6.28}$ & $72.36_{\pm3.41}$ & $\mathbf{88.58}_{\pm2.13}$ \\
			\multirow{-3}{*}{MU$\rightarrow$TR} & Kappa & $53.54_{\pm2.70}$ & $69.25_{\pm4.30}$ & $66.81_{\pm2.02}$ & $58.08_{\pm4.82}$ & $60.22_{\pm2.84}$ & $60.49_{\pm2.94}$ & $\underline{64.14}_{\pm0.51}$ & $56.15_{\pm3.77}$ & $64.67_{\pm0.28}$ & $64.59_{\pm1.18}$ & $58.49_{\pm8.52}$ & $\underline{69.65}_{\pm3.15}$ & $\mathbf{86.59}_{\pm2.16}$ \\ \midrule
			& OA & $27.64_{\pm4.44}$ & $16.93_{\pm1.98}$ & $23.17_{\pm4.11}$ & $13.38_{\pm3.03}$ & $61.60_{\pm1.76}$ & $60.11_{\pm1.62}$ & $56.20_{\pm1.15}$ & $67.96_{\pm2.07}$ & $\underline{70.39}_{\pm2.71}$ & $65.24_{\pm1.63}$ $\pm$ & $41.06_{\pm9.83}$ & $49.68_{\pm2.10}$ & $\mathbf{93.19}_{\pm2.08}$ \\
			& AA & $27.63_{\pm3.03}$ & $15.28_{\pm2.57}$ & $19.51_{\pm3.32}$ & $13.43_{\pm3.19}$ & $65.58_{\pm2.12}$  & $64.93_{\pm1.88}$ & $61.41_{\pm1.57}$ & $70.98_{\pm2.68}$ & $\underline{75.13}_{\pm2.47}$ & $71.12_{\pm1.54}$ & $41.85_{\pm7.83}$ & $51.09_{\pm2.09}$ & $\mathbf{93.28}_{\pm1.99}$ \\
			\multirow{-3}{*}{MU$\rightarrow$HU} & Kappa & $9.930_{\pm5.12}$ & $26.73_{\pm4.70}$ & $21.04_{\pm5.45}$ & $30.22_{\pm4.27}$ & $43.83_{\pm2.75}$ & $42.16_{\pm2.21}$ & $36.81_{\pm1.85}$ & $52.73_{\pm3.13}$ & $\underline{56.93}_{\pm3.78}$ & $49.86_{\pm2.30}$ & $14.16_{\pm12.5}$ & $24.76_{\pm2.97}$ & $\mathbf{89.64}_{\pm3.15}$ \\ \midrule
			& OA & $23.32_{\pm0.86}$ & $20.66_{\pm4.94}$ & $26.75_{\pm6.53}$ & $29.05_{\pm2.64}$ & $69.11_{\pm6.76}$ & $66.68_{\pm5.34}$ & $57.80_{\pm12.1}$ & $66.66_{\pm3.95}$ & $73.90_{\pm3.45}$ & $51.12_{\pm6.21}$ & $71.01_{\pm7.27}$ & $\underline{74.88}_{\pm4.94}$ & $\mathbf{82.87}_{\pm3.77}$ \\ 
			& AA & $38.00_{\pm1.51}$ & $35.13_{\pm7.72}$ & $45.34_{\pm10.3}$ & $48.15_{\pm4.69}$ & $60.96_{\pm3.72}$ & $59.19_{\pm3.38}$ & $55.28_{\pm7.05}$ & $60.43_{\pm2.26}$ & $64.02_{\pm2.12}$ & $51.45_{\pm3.34}$ & $60.24_{\pm8.56}$ & $\underline{63.01}_{\pm5.93}$ & $\mathbf{78.29}_{\pm5.23}$ \\
			\multirow{-3}{*}{HU$\rightarrow$TR} & Kappa & $1.530_{\pm1.53}$ & $4.120_{\pm8.09}$ & $6.300_{\pm9.89}$ & $10.10_{\pm3.58}$ & $49.01_{\pm8.64}$ & $45.42_{\pm7.05}$ & $36.06_{\pm13.9}$ & $45.81_{\pm4.97}$ & $55.49_{\pm4.88}$ & $28.18_{\pm6.35}$ & $47.78_{\pm13.9}$ & $\underline{55.55}_{\pm8.48}$ & $\mathbf{69.93}_{\pm6.40}$ \\ \midrule 
			& OA & $17.82_{\pm0.32}$ & $48.38_{\pm13.9}$ & $18.60_{\pm0.14}$ & $46.31_{\pm14.1}$ & $62.25_{\pm2.61}$ & $48.47_{\pm13.1}$ & $22.26_{\pm13.7}$ & $40.65_{\pm2.98}$ & $51.41_{\pm18.0}$ & $41.39_{\pm16.9}$ & $64.73_{\pm3.46}$ & $\underline{80.41}_{\pm3.02}$ & $\mathbf{82.33}_{\pm3.75}$ \\ 
			& AA & $31.95_{\pm0.62}$ & $28.21_{\pm5.86}$ & $33.29_{\pm0.14}$ & $47.07_{\pm13.0}$ & $44.36_{\pm5.13}$ & $42.66_{\pm7.26}$ & $35.58_{\pm5.53}$ & $43.11_{\pm2.49}$ & $42.03_{\pm8.23}$ & $41.90_{\pm5.58}$ & $37.67_{\pm12.4}$ & $\underline{74.38}_{\pm4.01}$ & $\mathbf{76.85}_{\pm7.05}$ \\
			\multirow{-3}{*}{HU$\rightarrow$MU} & Kappa & $4.130_{\pm1.47}$ & $11.79_{\pm11.8}$ & $0.070_{\pm0.15}$ & $10.37_{\pm14.8}$ & $24.74_{\pm6.89}$ & $17.80_{\pm11.0}$ & $3.610_{\pm9.93}$ & $14.48_{\pm2.40}$ & $16.26_{\pm16.3}$ & $14.53_{\pm11.0}$ & $5.610_{\pm16.14}$ & $64.14_{\pm4.86}$ & $\mathbf{66.44}_{\pm7.54}$ \\ \midrule 
			& OA & $28.94_{\pm9.62}$ & $11.11_{\pm5.29}$ & $23.36_{\pm2.53}$ & $25.06_{\pm5.91}$ & $53.84_{\pm2.39}$ & $58.51_{\pm1.57}$ & $56.17_{\pm2.02}$ & $\underline{77.89}_{\pm1.82}$ & $69.53_{\pm3.44}$ & $72.75_{\pm2.93}$ & $43.76_{\pm8.11}$ & $67.74_{\pm8.77}$ & $\mathbf{89.46}_{\pm3.35}$ \\ 
			& AA & $27.76_{\pm9.43}$ & $10.51_{\pm5.24}$ & $19.41_{\pm2.27}$ & $27.12_{\pm6.86}$ & $56.80_{\pm2.74}$ & $61.11_{\pm2.24}$ & $59.27_{\pm2.03}$ & $\underline{80.21}_{\pm1.70}$ & $71.84_{\pm3.16}$ & $75.53_{\pm2.34}$ & $44.25_{\pm8.73}$ & $68.30_{\pm9.61}$ & $\mathbf{90.47}_{\pm3.09}$ \\ 
			\multirow{-3}{*}{TR$\rightarrow$HU} & Kappa & $6.450_{\pm13.6}$ & $32.19_{\pm8.27}$ & $20.27_{\pm3.65}$ & $9.390_{\pm9.50}$ & $31.64_{\pm3.69}$ & $38.26_{\pm2.65}$ & $35.32_{\pm2.82}$ & $\underline{67.10}_{\pm2.62}$ & $54.61_{\pm4.88}$ & $59.62_{\pm4.12}$ & $16.59_{\pm12.0}$ & $51.32_{\pm13.5}$ & $\mathbf{84.12}_{\pm4.96}$ \\ \midrule 
			& OA & $66.50_{\pm3.14}$  & $68.66_{\pm6.38}$ & $62.09_{\pm3.22}$ & $49.60_{\pm13.0}$ & $67.72_{\pm2.87}$ & $70.98_{\pm2.97}$ & $66.64_{\pm1.97}$ & $63.20_{\pm2.27}$ & $68.77_{\pm4.89}$ & $67.11_{\pm3.03}$ & $73.99_{\pm6.77}$ & $\underline{74.42}_{\pm6.06}$ & $\mathbf{90.01}_{\pm2.27}$ \\ 
			& AA & $45.65_{\pm4.81}$ & $58.56_{\pm8.52}$ & $45.77_{\pm4.33}$ & $49.06_{\pm5.81}$ & $44.19_{\pm7.90}$ & $52.54_{\pm7.08}$ & $39.48_{\pm4.49}$ & $46.64_{\pm4.83}$ & $44.67_{\pm11.8}$ & $45.32_{\pm7.29}$ & $60.67_{\pm14.8}$ & $\underline{62.45}_{\pm11.2}$ & $\mathbf{86.94}_{\pm3.73}$ \\ 
			\multirow{-3}{*}{TR$\rightarrow$MU} & Kappa & $22.86_{\pm8.28}$ & $43.64_{\pm11.4}$ & $24.63_{\pm6.53}$ & $24.98_{\pm10.9}$ & $23.66_{\pm13.5}$ & $37.12_{\pm11.4}$ & $14.62_{\pm11.0}$ & $23.65_{\pm7.17}$ & $22.55_{\pm22.3}$ & $24.08_{\pm13.8}$ & $44.99_{\pm23.9}$ & $\underline{53.05}_{\pm11.8}$ & $\mathbf{81.27}_{\pm3.94}$ \\ \bottomrule
	\end{tabular}}
\vspace{-4mm}
\end{table*}

\noindent{\textbf{Implementation details}}. We select several different types of state-of-the-art methods for comparative analysis, including 1) multimodality-based methods, namely MFT~\cite{Ref-MFT}, MsFE-IFN~\cite{MRSIC2}, and CMFAEN~\cite{MRSIC3}; 2) diffusion model-based method, namely DKDMN~\cite{MHSI1}; 3) DG-based methods, namely SDENet~\cite{MMRS1}, LLURNet~\cite{SSDG2}, FDGNet~\cite{Feadis1}, TFTNet~\cite{Ref-TFTNet}, ADNet~\cite{Ref-ADNet}, and ISDGS~\cite{Ref-ISDGS}; 4) VLM \& DG-based methods, namely EHSNet~\cite{EHSNet}, and LDGNet~\cite{LDGNet}. In the relevant experiments, the learning rate, batch size, patch size, and epoch of FVMGN are set to $0.001$, $128$, $11$, and $20$, respectively. Adam optimizer with $1e$-$4$ weight decay and cosine annealing scheduler is employed for parameter optimization and learning rate adjustment. In addition, we select $10\%$ samples in the source domain for training and generalize the well-trained model to all test samples in the target domain, and all experiments are conducted on the PyTorch framework. For experimental reproducibility, we will release adjusted datasets and source code.

\noindent{\textbf{Evaluation metrics}}.
Following common evaluation metrics~\cite{MRSIC2,SSDG2,LDGNet}, we use overall accuracy (OA), average accuracy (AA), and kappa coefficient (Kappa) to assess model performance. Moreover, to ensure fairness, we calculated the mean and standard deviation of the OA, AA, and Kappa from $10$ independent experiments for comparative analysis.

\subsection{Comparison with SOTA Methods}
\noindent{\textbf{Quantitative analysis}}. As shown in Table \ref{Coms}, comparative experiments are conducted on six dataset combinations (SD$\leftrightarrow$TD: MU$\leftrightarrow$TR$\leftrightarrow$HU). Overall, the proposed FVMGN achieves $92.44$\%, $93.19$\%, $82.87$\%, $82.33$\%, $89.46$\%, and $90.01$\% in OA, respectively, demonstrating competitive classification performance compared with SOTA methods. 
It mainly benefits from two aspects: 1) FVMGN simultaneously focuses on multimodal heterogeneity and cross-scene generalization compared with diffusion-based (DKDMN), multimodality-based (SDENet, LLURNet, FDGNet, TFTNet, ADNet, and ISDGS) and DG-based (EHSNet and LDGNet) methods, achieving the effective multimodal cross-domain invariant feature extraction. 2) FVMGN designs modality-specific class text descriptions to characterize modal attributes while constructing a multiscale unified semantic space for spatial and frequency feature alignments compared with VLM \& DG-based methods.

\begin{figure}[t]
	\begin{minipage}[b]{0.115\linewidth}
		\centering
		\includegraphics[width=\textwidth]{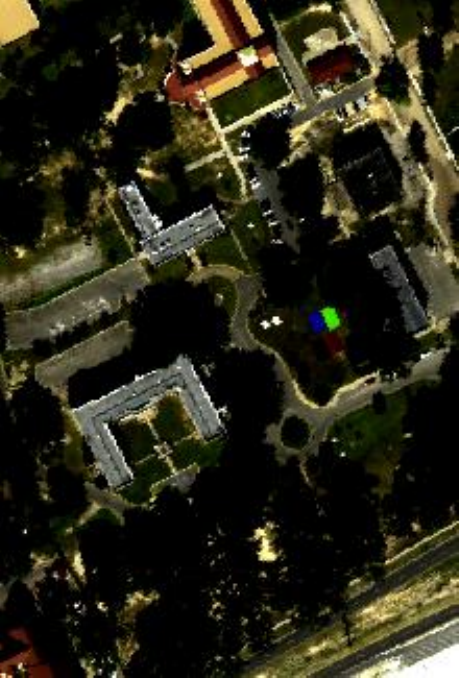}
		\centerline{(a)}\medskip
	\end{minipage}
	\hfill
	\begin{minipage}[b]{0.115\linewidth}
		\centering
		\includegraphics[width=\textwidth]{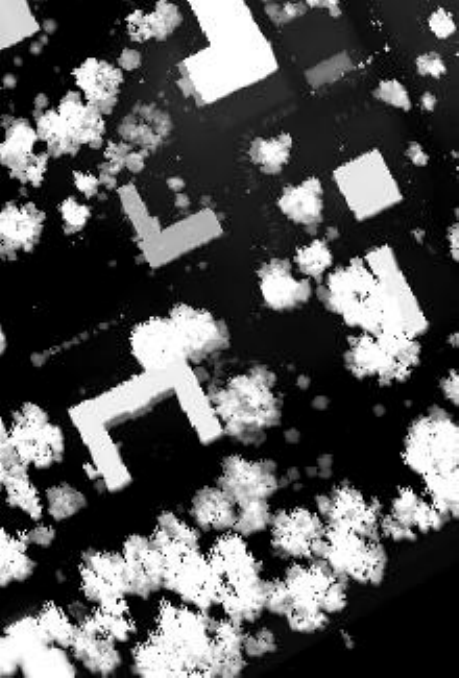}
		\centerline{(b)}\medskip
	\end{minipage}
	\hfill
	\begin{minipage}[b]{0.115\linewidth}
		\centering
		\includegraphics[width=\textwidth]{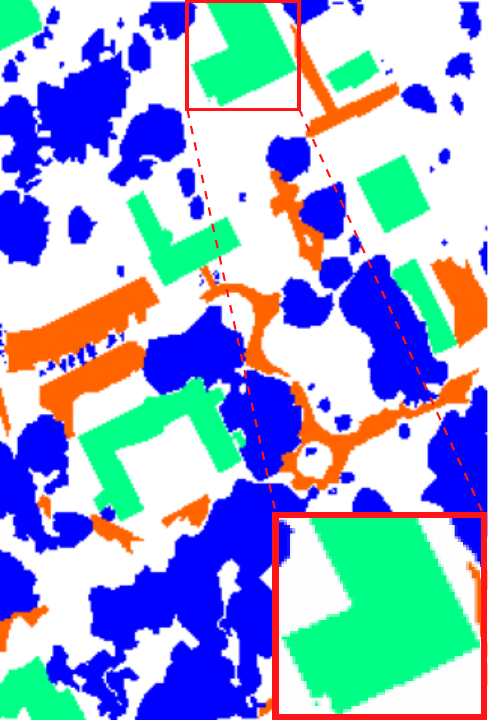}
		\centerline{(c)}\medskip
	\end{minipage}
	\hfill
	\begin{minipage}[b]{0.115\linewidth}
		\centering
		\includegraphics[width=\textwidth]{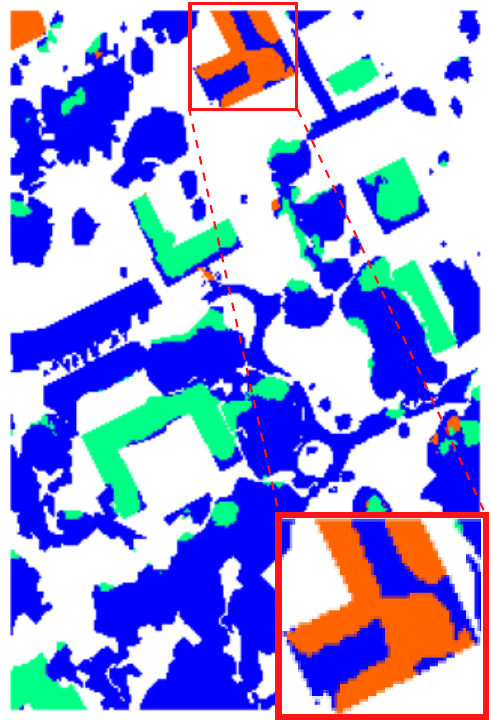}
		\centerline{(d)}\medskip
	\end{minipage}
	\hfill
	\begin{minipage}[b]{0.115\linewidth}
		\centering
		\includegraphics[width=\textwidth]{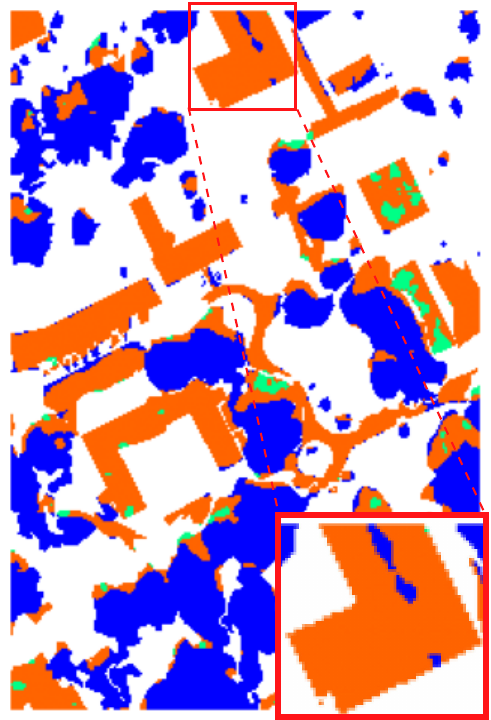}
		\centerline{(e)}\medskip
	\end{minipage}
	\hfill
	\begin{minipage}[b]{0.115\linewidth}
		\centering
		\includegraphics[width=\textwidth]{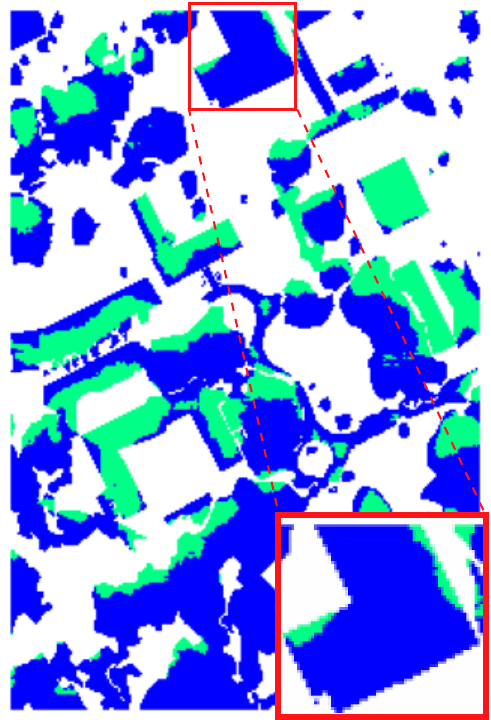}
		\centerline{(f)}\medskip
	\end{minipage}
	\hfill
	\begin{minipage}[b]{0.115\linewidth}
		\centering
		\includegraphics[width=\textwidth]{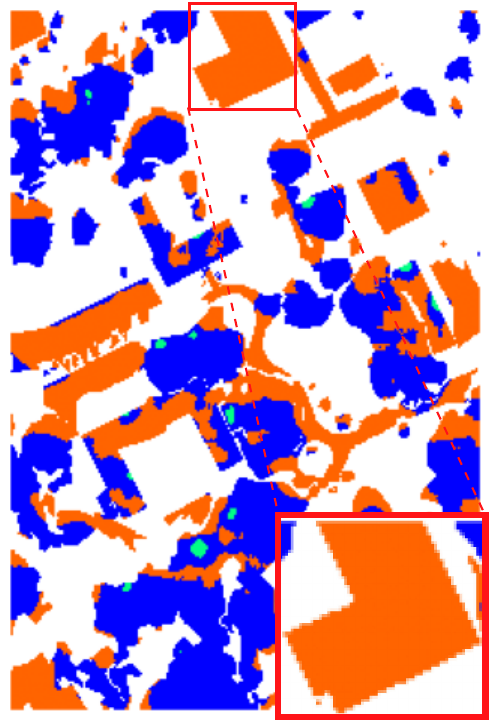}
		\centerline{(g)}\medskip
	\end{minipage}
	\hfill
	\begin{minipage}[b]{0.115\linewidth}
		\centering
		\includegraphics[width=\textwidth]{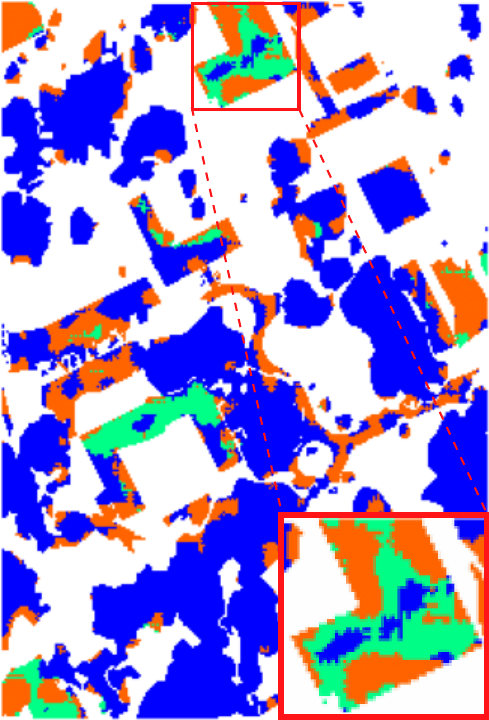}
		\centerline{(h)}\medskip
	\end{minipage}
	
	\begin{minipage}[b]{0.115\linewidth}
		\centering
		\includegraphics[width=\textwidth]{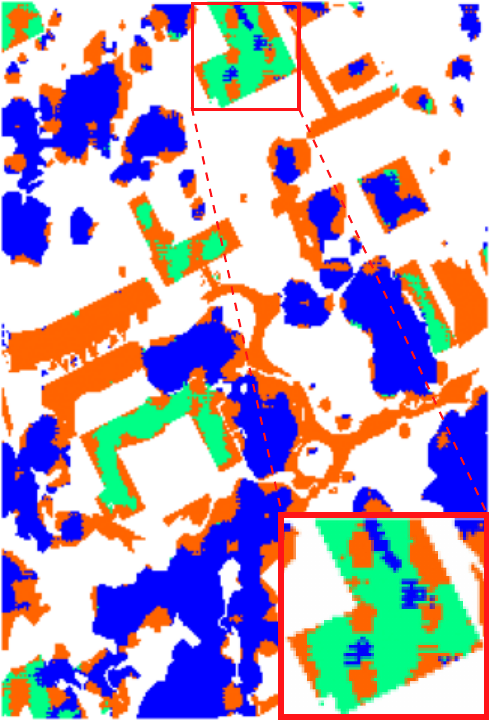}
		\centerline{(i)}\medskip
	\end{minipage}
	\hfill
	\begin{minipage}[b]{0.115\linewidth}
		\centering
		\includegraphics[width=\textwidth]{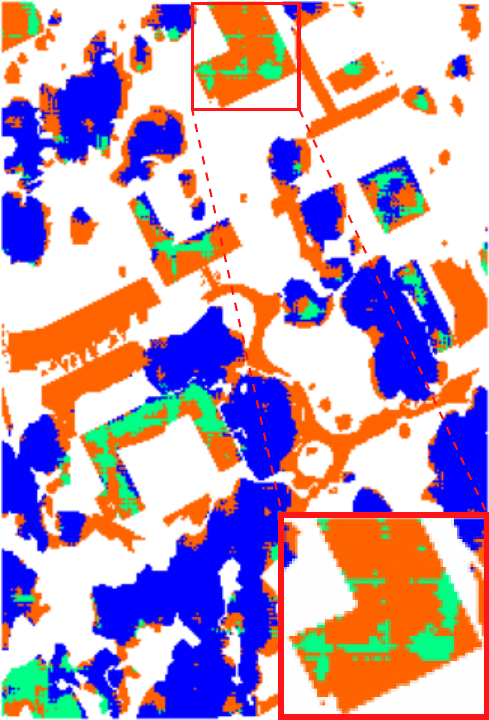}
		\centerline{(j)}\medskip
	\end{minipage}
	\hfill
	\begin{minipage}[b]{0.115\linewidth}
		\centering
		\includegraphics[width=\textwidth]{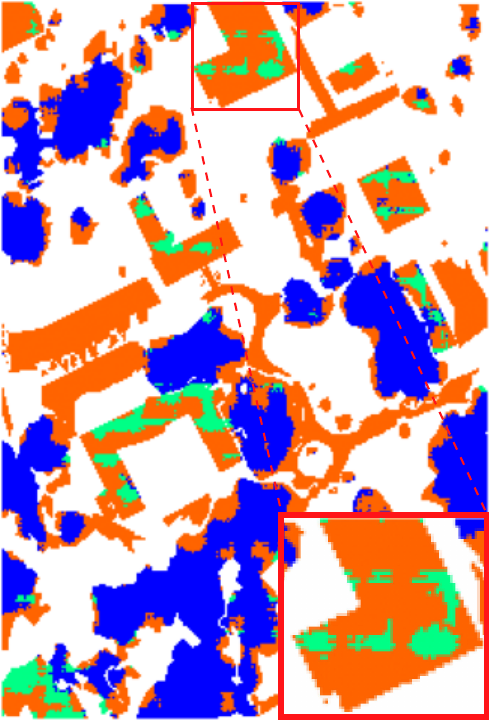}
		\centerline{(k)}\medskip
	\end{minipage}
	\hfill
	\begin{minipage}[b]{0.115\linewidth}
		\centering
		\includegraphics[width=\textwidth]{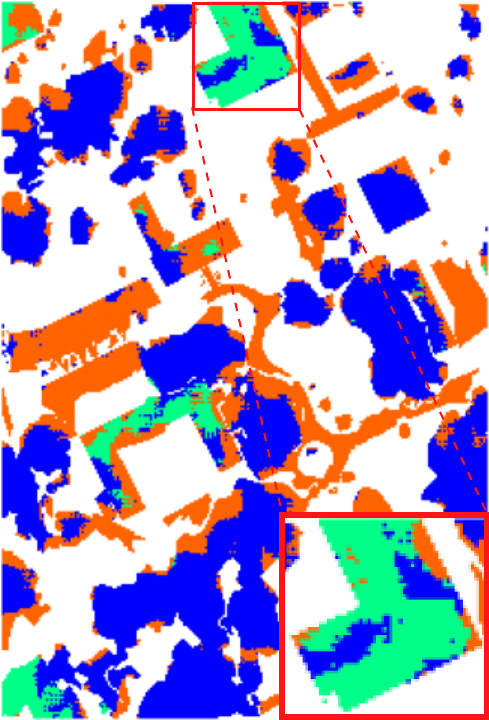}
		\centerline{(l)}\medskip
	\end{minipage}
	\hfill
	\begin{minipage}[b]{0.115\linewidth}
		\centering
		\includegraphics[width=\textwidth]{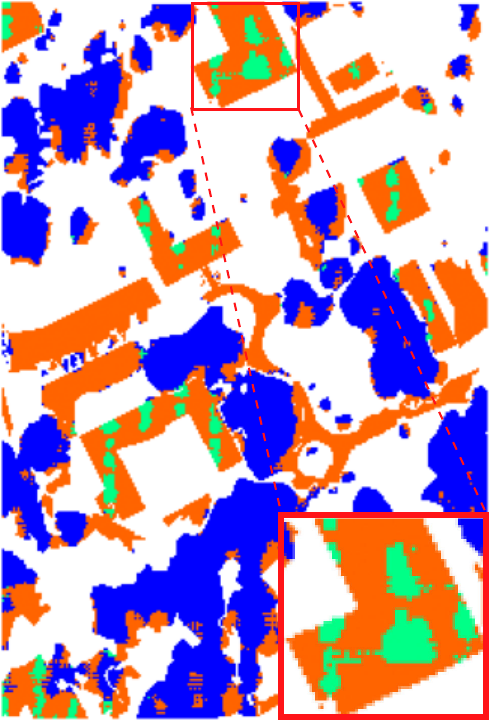}
		\centerline{(m)}\medskip
	\end{minipage}
	\hfill
	\begin{minipage}[b]{0.115\linewidth}
		\centering
		\includegraphics[width=\textwidth]{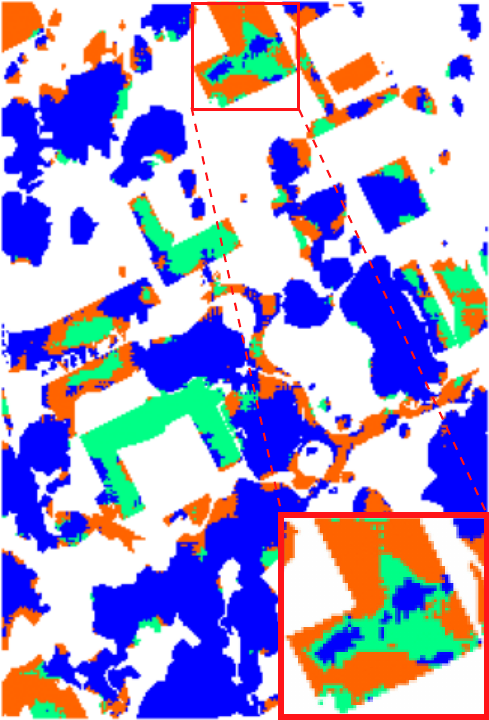}
		\centerline{(n)}\medskip
	\end{minipage}
	\hfill
	\begin{minipage}[b]{0.115\linewidth}
		\centering
		\includegraphics[width=\textwidth]{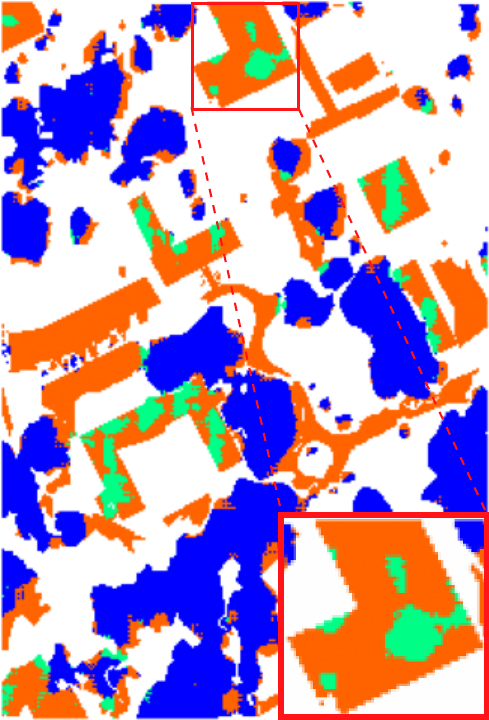}
		\centerline{(o)}\medskip
	\end{minipage}
	\hfill
	\begin{minipage}[b]{0.115\linewidth}
		\centering
		\includegraphics[width=\textwidth]{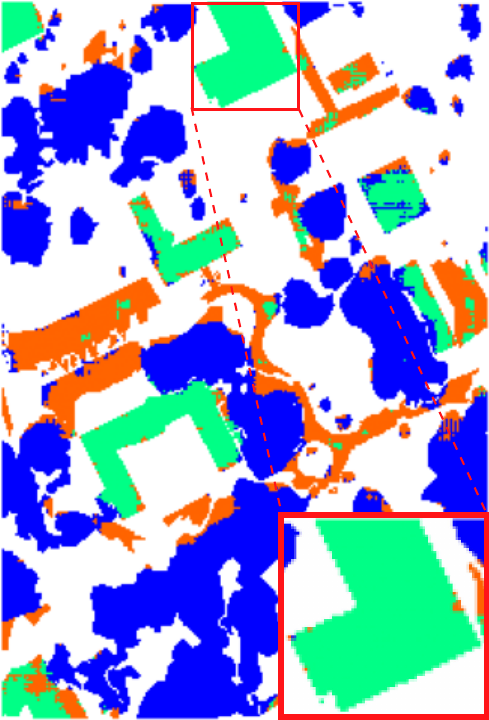}
		\centerline{(p)}\medskip
	\end{minipage}
	
	\vspace{-4mm}
	
	\caption{Classification maps (Blue/orange/green regions: Trees/Roads/Buildings) on the TR$\rightarrow$MU dataset combination. (a) HSI. (b) LiDAR image. (c) Ground truth. (d) MFT. (e) MsFE-IFN. (f) CMFAEN. (g) DKDMN. (h) SDENet. (i) LLURNet. (j) FDGNet. (k) TFTNet. (l) ADNet. (m) ISDGS. (n) EHSNet. (o) LDGNet. (p) FVMGN.} \label{comtabmuufl}
	\vspace{-6mm}
\end{figure}

\noindent{\textbf{Qualitative analysis}}. Taking the TR$\rightarrow$MU dataset combination as an example, classification maps of different methods are shown in Fig. \ref{comtabmuufl}. According to distribution difference descriptions for different datasets in \textit{Appl.Datasets}, it can be known that the source domain (Trento dataset) distribution is sparse, while the target domain (MUUFL dataset) distribution is dense, which poses certain challenges for RSMG. For example, CMFAEN fails to recognize roads as a separate class, and DKDMN recognizes a few building regions with a relatively low accuracy. Nevertheless, FVMGN still has a high classification accuracy on different class regions, and its classification map contains less noise and contaminated regions with a better visual effect.

\subsection{Ablation Study}
\noindent{\textbf{Contribution of different components}}.
We conduct performance evaluation on key components in FVMGN, as shown in Table \ref{moduleablation}. Taking the MUUFL$\rightarrow$Trento dataset combination as an example, OA of $\mathbf{NET}$-2 is 6.65\% higher than that of $\mathbf{NET}$-1, which indicates that DTAug is conducive to enhancing the input diversity. Secondly, $\mathbf{NET}$-3 achieves a 2.85\% OA improvement over $\mathbf{NET}$-2. This is mainly attributed to the fact that MWDis performs wavelet decomposition and resampling from the frequency domain, enhancing the ability to extract multimodal cross-domain invariant features. Furthermore, we apply the designed SFIE for $\mathbf{NET}$-4, achieving a 2.14\% improvement in OA. This is mainly because SFIE performs local-global feature representation and wavelet reconstruction from spatial and frequency perspectives, thereby deepening the learning and understanding for modality-specific attributes. Finally, $\mathbf{NET}$-5 (FVMGN) has a 1.75\% OA improvement by introducing the constructed MSFFA, which mainly benefits from that MSFFA achieves the refined multi-scale feature alignments for multimodal vision-vision and vision-text features in spatial and frequency domains, facilitating more accurate matching between positive sample pairs.

\begin{table}[t]\centering
	\caption{OAs of FVMGN with different parts. $\mathbf{NET}$-1 is a Baseline with the dual-branch structure of a ViT and three residual blocks, $\mathbf{NET}$-2 is $\mathbf{NET}$-1 with DTAug, $\mathbf{NET}$-3 is $\mathbf{NET}$-2 with MWDis, $\mathbf{NET}$-4 is $\mathbf{NET}$-3 with SFIE, and $\mathbf{NET}$-5 is $\mathbf{NET}$-4 with MSFFA. \label{moduleablation}}\vspace{-2mm}
	\resizebox{\linewidth}{9mm}{
		\begin{tabular}{ccccccc}
			\toprule
			Network & MU$\rightarrow$TR & MU$\rightarrow$HU & HU$\rightarrow$TR & HU$\rightarrow$MU & TR$\rightarrow$HU & TR$\rightarrow$MU \\ \midrule
			$\mathbf{NET}$-1 & $79.05_{\pm1.17}$ & $65.00_{\pm1.29}$ & $47.82_{\pm13.3}$ & $71.88_{\pm3.24}$ & $52.04_{\pm4.05}$ & $73.47_{\pm1.89}$ \\
			$\mathbf{NET}$-2 & $85.70_{\pm0.87}$ & $82.90_{\pm1.82}$ & $73.92_{\pm1.44}$ & $73.95_{\pm2.18}$ & $76.60_{\pm1.70}$ & $86.91_{\pm1.25}$ \\
			$\mathbf{NET}$-3 & $88.55_{\pm2.58}$ & $91.54_{\pm3.00}$ & $79.20_{\pm6.56}$ & $79.20_{\pm2.88}$ & $85.42_{\pm5.28}$ & $87.67_{\pm2.23}$ \\
			$\mathbf{NET}$-4 & $90.69_{\pm2.96}$ & $92.40_{\pm5.75}$ & $80.33_{\pm6.82}$ & $79.78_{\pm4.15}$ & $88.67_{\pm2.25}$ & $89.12_{\pm1.36}$ \\
			$\mathbf{NET}$-5 & $\mathbf{92.44}_{\pm1.24}$ & $\mathbf{93.19}_{\pm2.08}$ & $\mathbf{82.87}_{\pm3.77}$ & $\mathbf{82.33}_{\pm3.75}$ & $\mathbf{89.46}_{\pm3.35}$ & $\mathbf{90.01}_{\pm2.27}$ \\ \bottomrule
	\end{tabular}}
\vspace{-2mm}
\end{table}

\noindent{\textbf{Study on different feature alignments}}. 
As shown in Table \ref{fas}, OA of FVMGN with MSFFA outperforms that of FVMGN with spatial FA (Spat-FA) and that of FVMGN with frequency FA (Freq-FA), respectively. The performance improvement brought by MSFFA mainly benefits from multiscale and multi-frequency FAs in spatial and frequency domains, which facilitates fine-grained matching of positive vision-vision features and vision-text features.

\begin{table}[t]\centering
	\caption{OAs of FVMGN with different feature alignments (FA). Spat-FA is the spatial FA, Freq-FA is the frequency FA, and MSFFA is the multiscale spatial-frequency FA. \label{fas}}\vspace{-2mm}
	\resizebox{\linewidth}{7mm}{
		\begin{tabular}{ccccccc}
			\toprule
			FA & MU$\rightarrow$TR & MU$\rightarrow$HU & HU$\rightarrow$TR & HU$\rightarrow$MU & TR$\rightarrow$HU & TR$\rightarrow$MU \\ \midrule
			Spat-FA & $89.85_{\pm{3.31}}$ & $89.68_{\pm{2.50}}$ & $77.16_{\pm{5.59}}$ & $79.64_{\pm{10.0}}$ & $85.73_{\pm{6.36}}$ & $86.15_{\pm{4.86}}$ \\
			Freq-FA & $91.93_{\pm{2.02}}$ & $86.36_{\pm{3.93}}$ & $81.43_{\pm{4.43}}$ & $80.93_{\pm{6.04}}$ & $87.11_{\pm{2.50}}$ & $88.19_{\pm{4.63}}$ \\
			MSFFA & $92.44_{\pm{1.24}}$ & $93.19_{\pm{2.08}}$ & $82.33_{\pm{3.75}}$ & $82.87_{\pm{3.77}}$ & $89.46_{\pm{3.35}}$ & $90.01_{\pm{2.27}}$ \\ \bottomrule
	\end{tabular}}
\vspace{-2mm}
\end{table}

\begin{figure}[!t]\centering\rmfamily
	\begin{minipage}[b]{0.325\linewidth}
		\centering
		\includegraphics[width=\textwidth]{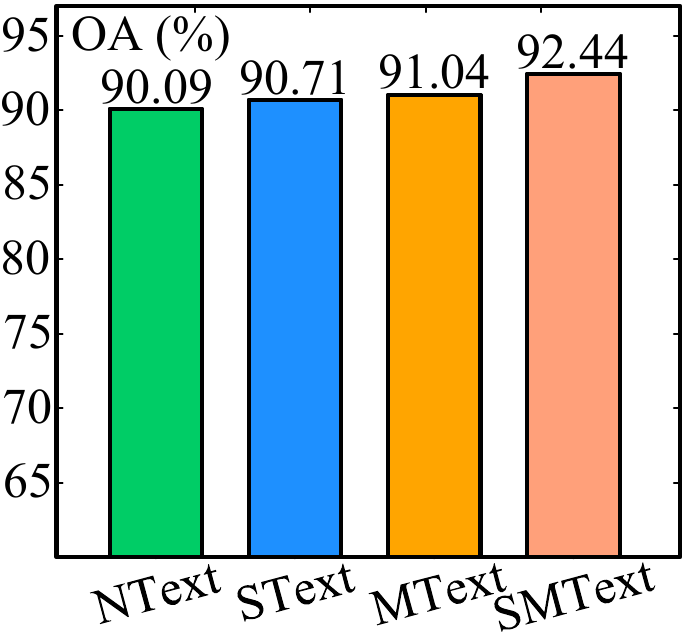}
		\centerline{(a)}\medskip
	\end{minipage}
	\hfill
	\begin{minipage}[b]{0.325\linewidth}
		\centering
		\includegraphics[width=\textwidth]{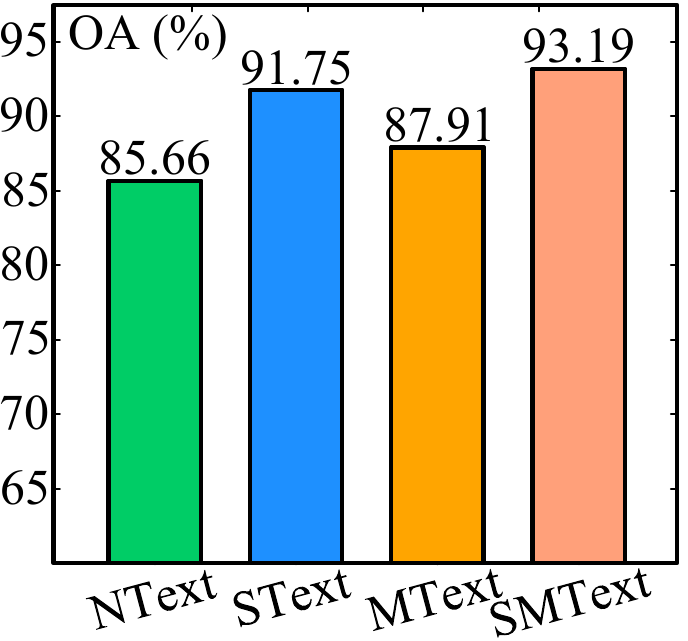}
		\centerline{(b)}\medskip
	\end{minipage}
	\hfill
	\begin{minipage}[b]{0.325\linewidth}
		\centering
		\includegraphics[width=\textwidth]{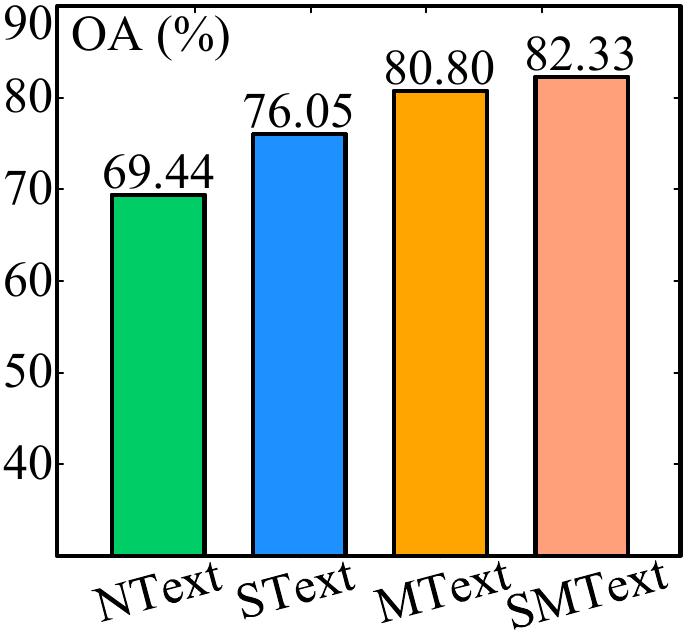}
		\centerline{(c)}\medskip
	\end{minipage}
	
	\begin{minipage}[b]{0.325\linewidth}
		\centering
		\includegraphics[width=\textwidth]{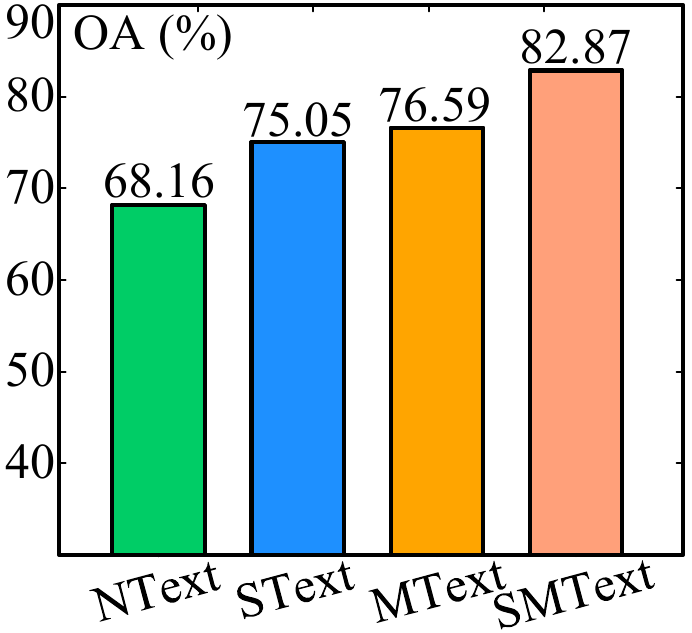}
		\centerline{(d)}\medskip
	\end{minipage}
	\hfill
	\begin{minipage}[b]{0.325\linewidth}
		\centering
		\includegraphics[width=\textwidth]{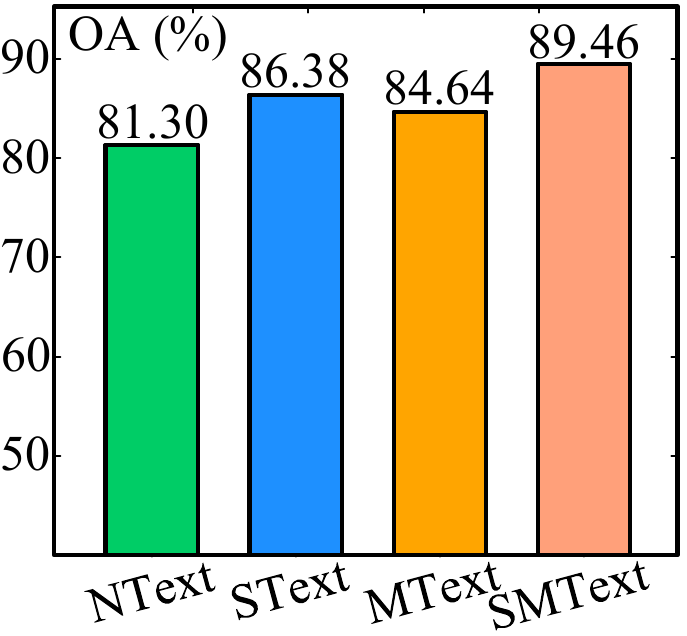}
		\centerline{(e)}\medskip
	\end{minipage}
	\hfill
	\begin{minipage}[b]{0.325\linewidth}
		\centering
		\includegraphics[width=\textwidth]{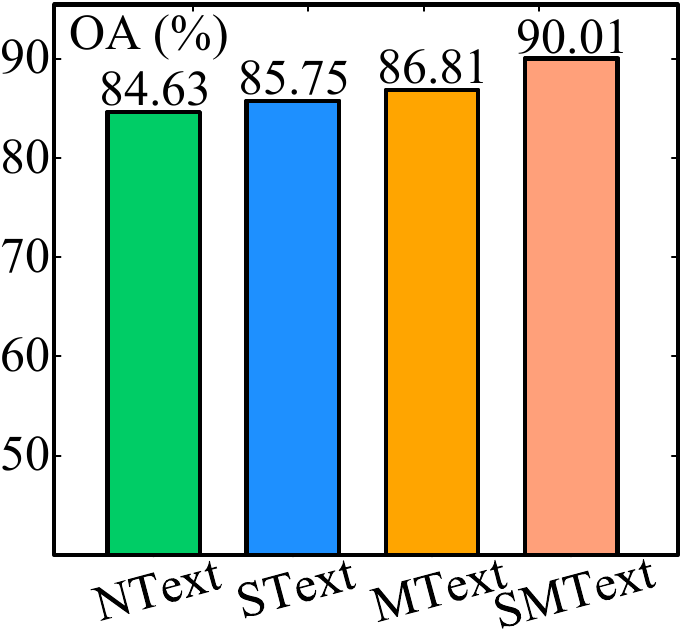}
		\centerline{(f)}\medskip
	\end{minipage}
	
	\vspace{-4mm}
	
	\caption{Classification performance of FVMGN with different texts. (a) MU$\rightarrow$TR. (b) MU$\rightarrow$HU. (c) HU$\rightarrow$TR. (d) HU$\rightarrow$MU. (e) TR$\rightarrow$HU. (f) TR$\leftrightarrow$MU. NText: No text, SText: Shared text, MText: Modality-specific (proprietary) texts, and SMText: Shared and modality-specific texts.}
	\label{textlinfluence}
	\vspace{-4mm}
\end{figure}

\noindent{\textbf{Study on different text descriptions}}. 
As shown in Fig. \ref{textlinfluence}, FVMGN with both shared and modality-specific texts achieves superior classification performance. This indicates that shared and modality-specific texts can provide general and proprietary linguistic prior knowledge for vision feature representation, respectively, thereby enhancing multimodality generalization ability.

\begin{figure}[t]
	\begin{center}
		\includegraphics[width=\linewidth]{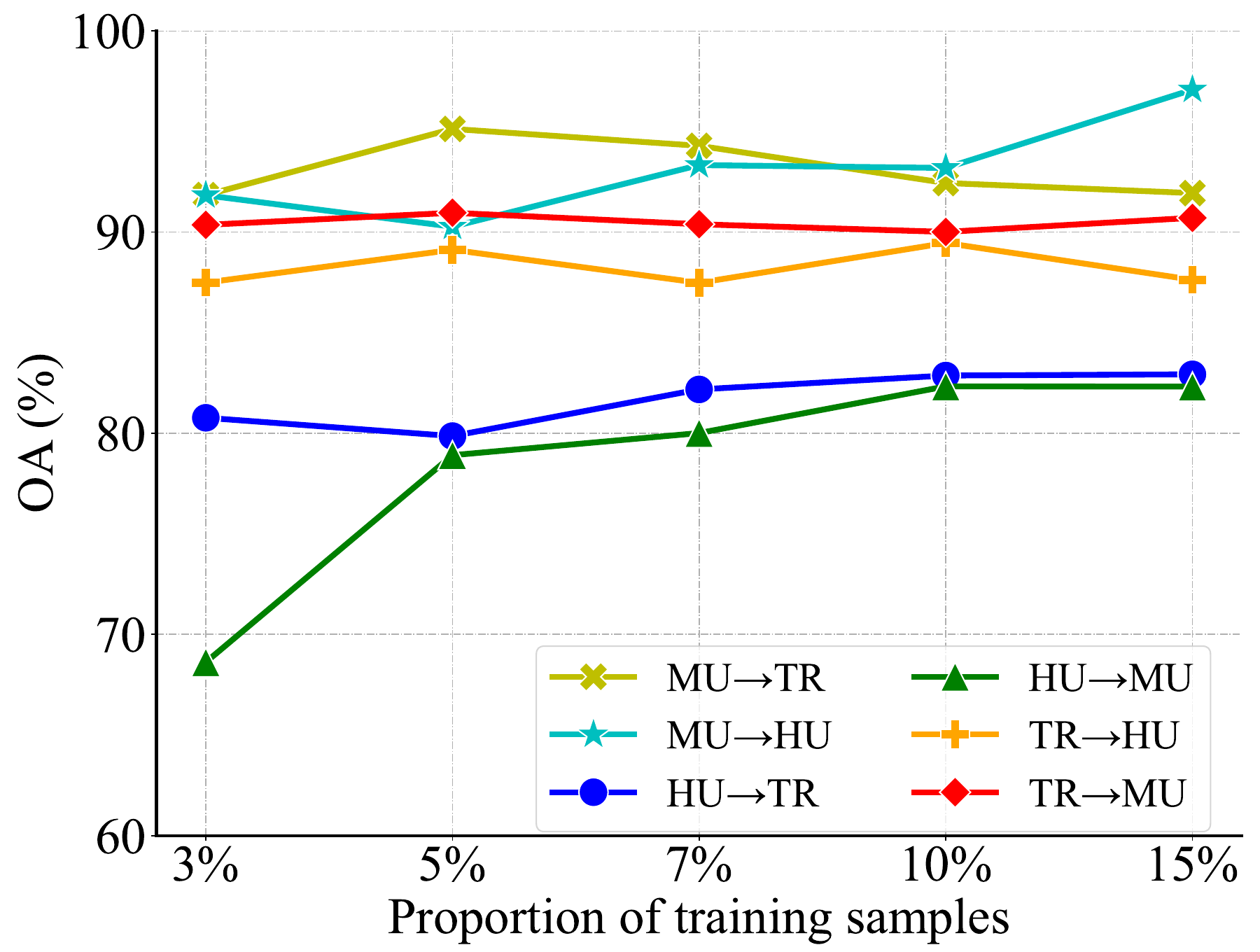}
	\end{center}
	\vspace{-2mm}
	\caption{OA of FVMGN on different dataset combinations.}
	\label{line}
\vspace{-4mm}
\end{figure}

\noindent{\textbf{Effect of training sample proportions}}.
As shown in Fig. \ref{line}, FVMGN has relatively stable classification performance on most training sample ratios. However, OA of FVMGN is poor when the proportion of training sample is 3\% on the HU$\rightarrow$MU dataset combination, mainly because there is a huge gap in the number of training samples between HU and MU datasets, making it difficult for FVMGN to learn cross-domain invariant features.

\section{Conclusion}
In this paper, we propose a FVMGN for RSIC that facilitates cross-scene multimodal information representation. Specifically, FVMGN leverages land-cover distributions generated by DTAug strategy to enrich input diversity, exploring the potential and effectiveness of DDPM in data augmentation. On the other hand, MWDis performs Gaussian modeling and histogram equalization on different frequency components in the frequency domain to learn cross-domain-invariant representations. Moreover, WCTMixer elegantly integrates the wavelet transform and attention mechanism to achieve fine multi-frequency analysis and feature reconstruction. Finally, MSFFA introduces multiscale properties and modality attributes based on the vanilla vision-text feature alignment, thereby enhancing the model ability to match and understand positive feature pairs. Extensive experiment results confirm the generalization of FVMGN.

\clearpage

\bibliography{aaai2026}

\clearpage

\section{Supplementary Materials}
\subsection{Methodology}
\begin{figure}[H]
	\begin{center}\vspace{-2mm}
		\includegraphics[width=\linewidth]{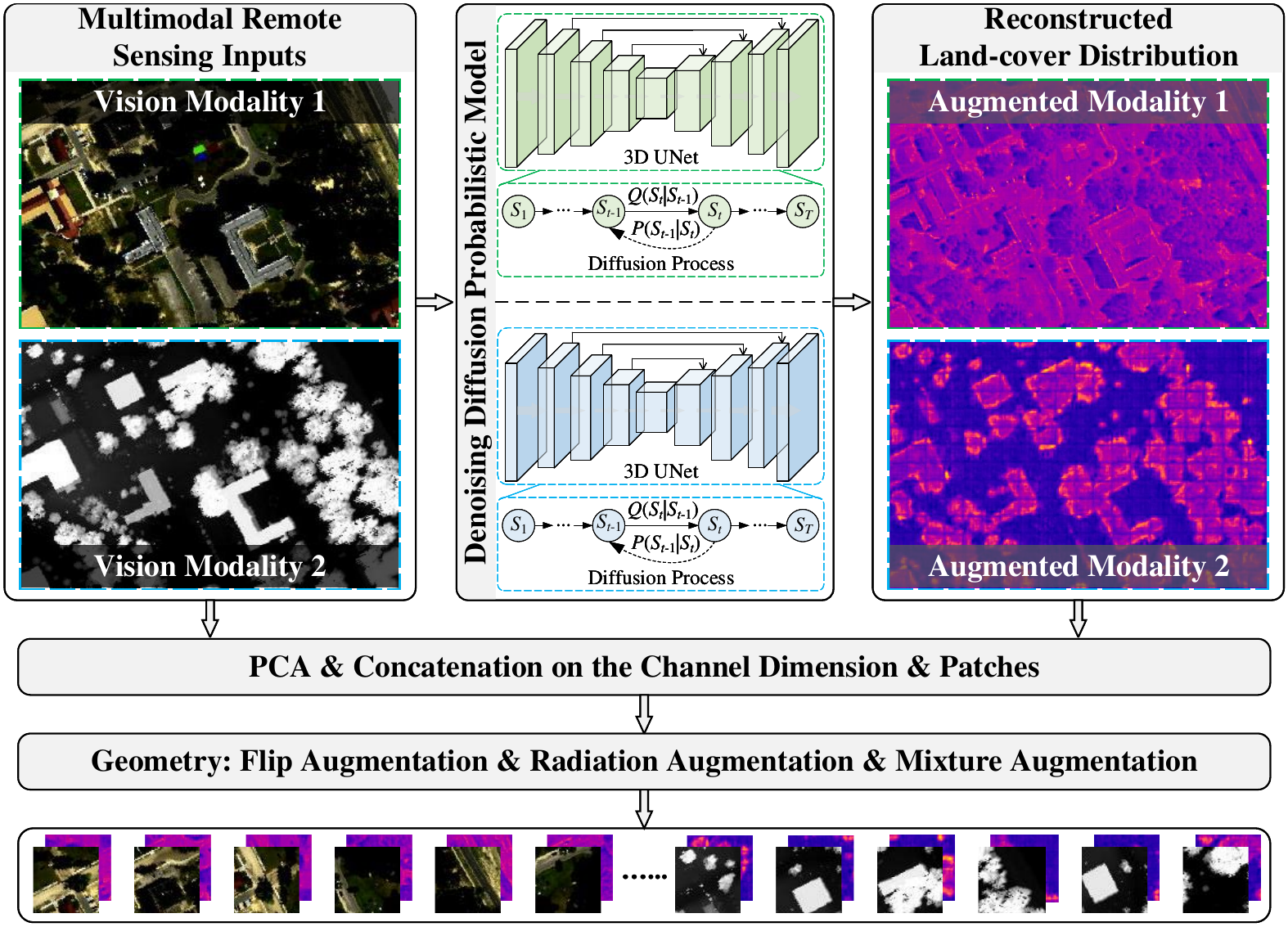}
	\end{center}\vspace{-2mm}
	\caption{Diagram of DTAug strategy.}
	\label{DTAug}
	\vspace{-2mm}
\end{figure}

\noindent{\textbf{DTAug}}. As shown in Fig. \ref{DTAug}, we present the detailed process of DTAug strategy.

\begin{figure}[H]
	\begin{center}\vspace{-2mm}
		\includegraphics[width=\linewidth]{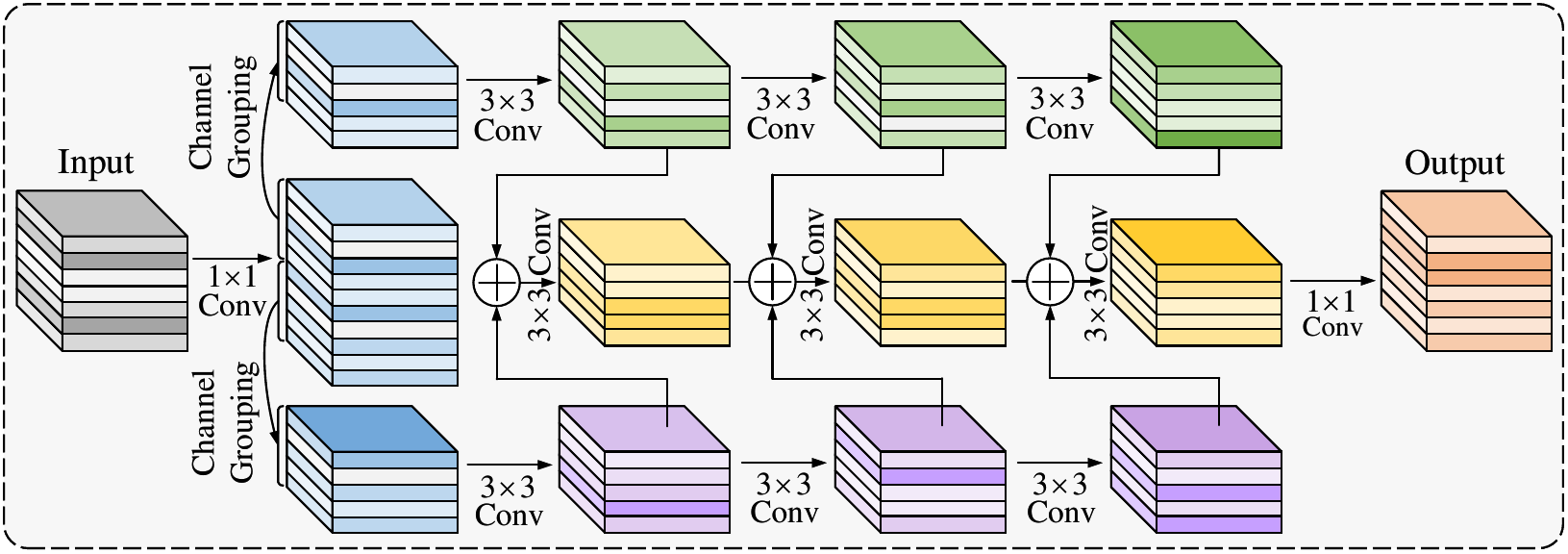}
	\end{center}\vspace{-2mm}
	\caption{Structure of FRGCM.}
	\label{FRGCM}\vspace{-2mm}
\end{figure}

\noindent{\textbf{FRGCM}}. Detailed structure of FRGCM in SFIE is shown in Fig. \ref{FRGCM}. Specifically, the input information is first fed into a $1\times1$ Conv layer to adjust the number of channels, and then the channels are divided into two groups for local feature extraction within each group, respectively. Furthermore, to achieve progressive inter-group information interaction, output features of different $3\times3$ Conv layers in each group are performed progressive summed pixel-wise summation and then fed into inter-group interaction Conv branch for further feature extraction.

\subsection{Experiments}
\begin{figure}[H]
	\begin{center}\vspace{-2mm}
		\includegraphics[width=0.99\linewidth]{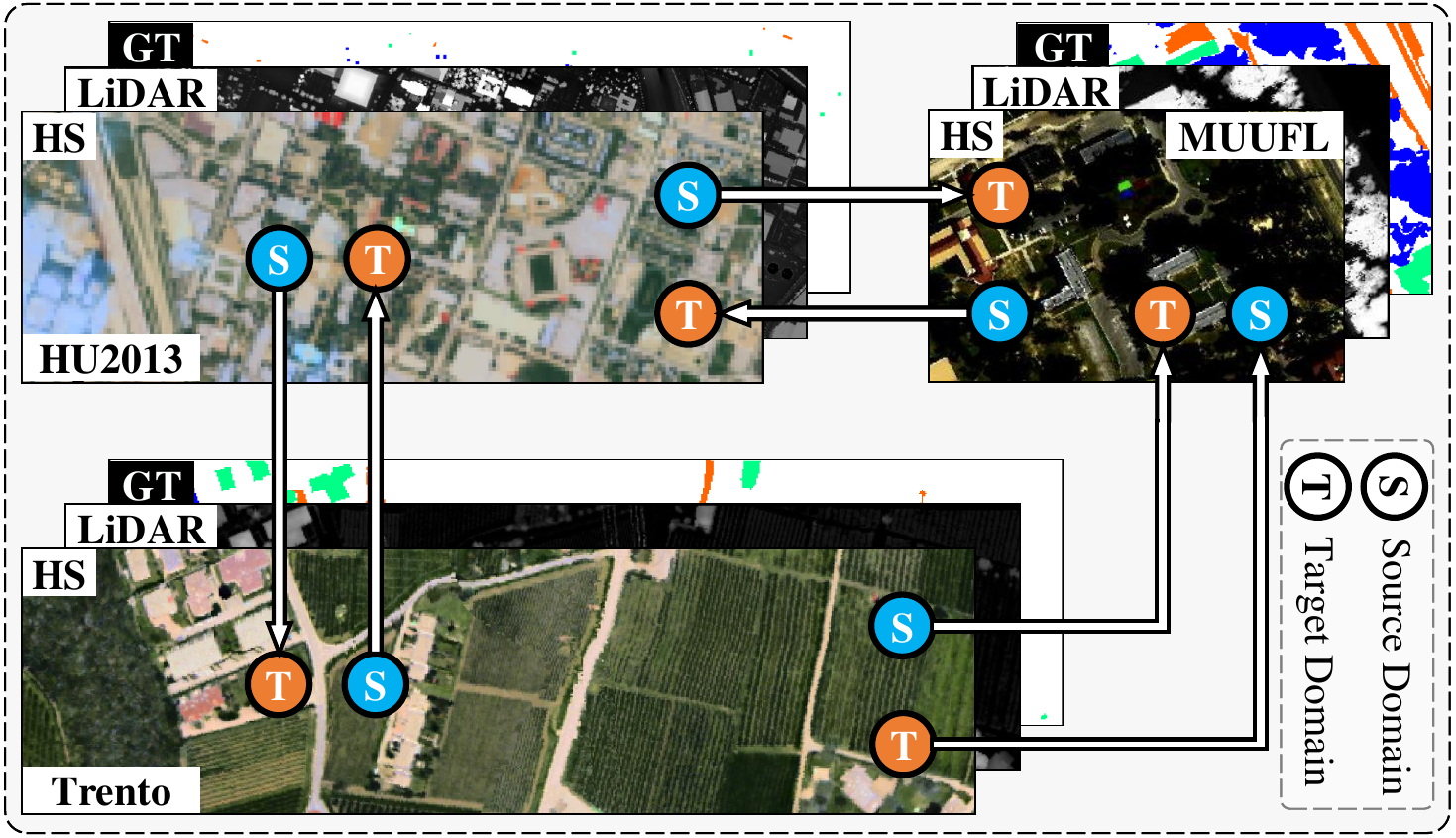}
	\end{center}\vspace{-2mm}
	\caption{Mutual multimodality generalization description. MUUFL, Trento, and HU2013 datasets are used as source and target domains for mutual generalization, respectively.}\vspace{-2mm}
	\label{Datasetsdes}
\end{figure}

\noindent{\textbf{Datasets}}.
We provide a detailed description of three datasets used in Table \ref{CLSParams}. For the MUUFL dataset, different class regions are large and densely distributed. In contrast, for the Trento dataset, different class regions are relatively sparsely distributed, and there are significant variations in region shapes. For the HU2013 dataset, different class regions show point-like patterns and have a wide distribution range. Therefore, there are significant distribution differences between the multimodality datasets, which pose higher demands on the model generalization ability. Additionally, we present a overview of mutual RSMG in Fig. \ref{Datasetsdes}, namely SD$\leftrightarrow$TD: MUUFL$\leftrightarrow$Trento$\leftrightarrow$HU2013 (MU$\leftrightarrow$TR$\leftrightarrow$HU).

\begin{table}[H]\centering
	\caption{Remote sensing dataset descriptions, including MUUFL, Trento, and HU2013 datasets.\label{CLSParams}}\vspace{-2mm}
	\resizebox{\linewidth}{13mm}{
		\begin{tabular}{c|ccccc} \toprule
			Dataset & Spatial Size & Data Type & Band No. & Class No. & Samples \\ \midrule
			& & HS & $64$ & & \\
			\multirow{-2}{*}{MUUFL} & \multirow{-2}{*}{$325\times220$} & LiDAR & $2$ & \multirow{-2}{*}{3} & \multirow{-2}{*}{$36173$} \\
			& & HS & $63$ & & \\
			\multirow{-2}{*}{Trento} & \multirow{-2}{*}{$166\times600$} & LiDAR & $1$ & \multirow{-2}{*}{3} & \multirow{-2}{*}{$15200$} \\
			& & HS & $144$ & & \\
			\multirow{-2}{*}{HU2013} & \multirow{-2}{*}{$349\times960$} & LiDAR & $1$ & \multirow{-2}{*}{3} & \multirow{-2}{*}{$2274$} \\ \bottomrule
	\end{tabular}}\vspace{-2mm}
\end{table}

\noindent{\textbf{Experimental environment}}. 
Python 3.9.18; PyTorch 2.0.1; Software VSCode; GPU NVIDIA GeForce RTX 4060; CPU Intel 12th Gen Core i7-12650H; Memory 16GB; Operation System Win 11.

\noindent{\textbf{Hyper-parameters}}. All hyper-parameters for the comparison methods are set to their default configurations.

\noindent{\textbf{Qualitative experimental results}}.
To achieve a clearer comparison for classification performance in different classes, we expand the classification results presented in the paper, as shown in Table \ref{Coms1}--\ref{Coms6}, where the symbol `$---$' denotes no pixels correctly identified.

\noindent{\textbf{Modal extensibility}}. To preliminarily verify the classification performance of the proposed method on more multimodal remote sensing datasets, we conduct experiments on the MS-SAR Augsburg (Aug) and Berlin (Ber) datasets. The two datasets are divided into four sub-parts respectively for generalization, as shown in Table \ref{more}. This datset combination is more diverse and larger scene. Here, Aug1-Ber1 refers to the first sub-part of the MS-SAR Aug and Ber datasets. We select 7 classes for the RSMG experiment. The results (OA, AA, and Kappa) show that the proposed method has good modal extensibility.

\begin{table}[H]\centering\vspace{-2mm}
	\caption{Preliminary generalization performance on the MS-SAR Augsburg and Berlin datasets. \label{more}}\vspace{-2mm}
	\resizebox{\linewidth}{9mm}{
		\begin{tabular}{ccccc}
			\toprule
			Metric & Aug1$\rightarrow$Ber1 & Aug2$\rightarrow$Ber2 & Aug3$\rightarrow$Ber3 & Aug4$\rightarrow$Ber4 \\ \midrule
			OA     & $65.18$     & $65.39$     & $72.23$     & $70.09$     \\
			AA     & $50.50$     & $43.66$     & $48.13$     & $44.10$     \\
			Kappa  & $47.26$     & $39.26$     & $61.96$     & $57.72$     \\ \bottomrule
	\end{tabular}}\vspace{-2mm}
\end{table}

\noindent{\textbf{Computational complexity}}. Computational complexity of different methods is shown in Table \ref{time}. Here, FVMGN-NW represents the FVMGN without the MWDis module. Observation shows that the proposed method has an moderate computational complexity.

\noindent{\textbf{Visualization}}. We present the vision results of the proposed method, as shown in Figs. \ref{MT}--\ref{TM}. Overall, the classification maps of the proposed method contain less noise and fewer contaminated areas, resulting in a good visual effect.

\noindent{\textbf{Manual fine-tuning for text descriptions}}: We provide GPT with modality-specific text generation prompts, and the generated texts may have redundancy such as repetition or excessive modifiers. Manual fine-tuning refers to removing irrelevant/repeated words to limit text length. Future work will explore more automated GPT post-fine-tuning process.

\noindent{\textbf{Label information}}: Different texts are generated from class labels, and they are available for training but are prohibited for inference, avoiding label information leakage.

\noindent{\textbf{Loss weight}: Weights of different loss items are the same in this work. Future work will explore adaptive weighting, such as gradient norm or soft weighting methods.
	
	\noindent{\textbf{Reproducibility}}. 
	We will release the \textit{\textbf{source code}} and the adjusted remote sensing \textit{\textbf{multimodality generalization datasets}} in this work soon, which is beneficial for the development of the remote sensing community.
	
	\begin{table*}[t]\centering
		\caption{Classification results of different methods on the MU$\rightarrow$TR dataset combination. \label{Coms1}}
		\resizebox{\linewidth}{11.5mm}{
			\begin{tabular}{cccccccccccccc}
				\toprule
				Class No. & MFT & MsFE-IFN & CMFAEN & DKDMN & SDENet & LLURNet & FDGNet & TFTNet & ADNet                 & ISDGS & EHSNet & LDGNet & FVMGN \\ \midrule
				1 & $\underline{99.80}_{\pm{0.33}}$ & $98.03_{\pm{2.40}}$ & $98.94_{\pm{1.89}}$ & $\mathbf{99.81}_{\pm{0.51}}$ & $97.98_{\pm{1.32}}$ & $98.76_{\pm{1.29}}$ & $99.64_{\pm{0.24}}$ & $98.49_{\pm{0.81}}$ & $99.64_{\pm{0.27}}$ & $98.79_{\pm{0.96}}$ & $99.71_{\pm{0.49}}$ & $98.76_{\pm{1.86}}$ & $98.08_{\pm{2.11}}$ \\
				2 & $73.98_{\pm{8.81}}$ & $86.75_{\pm{5.72}}$ & $90.39_{\pm{6.08}}$ & $55.82_{\pm{13.5}}$ & $85.28_{\pm{10.7}}$ & $90.75_{\pm{6.87}}$ & $97.16_{\pm{1.11}}$ & $77.81_{\pm{10.3}}$ & $\mathbf{98.45}_{\pm{0.61}}$ & $\underline{97.10}_{\pm{2.48}}$ & $35.39_{\pm{40.65}}$ & $72.92_{\pm{13.98}}$ & $86.85_{\pm{6.40}}$ \\
				3 & $10.59_{\pm{6.91}}$ & $36.01_{\pm{8.31}}$ & $22.94_{\pm{5.42}}$ & $37.81_{\pm{10.1}}$ & $8.000_{\pm{9.59}}$ & $0.010_{\pm{0.02}}$ & $0.070_{\pm{0.22}}$ & $5.030_{\pm{3.62}}$ & $---$ & $3.650_{\pm{6.05}}$ & $\underline{54.16}_{\pm{27.25}}$ & $45.39_{\pm{20.79}}$ & $\mathbf{80.83}_{\pm{10.0}}$ \\ \midrule
				OA & $77.59_{\pm{1.01}}$ & $\underline{83.98}_{\pm{1.86}}$ & $82.82_{\pm{1.06}}$ & $78.95_{\pm{2.38}}$ & $78.14_{\pm{1.40}}$ & $78.23_{\pm{1.42}}$ & $80.11_{\pm{0.26}}$ & $76.32_{\pm{1.67}}$ & $80.36_{\pm{0.16}}$ & $80.27_{\pm{0.67}}$ & $77.58_{\pm{4.17}}$ & $83.17_{\pm{1.63}}$ & $\mathbf{92.44}_{\pm{1.24}}$ \\
				AA & $61.46_{\pm{1.64}}$ & $\underline{73.60}_{\pm{3.61}}$ & $70.75_{\pm{1.46}}$ & $64.48_{\pm{3.77}}$ & $63.75_{\pm{2.17}}$ & $63.17_{\pm{2.20}}$ & $65.62_{\pm{0.37}}$ & $60.44_{\pm{2.72}}$ & $66.03_{\pm{0.19}}$ & $66.51_{\pm{1.23}}$ & $63.09_{\pm{6.28}}$ & $72.36_{\pm{3.41}}$ & $\mathbf{88.58}_{\pm{2.13}}$ \\
				Kappa & $53.54_{\pm{2.70}}$ & $69.25_{\pm{4.30}}$ & $66.81_{\pm{2.02}}$ & $58.08_{\pm{4.82}}$ & $60.22_{\pm{2.84}}$ & $60.49_{\pm{2.94}}$ & $64.14_{\pm{0.51}}$ & $56.15_{\pm{3.77}}$ & $64.67_{\pm{0.28}}$ & $64.59_{\pm{1.18}}$ & $58.49_{\pm{8.52}}$ & $\underline{69.65}_{\pm{3.15}}$ & $\mathbf{86.59}_{\pm{2.16}}$ \\ \bottomrule
		\end{tabular}}
	\end{table*}
	
	\begin{figure*}[t]
		\centering
		\rmfamily
		\includegraphics[width=0.05\linewidth]{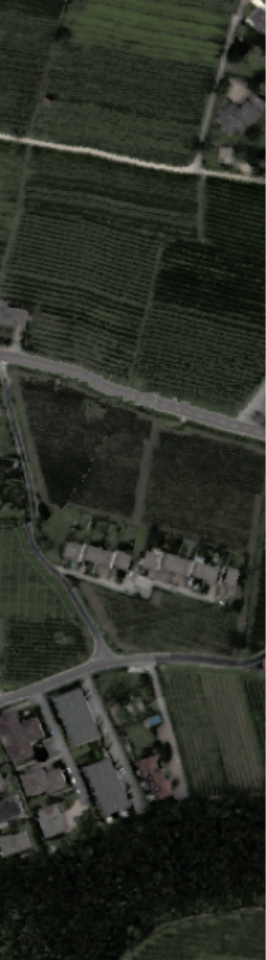}
		\hfill
		\includegraphics[width=0.05\linewidth]{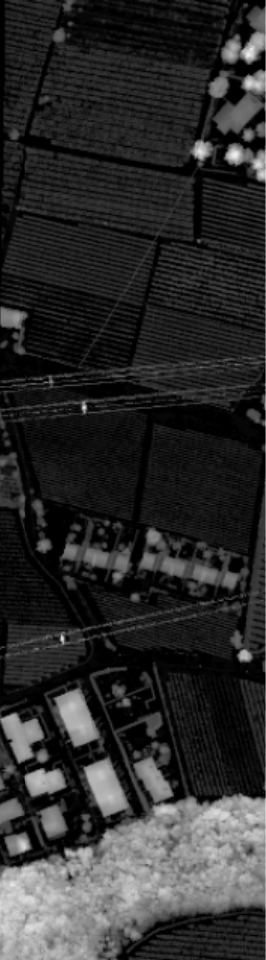}
		\hfill
		\includegraphics[width=0.05\linewidth]{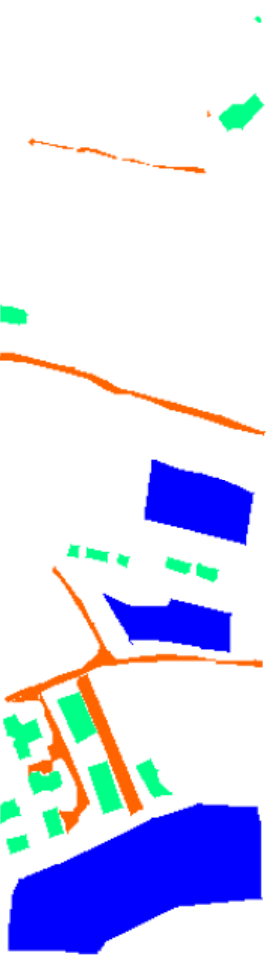}
		\hfill
		\includegraphics[width=0.05\linewidth]{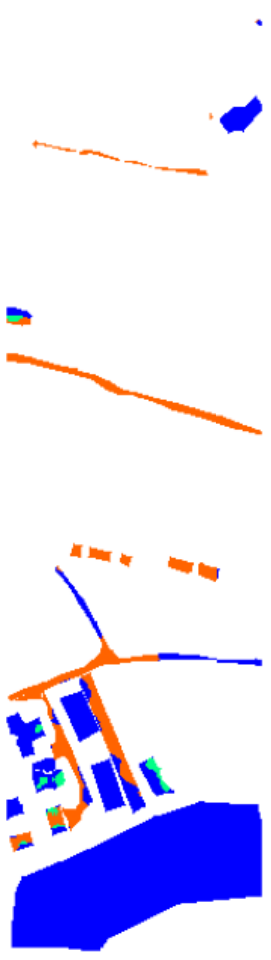}
		\hfill
		\includegraphics[width=0.05\linewidth]{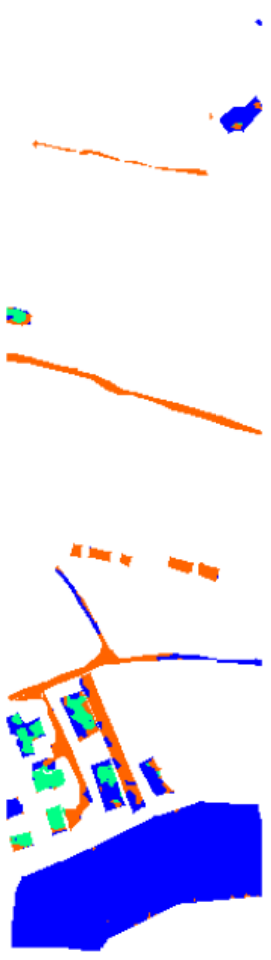}
		\hfill
		\includegraphics[width=0.05\linewidth]{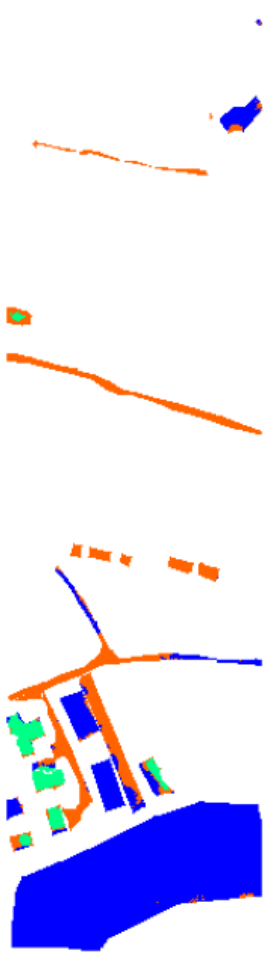}
		\hfill
		\includegraphics[width=0.05\linewidth]{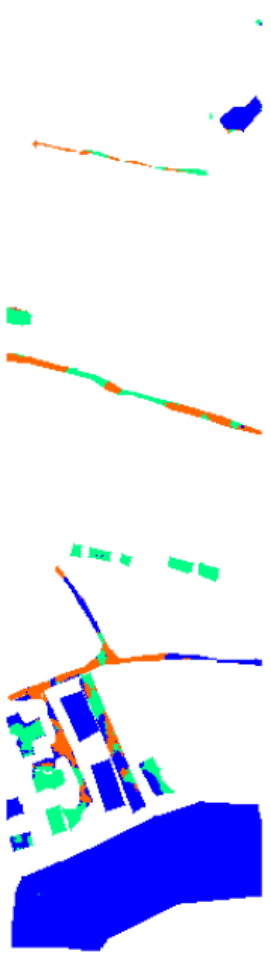}
		\hfill
		\includegraphics[width=0.05\linewidth]{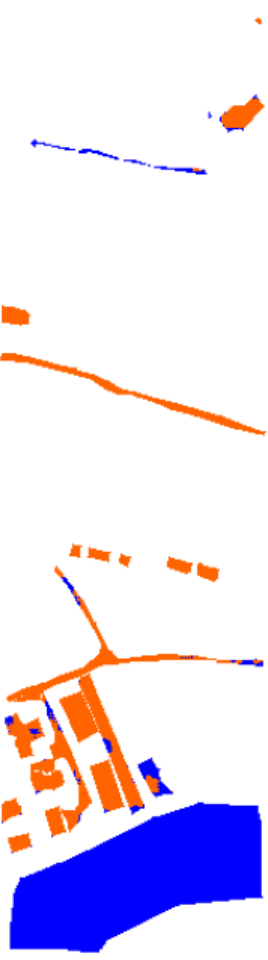}
		\hfill
		\includegraphics[width=0.05\linewidth]{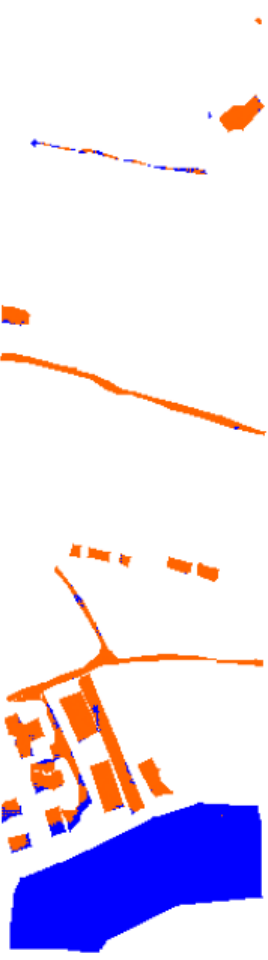}
		\hfill
		\includegraphics[width=0.05\linewidth]{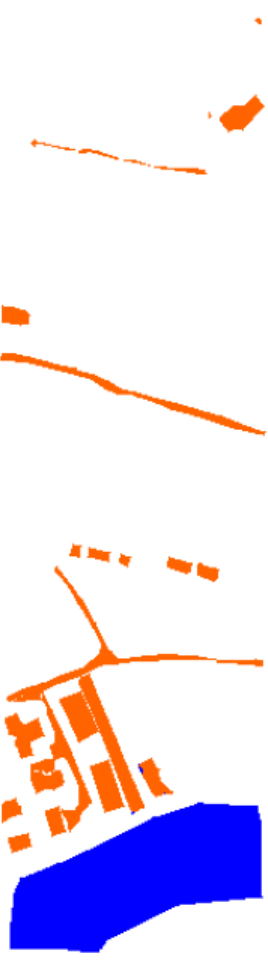}
		\hfill
		\includegraphics[width=0.05\linewidth]{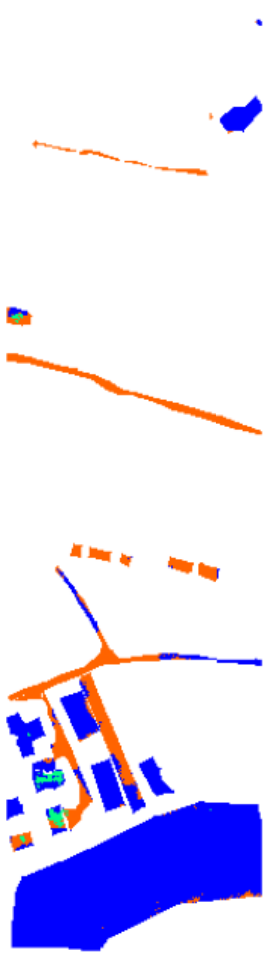}
		\hfill
		\includegraphics[width=0.05\linewidth]{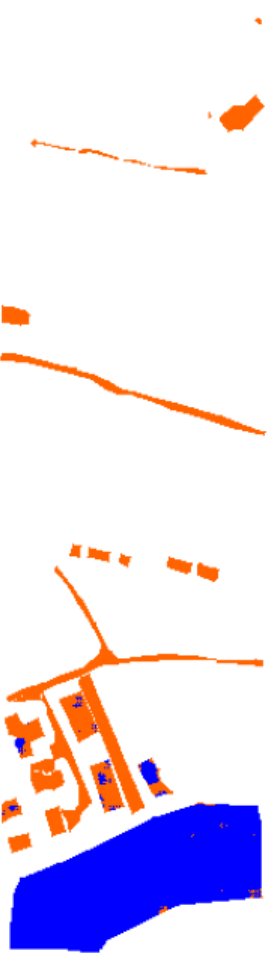}
		\hfill
		\includegraphics[width=0.05\linewidth]{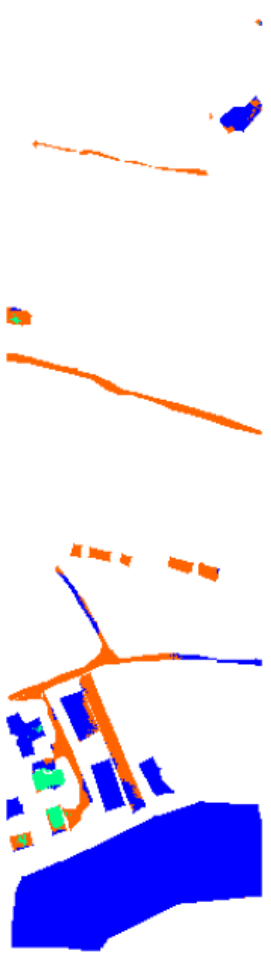}
		\hfill
		\includegraphics[width=0.05\linewidth]{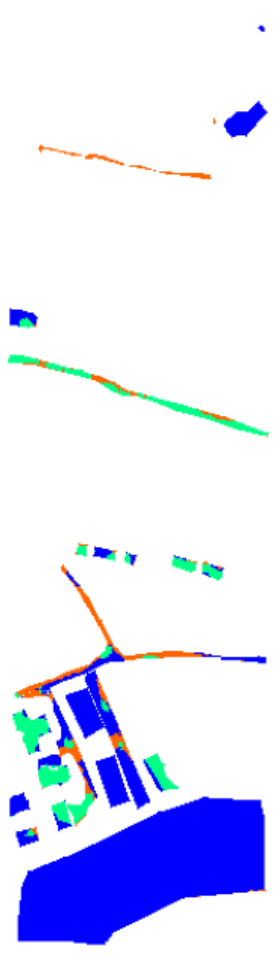}
		\hfill
		\includegraphics[width=0.05\linewidth]{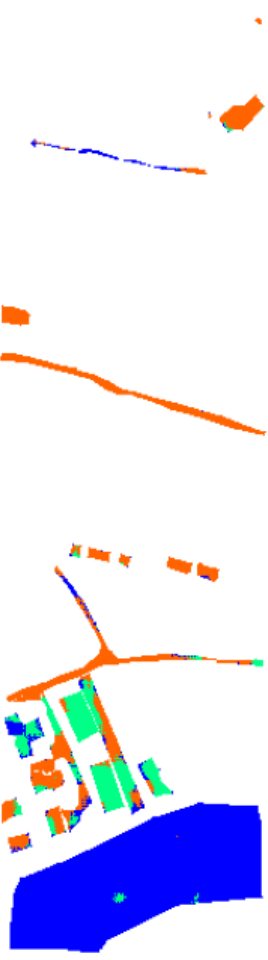}
		\hfill
		\includegraphics[width=0.05\linewidth]{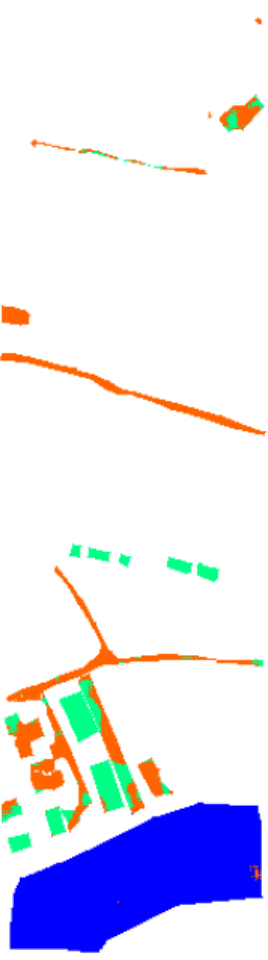}
		
		\makebox[0.05\linewidth][c]{(a)}
		\hfill
		\makebox[0.05\linewidth][c]{(b)}
		\hfill
		\makebox[0.05\linewidth][c]{(c)}
		\hfill
		\makebox[0.05\linewidth][c]{(d)}
		\hfill
		\makebox[0.05\linewidth][c]{(e)}
		\hfill
		\makebox[0.05\linewidth][c]{(f)}
		\hfill
		\makebox[0.05\linewidth][c]{(g)}
		\hfill
		\makebox[0.05\linewidth][c]{(h)}
		\hfill
		\makebox[0.05\linewidth][c]{(i)}
		\hfill
		\makebox[0.05\linewidth][c]{(j)}
		\hfill
		\makebox[0.05\linewidth][c]{(k)}
		\hfill
		\makebox[0.05\linewidth][c]{(l)}
		\hfill
		\makebox[0.05\linewidth][c]{(m)}
		\hfill
		\makebox[0.05\linewidth][c]{(n)}
		\hfill
		\makebox[0.05\linewidth][c]{(o)}
		\hfill
		\makebox[0.05\linewidth][c]{(p)}
		
		\caption{Classification maps (Blue/orange/green regions: Trees/Roads/Buildings) on the MU$\rightarrow$TR dataset combination. (a) HSI. (b) LiDAR image. (c) Ground truth. (d) MFT. (e) MsFE. (f) CMFAEN. (g) DKDMN. (h) SDENet. (i) LLURNet. (j) FDGNet. (k) TFTNet. (l) ADNet. (m) ISDGS. (n) EHSNet. (o) LDGNet. (p) FVMGN.} \label{MT}
	\end{figure*}
	
	\begin{table*}[t]\centering
		\caption{Classification results of different methods on the MU$\rightarrow$HU dataset combination. \label{Coms2}}
		\resizebox{\linewidth}{11.5mm}{
			\begin{tabular}{cccccccccccccc}
				\toprule
				Class No. & MFT & MsFE-IFN & CMFAEN & DKDMN & SDENet & LLURNet & FDGNet & TFTNet & ADNet & ISDGS & EHSNet & LDGNet & FVMGN \\ \midrule
				1 & $2.530_{\pm{1.41}}$ & $7.540_{\pm{6.56}}$ & $3.220_{\pm{2.33}}$ & $7.560_{\pm{4.18}}$ & $98.64_{\pm{0.96}}$ & $98.53_{\pm{2.04}}$ & $\underline{99.72}_{\pm{0.39}}$ & $97.40_{\pm{2.33}}$ & $\mathbf{100.0}_{\pm{0.00}}$ & $99.51_{\pm{0.66}}$ & $84.39_{\pm{14.9}}$ & $98.44_{\pm{1.41}}$ & $94.25_{\pm{3.29}}$ \\
				2 & $51.02_{\pm{18.4}}$ & $9.720_{\pm{7.65}}$ & $6.430_{\pm{6.10}}$ & $19.37_{\pm{9.51}}$ & $64.76_{\pm{11.2}}$ & $70.14_{\pm{13.1}}$ & $65.11_{\pm{6.90}}$ & $69.12_{\pm{14.9}}$ & $88.13_{\pm{10.2}}$ & $\underline{89.63}_{\pm{6.11}}$ & $7.11_{\pm{11.12}}$ & $16.53_{\pm{13.71}}$ & $\mathbf{93.11}_{\pm{2.85}}$ \\
				3 & $29.35_{\pm{20.4}}$ & $28.57_{\pm{5.15}}$ & $\underline{48.90}_{\pm{11.1}}$ & $13.37_{\pm{4.51}}$ & $33.33_{\pm{7.69}}$ & $26.13_{\pm{11.5}}$ & $19.39_{\pm{2.96}}$ & $46.44_{\pm{8.07}}$ & $37.26_{\pm{9.28}}$ & $24.23_{\pm{5.21}}$ & $34.06_{\pm{26.1}}$ & $38.30_{\pm{11.5}}$ & $\mathbf{92.50}_{\pm{3.28}}$ \\ \midrule
				OA & $27.64_{\pm{4.44}}$ & $16.93_{\pm{1.98}}$ & $23.17_{\pm{4.11}}$ & $13.38_{\pm{3.03}}$ & $61.60_{\pm{1.76}}$ & $60.11_{\pm{1.62}}$ & $56.20_{\pm{1.15}}$ & $67.96_{\pm{2.07}}$ & $\underline{70.39}_{\pm{2.71}}$ & $65.24_{\pm{1.63}}$ & $41.06_{\pm{9.83}}$ & $49.68_{\pm{2.10}}$ & $\mathbf{93.19}_{\pm{2.08}}$ \\
				AA & $27.63_{\pm{3.03}}$ & $15.28_{\pm{2.57}}$ & $19.51_{\pm{3.32}}$ & $13.43_{\pm{3.19}}$ & $65.58_{\pm{2.12}}$ & $64.93_{\pm{1.88}}$ & $61.41_{\pm{1.57}}$ & $70.98_{\pm{2.68}}$ & $\underline{75.13}_{\pm{2.47}}$ & $71.12_{\pm{1.54}}$ & $41.85_{\pm{7.83}}$ & $51.09_{\pm{2.09}}$ & $\mathbf{93.28}_{\pm{1.99}}$ \\
				Kappa & $9.930_{\pm{5.12}}$ & $26.73_{\pm{4.70}}$ & $21.04_{\pm{5.45}}$ & $30.22_{\pm{4.27}}$ & $43.83_{\pm{2.75}}$ & $42.16_{\pm{2.21}}$ & $36.81_{\pm{1.85}}$ & $52.73_{\pm{3.13}}$ & $\underline{56.93}_{\pm{3.78}}$ & $49.86_{\pm{2.30}}$ & $14.16_{\pm{12.5}}$ & $24.76_{\pm{2.97}}$ & $\mathbf{89.64}_{\pm{3.15}}$ \\ \bottomrule
		\end{tabular}}
	\end{table*}
	
	\begin{figure*}[t]
		\centering
		\rmfamily
		\includegraphics[width=0.05\linewidth]{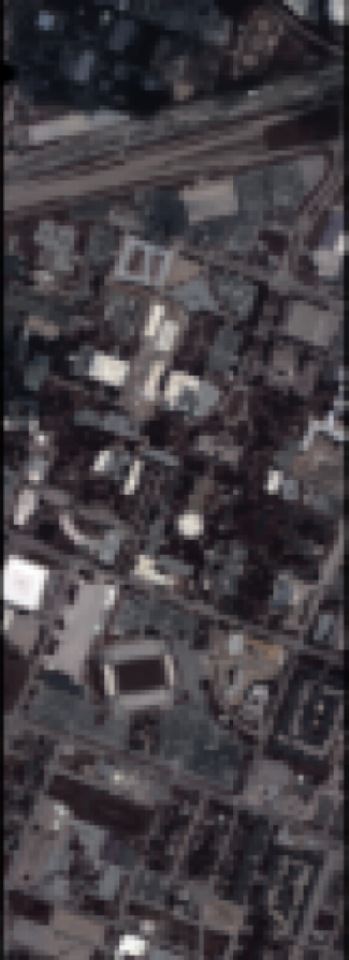}
		\hfill
		\includegraphics[width=0.05\linewidth]{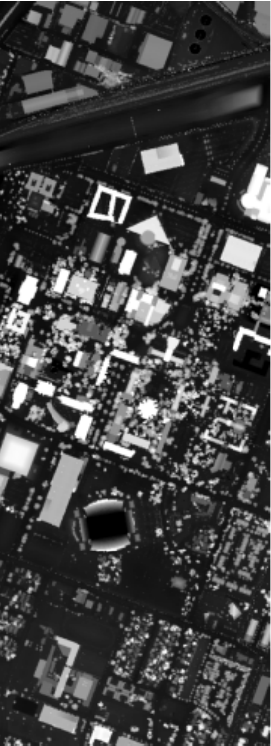}
		\hfill
		\includegraphics[width=0.05\linewidth]{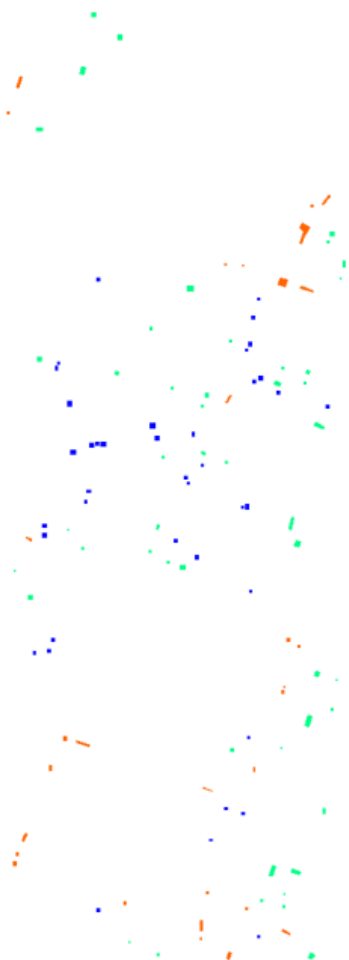}
		\hfill
		\includegraphics[width=0.05\linewidth]{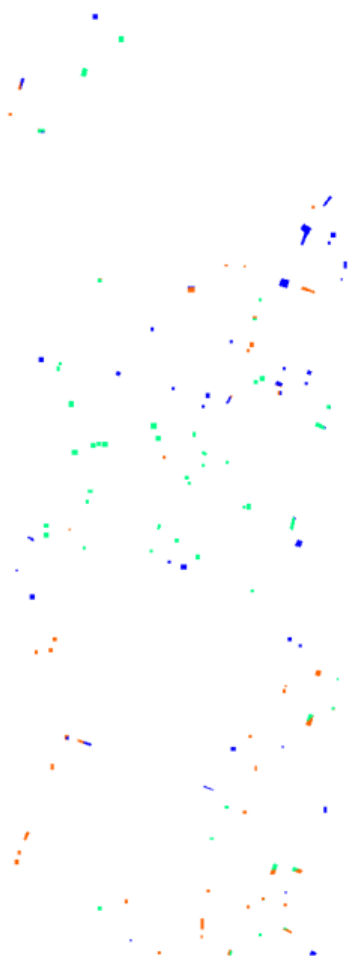}
		\hfill
		\includegraphics[width=0.05\linewidth]{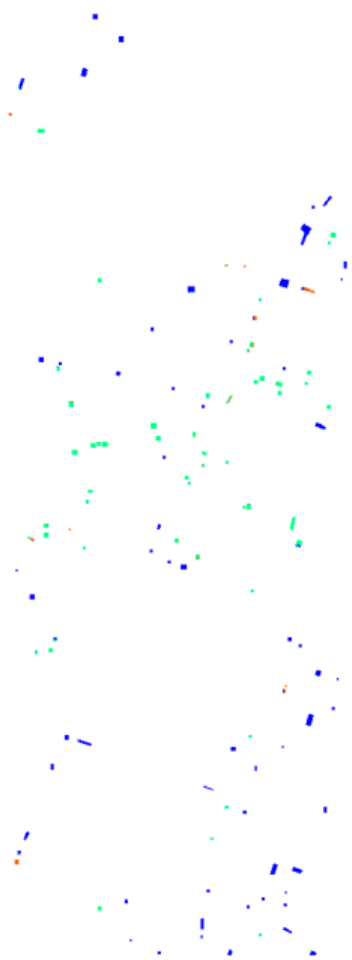}
		\hfill
		\includegraphics[width=0.05\linewidth]{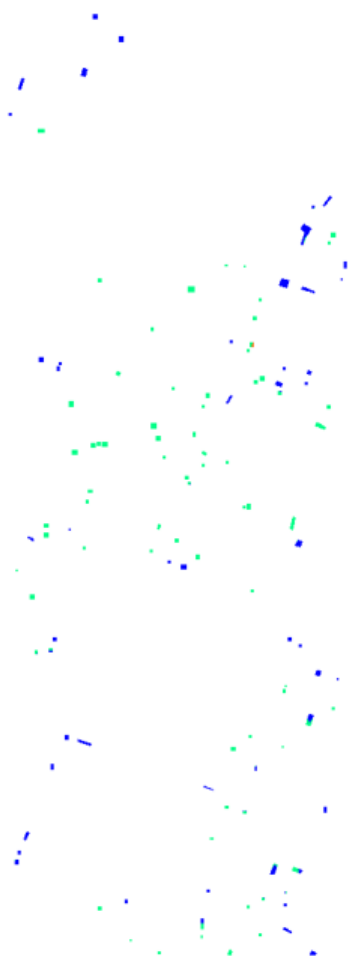}
		\hfill
		\includegraphics[width=0.05\linewidth]{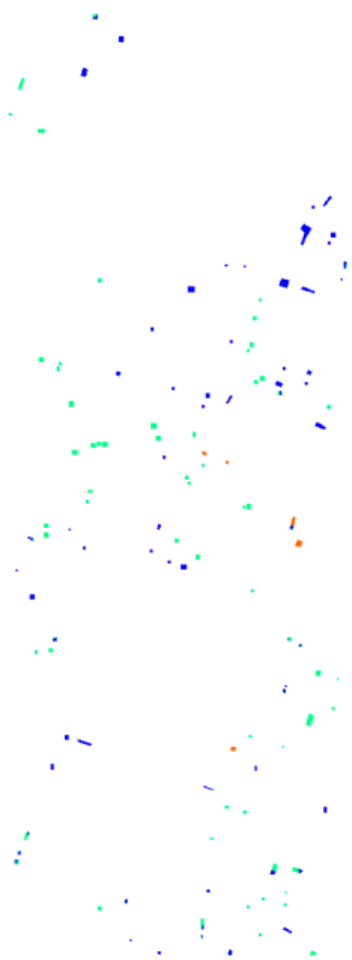}
		\hfill
		\includegraphics[width=0.05\linewidth]{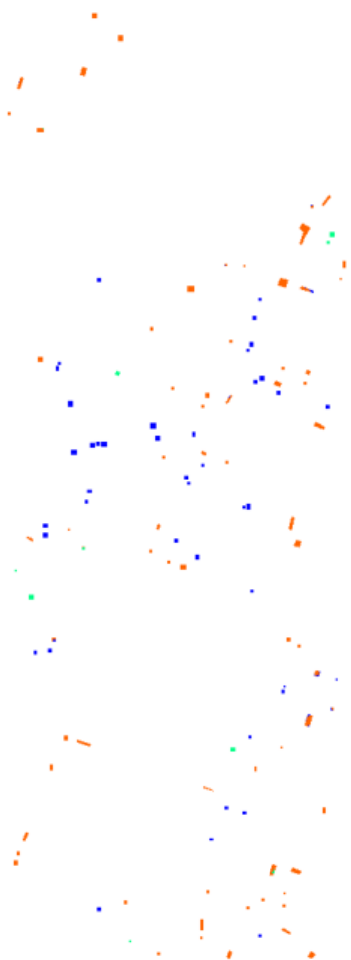}
		\hfill
		\includegraphics[width=0.05\linewidth]{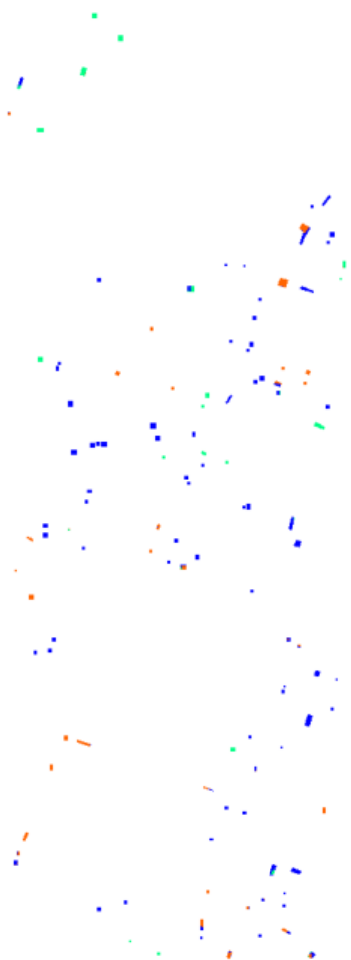}
		\hfill
		\includegraphics[width=0.05\linewidth]{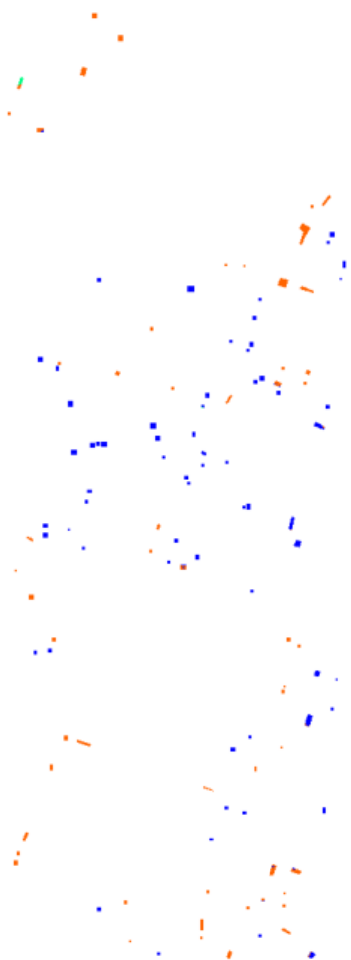}
		\hfill
		\includegraphics[width=0.05\linewidth]{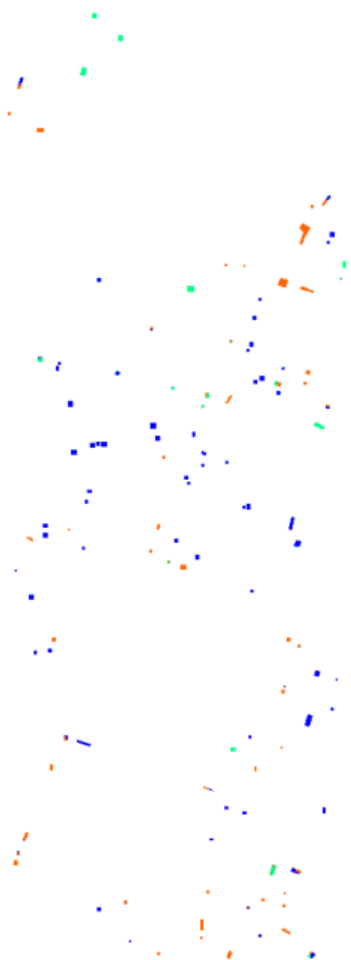}
		\hfill
		\includegraphics[width=0.05\linewidth]{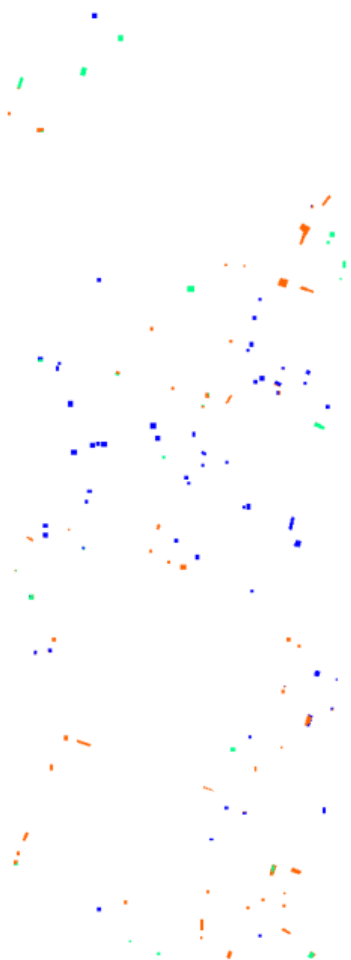}
		\hfill
		\includegraphics[width=0.05\linewidth]{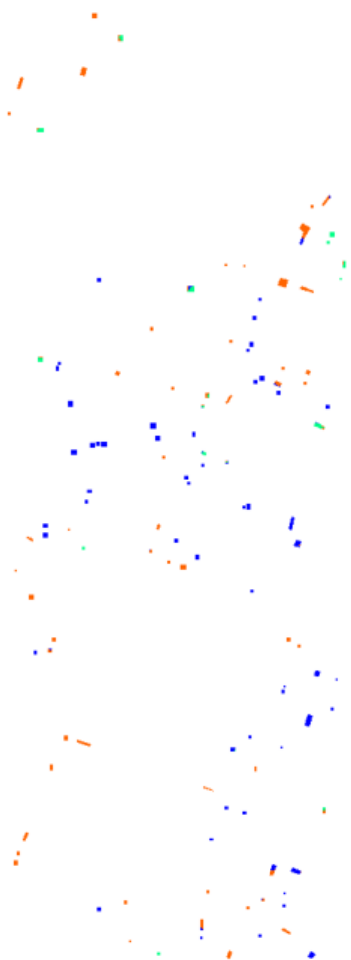}
		\hfill
		\includegraphics[width=0.05\linewidth]{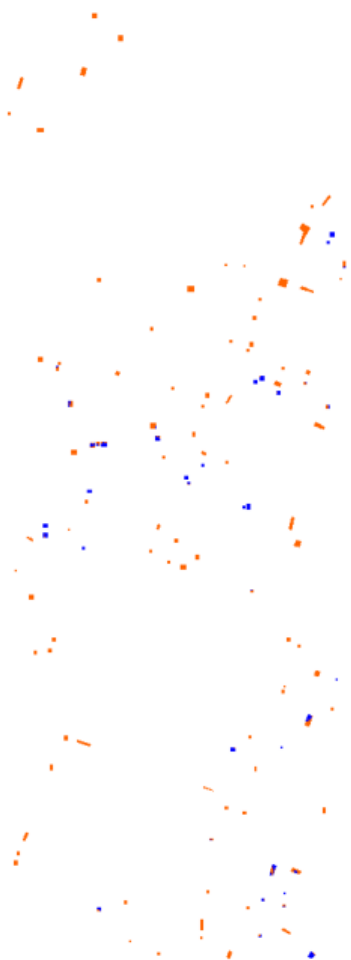}
		\hfill
		\includegraphics[width=0.05\linewidth]{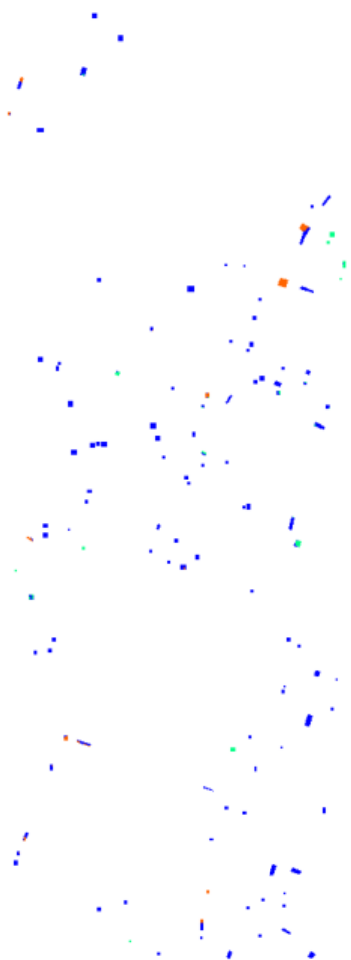}
		\hfill
		\includegraphics[width=0.05\linewidth]{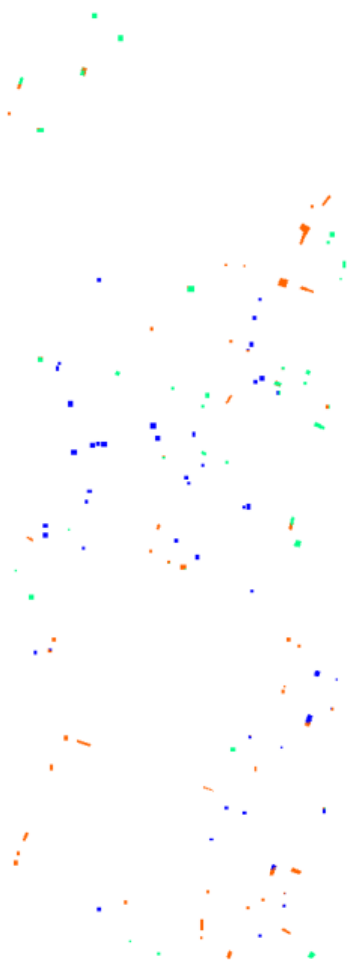}
		
		\makebox[0.05\linewidth][c]{(a)}
		\hfill
		\makebox[0.05\linewidth][c]{(b)}
		\hfill
		\makebox[0.05\linewidth][c]{(c)}
		\hfill
		\makebox[0.05\linewidth][c]{(d)}
		\hfill
		\makebox[0.05\linewidth][c]{(e)}
		\hfill
		\makebox[0.05\linewidth][c]{(f)}
		\hfill
		\makebox[0.05\linewidth][c]{(g)}
		\hfill
		\makebox[0.05\linewidth][c]{(h)}
		\hfill
		\makebox[0.05\linewidth][c]{(i)}
		\hfill
		\makebox[0.05\linewidth][c]{(j)}
		\hfill
		\makebox[0.05\linewidth][c]{(k)}
		\hfill
		\makebox[0.05\linewidth][c]{(l)}
		\hfill
		\makebox[0.05\linewidth][c]{(m)}
		\hfill
		\makebox[0.05\linewidth][c]{(n)}
		\hfill
		\makebox[0.05\linewidth][c]{(o)}
		\hfill
		\makebox[0.05\linewidth][c]{(p)}
		
		\caption{Classification maps (Blue/orange/green regions: Trees/Roads/Buildings) on the MU$\rightarrow$HU dataset combination. (a) HSI. (b) LiDAR image. (c) Ground truth. (d) MFT. (e) MsFE. (f) CMFAEN. (g) DKDMN. (h) SDENet. (i) LLURNet. (j) FDGNet. (k) TFTNet. (l) ADNet. (m) ISDGS. (n) EHSNet. (o) LDGNet. (p) FVMGN.} \label{MH}
	\end{figure*}
	
	\begin{table*}[t]\centering
		\caption{Classification results of different methods on the HU$\rightarrow$TR dataset combination. \label{Coms3}}
		\resizebox{\linewidth}{11.5mm}{
			\begin{tabular}{cccccccccccccc}
				\toprule
				Class No. & MFT & MsFE-IFN & CMFAEN & DKDMN & SDENet & LLURNet & FDGNet & TFTNet & ADNet & ISDGS & EHSNet & LDGNet & FVMGN \\ \midrule
				1 & $---$ & $0.230_{\pm{0.43}}$ & $0.190_{\pm{0.32}}$ & $0.030_{\pm{0.06}}$ & $83.49_{\pm{11.3}}$ & $79.92_{\pm{8.75}}$ & $63.67_{\pm{19.7}}$ & $78.11_{\pm{6.53}}$ & $\underline{90.81}_{\pm{5.55}}$ & $52.78_{\pm{10.6}}$ & $88.37_{\pm{10.3}}$ & $\mathbf{93.92}_{\pm{5.55}}$ & $90.48_{\pm{6.43}}$ \\
				2 & $\mathbf{92.32}_{\pm{4.79}}$ & $34.64_{\pm{32.8}}$ & $51.39_{\pm{34.8}}$ & $\underline{90.77}_{\pm{2.95}}$ & $1.160_{\pm{1.30}}$ & $3.340_{\pm{3.45}}$ & $4.620_{\pm{4.03}}$ & $4.20_{\pm{1.74}}$ & $3.450_{\pm{2.47}}$ & $2.53_{\pm{2.67}}$ & $18.43_{\pm{17.1}}$ & $19.47_{\pm{19.9}}$ & $55.47_{\pm{22.7}}$ \\
				3 & $21.69_{\pm{6.75}}$ & $70.52_{\pm{14.8}}$ & $84.44_{\pm{12.5}}$ & $53.66_{\pm{14.9}}$ & $98.24_{\pm{1.76}}$ & $94.31_{\pm{4.96}}$ & $97.54_{\pm{3.81}}$ & $\underline{98.98}_{\pm{0.81}}$ & $97.79_{\pm{2.80}}$ & $\mathbf{99.05}_{\pm{1.91}}$ & $73.93_{\pm{29.6}}$ & $75.63_{\pm{23.2}}$ & $88.90_{\pm{13.0}}$ \\ \midrule
				OA & $23.32_{\pm{0.86}}$ & $20.66_{\pm{4.94}}$ & $26.75_{\pm{6.53}}$ & $29.05_{\pm{2.64}}$ & $69.11_{\pm{6.76}}$ & $66.68_{\pm{5.34}}$ & $57.80_{\pm{12.1}}$ & $66.66_{\pm{3.95}}$ & $73.90_{\pm{3.45}}$ & $51.12_{\pm{6.21}}$ & $71.01_{\pm{7.27}}$ & $\underline{74.88}_{\pm{4.94}}$ & $\mathbf{82.87}_{\pm{3.77}}$ \\
				AA & $38.00_{\pm{1.51}}$ & $35.13_{\pm{7.72}}$ & $45.34_{\pm{10.3}}$ & $48.15_{\pm{4.69}}$ & $60.96_{\pm{3.72}}$ & $59.19_{\pm{3.38}}$ & $55.28_{\pm{7.05}}$ & $60.43_{\pm{2.26}}$ & $64.02_{\pm{2.12}}$ & $51.45_{\pm{3.34}}$ & $60.24_{\pm{8.56}}$ & $\underline{63.01}_{\pm{5.93}}$ & $\mathbf{78.29}_{\pm{5.23}}$ \\
				Kappa & $1.530_{\pm{1.53}}$ & $4.120_{\pm{8.09}}$ & $6.300_{\pm{9.89}}$ & $10.10_{\pm{3.58}}$ & $49.01_{\pm{8.64}}$ & $45.42_{\pm{7.05}}$ & $36.06_{\pm{13.9}}$ & $45.81_{\pm{4.97}}$ & $55.49_{\pm{4.88}}$ & $28.18_{\pm{6.35}}$ & $47.78_{\pm{13.9}}$ & $\underline{55.55}_{\pm{8.48}}$ & $\mathbf{69.93}_{\pm{6.40}}$ \\ \bottomrule
		\end{tabular}}
	\end{table*}
	
	\begin{figure*}[t]
		\centering
		\rmfamily
		\includegraphics[width=0.05\linewidth]{trentohsi}
		\hfill
		\includegraphics[width=0.05\linewidth]{trentolidar}
		\hfill
		\includegraphics[width=0.05\linewidth]{trento_gt}
		\hfill
		\includegraphics[width=0.05\linewidth]{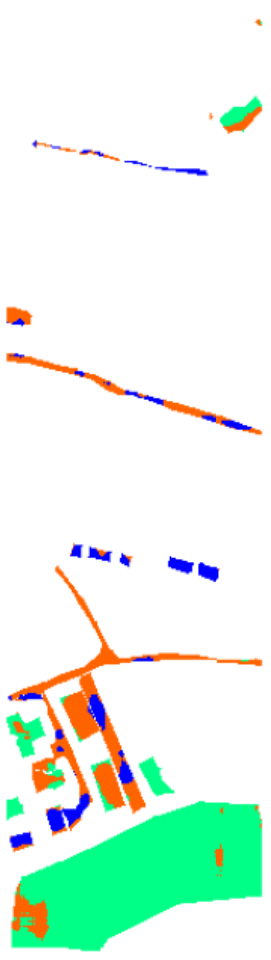}
		\hfill
		\includegraphics[width=0.05\linewidth]{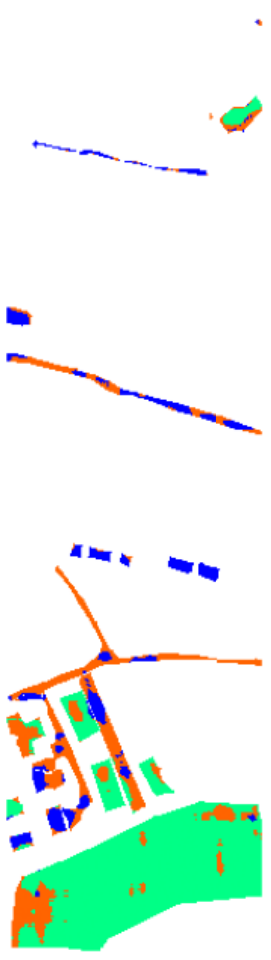}
		\hfill
		\includegraphics[width=0.05\linewidth]{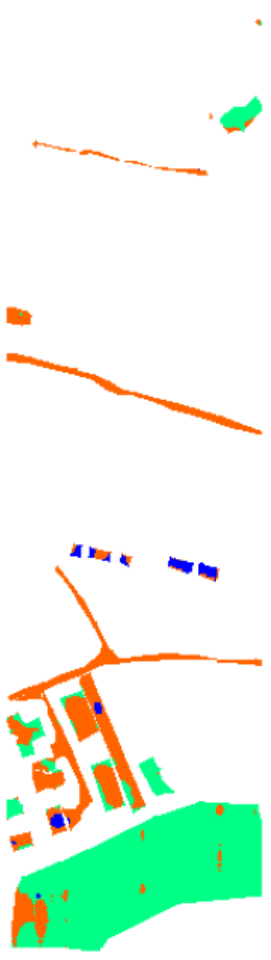}
		\hfill
		\includegraphics[width=0.05\linewidth]{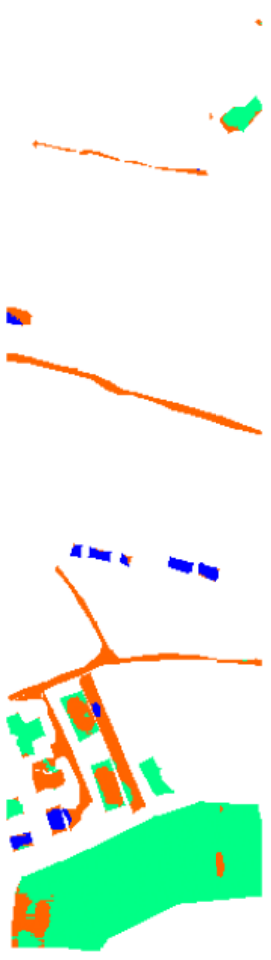}
		\hfill
		\includegraphics[width=0.05\linewidth]{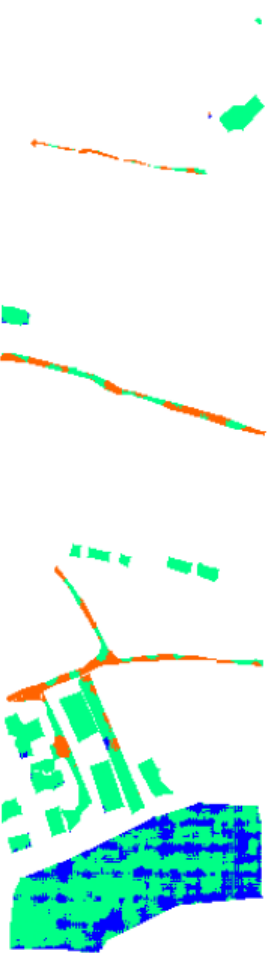}
		\hfill
		\includegraphics[width=0.05\linewidth]{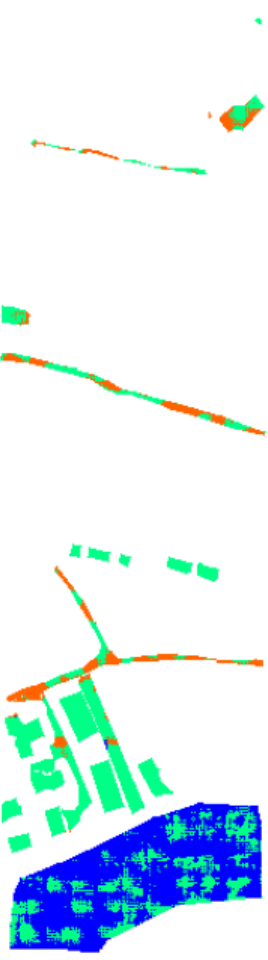}
		\hfill
		\includegraphics[width=0.05\linewidth]{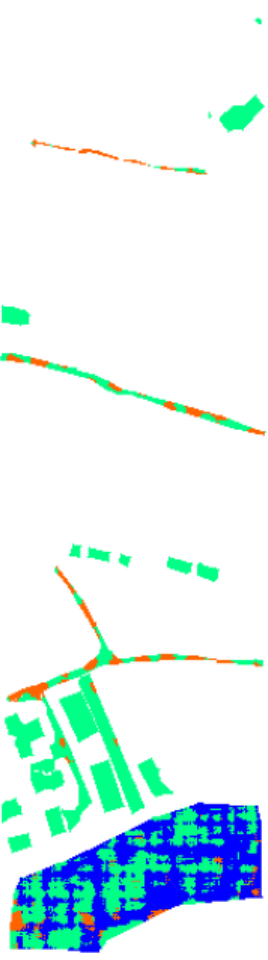}
		\hfill
		\includegraphics[width=0.05\linewidth]{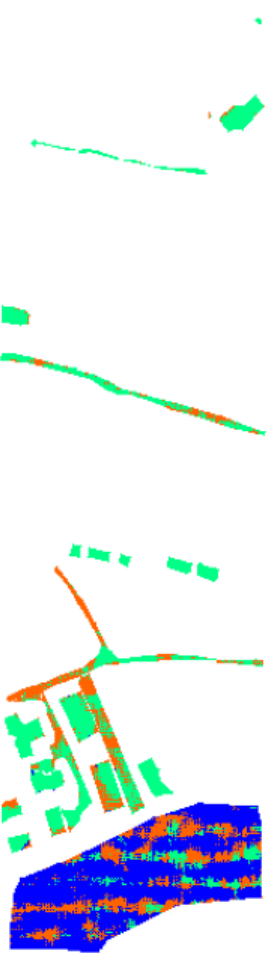}
		\hfill
		\includegraphics[width=0.05\linewidth]{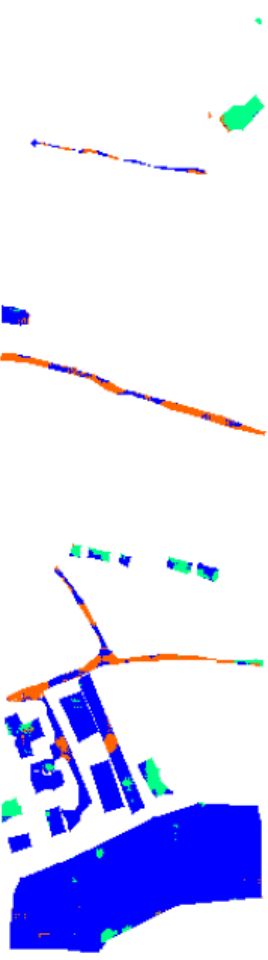}
		\hfill
		\includegraphics[width=0.05\linewidth]{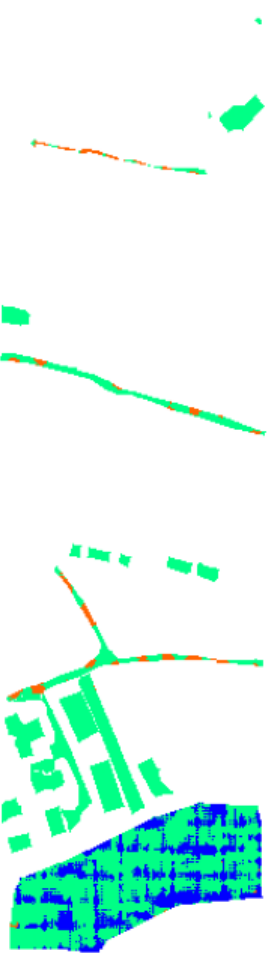}
		\hfill
		\includegraphics[width=0.05\linewidth]{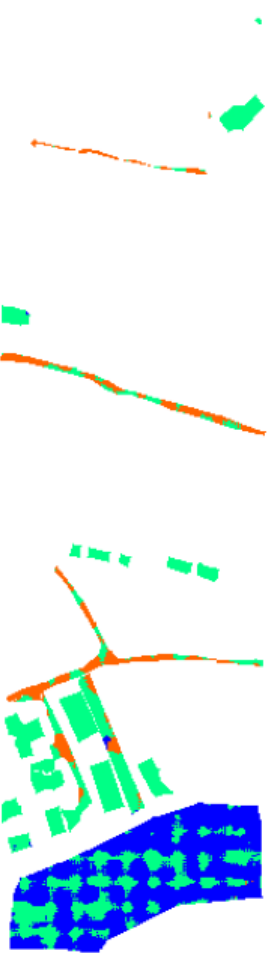}
		\hfill
		\includegraphics[width=0.05\linewidth]{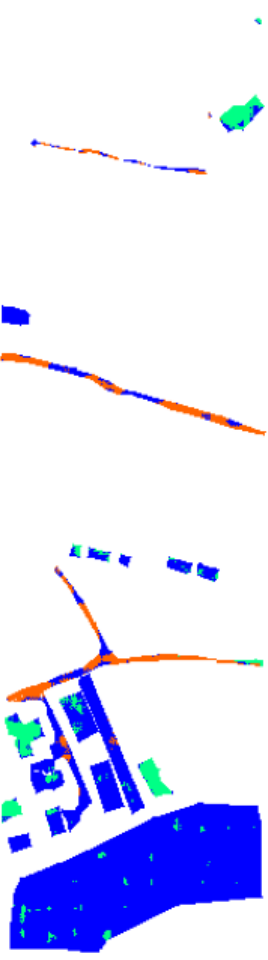}
		\hfill
		\includegraphics[width=0.05\linewidth]{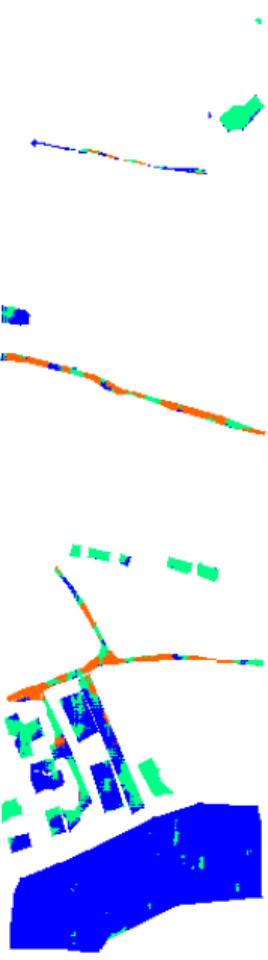}
		
		\makebox[0.05\linewidth][c]{(a)}
		\hfill
		\makebox[0.05\linewidth][c]{(b)}
		\hfill
		\makebox[0.05\linewidth][c]{(c)}
		\hfill
		\makebox[0.05\linewidth][c]{(d)}
		\hfill
		\makebox[0.05\linewidth][c]{(e)}
		\hfill
		\makebox[0.05\linewidth][c]{(f)}
		\hfill
		\makebox[0.05\linewidth][c]{(g)}
		\hfill
		\makebox[0.05\linewidth][c]{(h)}
		\hfill
		\makebox[0.05\linewidth][c]{(i)}
		\hfill
		\makebox[0.05\linewidth][c]{(j)}
		\hfill
		\makebox[0.05\linewidth][c]{(k)}
		\hfill
		\makebox[0.05\linewidth][c]{(l)}
		\hfill
		\makebox[0.05\linewidth][c]{(m)}
		\hfill
		\makebox[0.05\linewidth][c]{(n)}
		\hfill
		\makebox[0.05\linewidth][c]{(o)}
		\hfill
		\makebox[0.05\linewidth][c]{(p)}
		
		\caption{Classification maps (Blue/orange/green regions: Trees/Roads/Buildings) on the HU$\rightarrow$TR dataset combination. (a) HSI. (b) LiDAR image. (c) Ground truth. (d) MFT. (e) MsFE. (f) CMFAEN. (g) DKDMN. (h) SDENet. (i) LLURNet. (j) FDGNet. (k) TFTNet. (l) ADNet. (m) ISDGS. (n) EHSNet. (o) LDGNet. (p) FVMGN.} \label{HT}
	\end{figure*}
	
	\begin{table*}[t]\centering
		\caption{Classification results of different methods on the HU$\rightarrow$MU dataset combination. \label{Coms4}}
		\resizebox{\linewidth}{11.5mm}{
			\begin{tabular}{cccccccccccccc}
				\toprule
				Class No. & MFT & MsFE-IFN & CMFAEN & DKDMN & SDENet & LLURNet & FDGNet & TFTNet & ADNet & ISDGS & EHSNet & LDGNet & FVMGN \\ \midrule
				1 & $0.230_{\pm{0.18}}$ & $71.65_{\pm{24.2}}$ & $0.260_{\pm{0.47}}$ & $45.07_{\pm{39.3}}$ & $83.00_{\pm{11.1}}$ & $55.87_{\pm{24.0}}$ & $7.640_{\pm{22.9}}$ & $38.55_{\pm{5.58}}$ & $62.96_{\pm{33.61}}$ & $41.11_{\pm{31.2}}$ & $\mathbf{96.02}_{\pm{8.43}}$ & $\underline{87.06}_{\pm{4.47}}$ & $88.49_{\pm{5.55}}$ \\
				2 & $\underline{94.80}_{\pm{1.47}}$ & $0.220_{\pm{0.62}}$ & $\mathbf{99.28}_{\pm{1.55}}$ & $59.92_{\pm{37.6}}$ & $22.35_{\pm{26.6}}$ & $10.15_{\pm{18.4}}$ & $20.22_{\pm{31.1}}$ & $17.66_{\pm{15.2}}$ & $5.36_{\pm{10.50}}$ & $30.56_{\pm{19.7}}$ & $7.510_{\pm{21.0}}$ & $80.23_{\pm{12.6}}$ & $77.32_{\pm{13.4}}$ \\
				3 & $0.830_{\pm{0.77}}$ & $12.76_{\pm{15.3}}$ & $0.350_{\pm{0.85}}$ & $36.23_{\pm{42.1}}$ & $27.74_{\pm{18.6}}$ & $61.97_{\pm{25.3}}$ & $\mathbf{78.89}_{\pm{34.1}}$ & $\underline{73.11}_{\pm{14.0}}$ & $57.75_{\pm{31.3}}$ & $54.02_{\pm{20.1}}$ & $9.480_{\pm{24.1}}$ & $55.84_{\pm{15.8}}$ & $64.74_{\pm{24.0}}$ \\ \midrule
				OA & $17.82_{\pm{0.32}}$ & $48.38_{\pm{13.9}}$ & $18.60_{\pm{0.14}}$ & $46.31_{\pm{14.1}}$ & $62.25_{\pm{2.61}}$ & $48.47_{\pm{13.1}}$ & $22.26_{\pm{13.7}}$ & $40.65_{\pm{2.98}}$ & $51.41_{\pm{18.0}}$ & $41.39_{\pm{16.9}}$ & $64.73_{\pm{3.46}}$ & $\underline{80.41}_{\pm{3.02}}$ & $\mathbf{82.33}_{\pm{3.75}}$ \\
				AA & $31.95_{\pm{0.62}}$ & $28.21_{\pm{5.86}}$ & $33.29_{\pm{0.14}}$ & $47.07_{\pm{13.0}}$ & $44.36_{\pm{5.13}}$ & $42.66_{\pm{7.26}}$ & $35.58_{\pm{5.53}}$ & $43.11_{\pm{2.49}}$ & $42.03_{\pm{8.23}}$ & $41.90_{\pm{5.58}}$ & $37.67_{\pm{12.4}}$ & $\underline{74.38}_{\pm{4.01}}$ & $\mathbf{76.85}_{\pm{7.05}}$ \\
				Kappa & $4.130_{\pm{1.47}}$ & $11.79_{\pm{11.8}}$ & $0.070_{\pm{0.15}}$ & $10.37_{\pm{14.8}}$ & $24.74_{\pm{6.89}}$ & $17.80_{\pm{11.0}}$ & $3.610_{\pm{9.93}}$ & $14.48_{\pm{2.40}}$ & $16.26_{\pm{16.3}}$ & $14.53_{\pm{11.0}}$ & $5.610_{\pm{16.14}}$ & $\underline{64.14}_{\pm{4.86}}$ & $\mathbf{66.44}_{\pm{7.54}}$ \\ \bottomrule
		\end{tabular}}
	\end{table*}
	
	\begin{figure*}[t]
		\centering
		\rmfamily
		\includegraphics[width=0.05\linewidth]{MUUFL_HSI}
		\hfill
		\includegraphics[width=0.05\linewidth]{MUUFL_LiDAR}
		\hfill
		\includegraphics[width=0.05\linewidth]{MUUFL_GT}
		\hfill
		\includegraphics[width=0.05\linewidth]{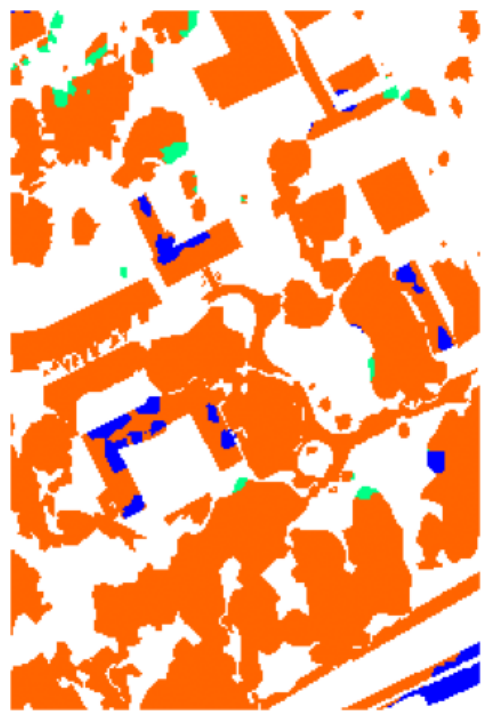}
		\hfill
		\includegraphics[width=0.05\linewidth]{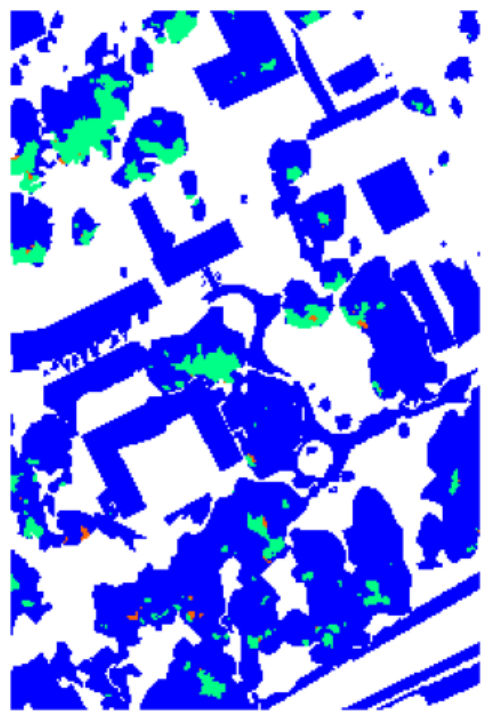}
		\hfill
		\includegraphics[width=0.05\linewidth]{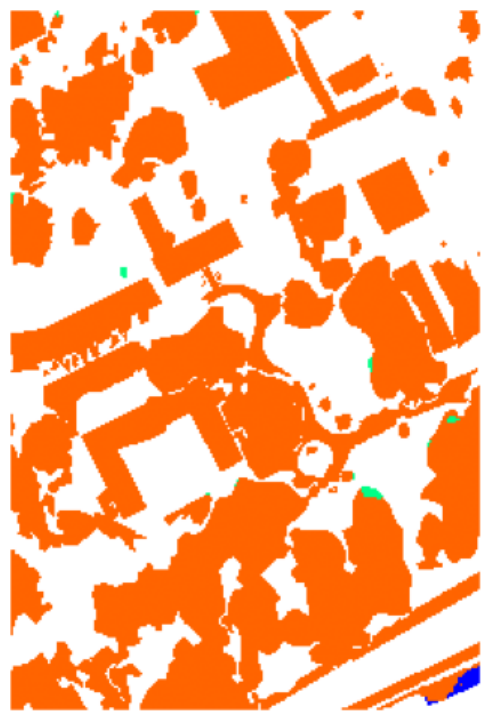}
		\hfill
		\includegraphics[width=0.05\linewidth]{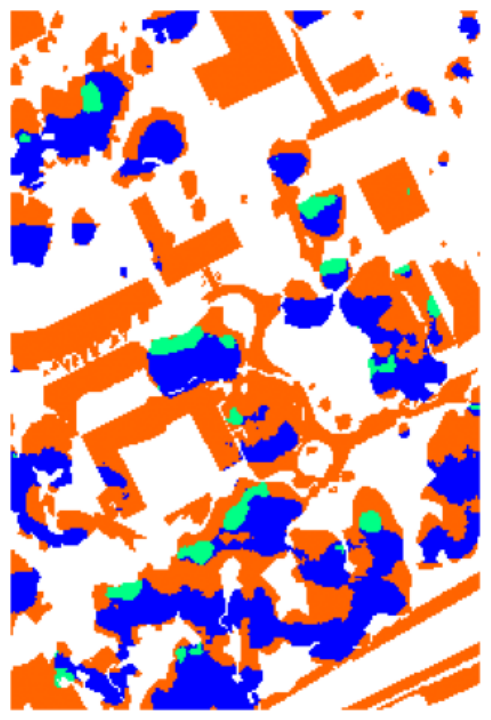}
		\hfill
		\includegraphics[width=0.05\linewidth]{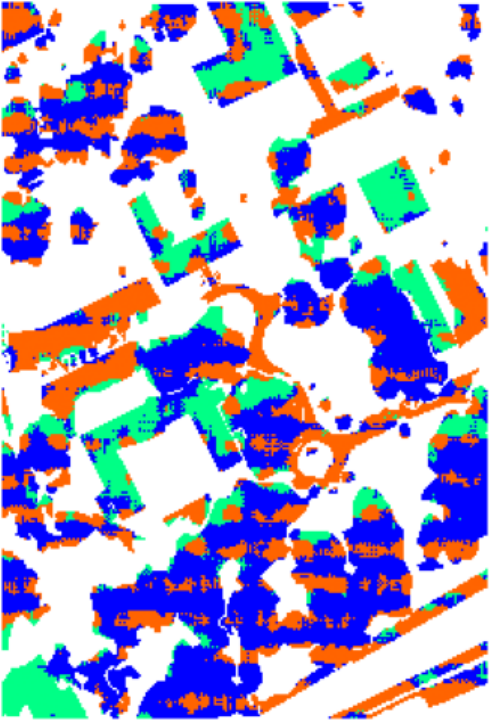}
		\hfill
		\includegraphics[width=0.05\linewidth]{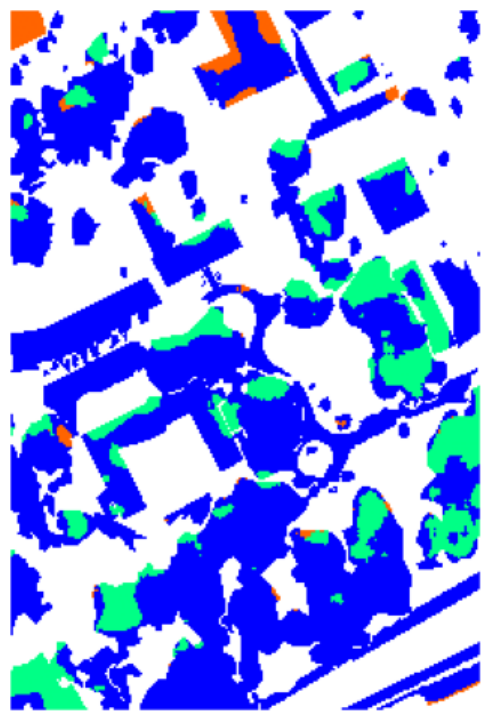}
		\hfill
		\includegraphics[width=0.05\linewidth]{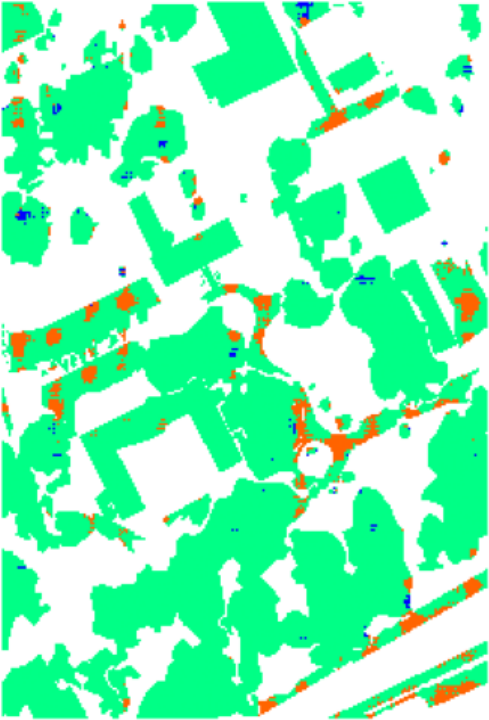}
		\hfill
		\includegraphics[width=0.05\linewidth]{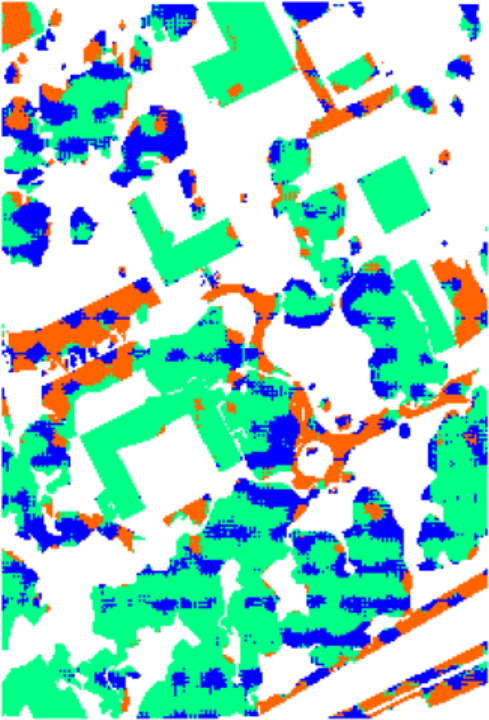}
		\hfill
		\includegraphics[width=0.05\linewidth]{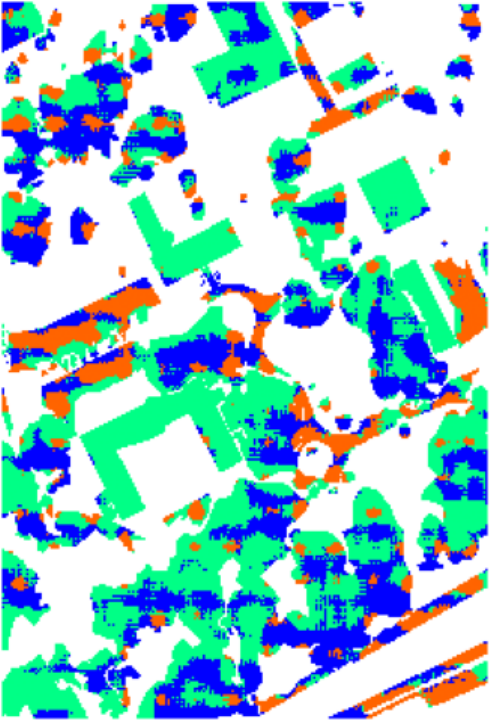}
		\hfill
		\includegraphics[width=0.05\linewidth]{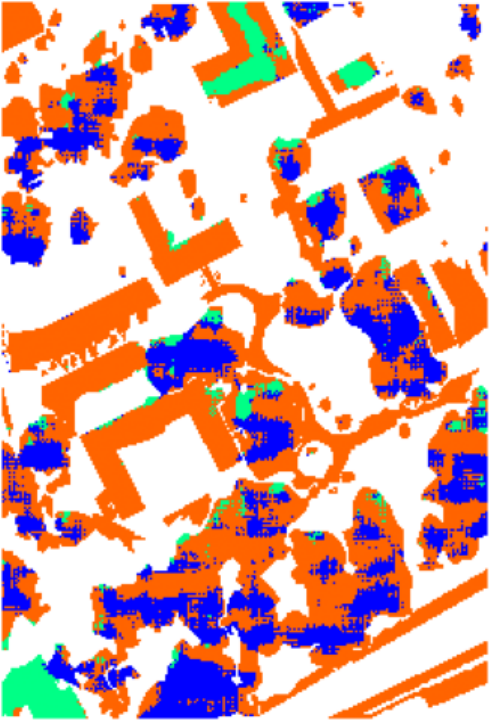}
		\hfill
		\includegraphics[width=0.05\linewidth]{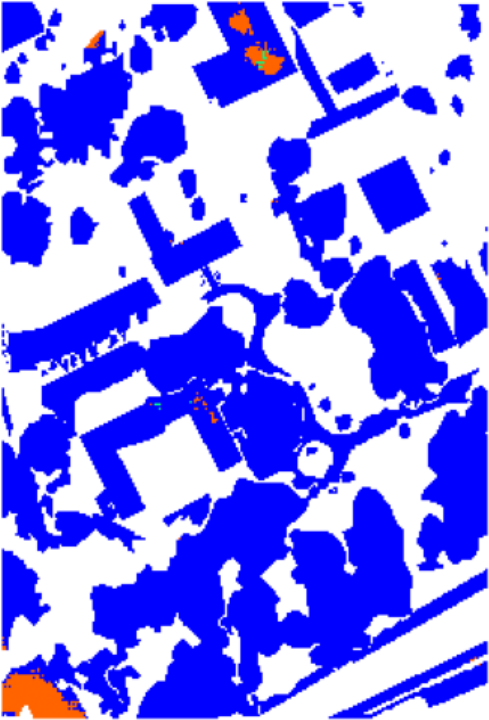}
		\hfill
		\includegraphics[width=0.05\linewidth]{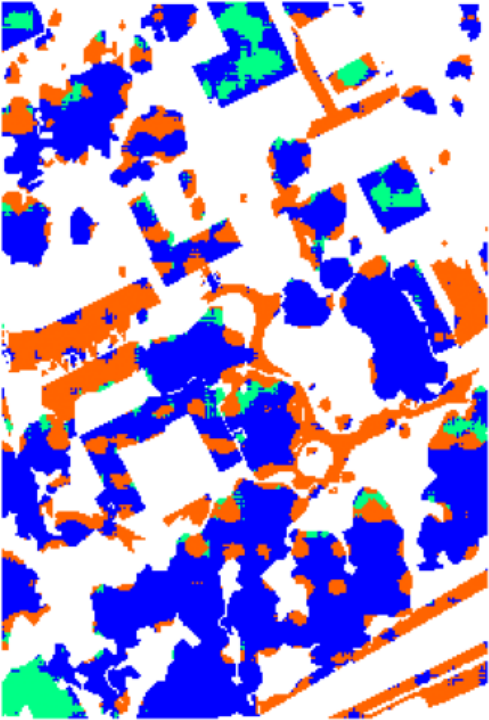}
		\hfill
		\includegraphics[width=0.05\linewidth]{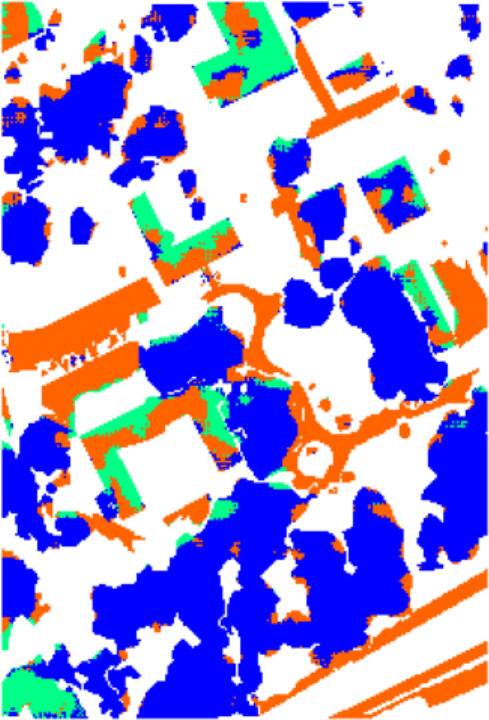}
		
		\makebox[0.05\linewidth][c]{(a)}
		\hfill
		\makebox[0.05\linewidth][c]{(b)}
		\hfill
		\makebox[0.05\linewidth][c]{(c)}
		\hfill
		\makebox[0.05\linewidth][c]{(d)}
		\hfill
		\makebox[0.05\linewidth][c]{(e)}
		\hfill
		\makebox[0.05\linewidth][c]{(f)}
		\hfill
		\makebox[0.05\linewidth][c]{(g)}
		\hfill
		\makebox[0.05\linewidth][c]{(h)}
		\hfill
		\makebox[0.05\linewidth][c]{(i)}
		\hfill
		\makebox[0.05\linewidth][c]{(j)}
		\hfill
		\makebox[0.05\linewidth][c]{(k)}
		\hfill
		\makebox[0.05\linewidth][c]{(l)}
		\hfill
		\makebox[0.05\linewidth][c]{(m)}
		\hfill
		\makebox[0.05\linewidth][c]{(n)}
		\hfill
		\makebox[0.05\linewidth][c]{(o)}
		\hfill
		\makebox[0.05\linewidth][c]{(p)}
		
		\caption{Classification maps (Blue/orange/green regions: Trees/Roads/Buildings) on the HU$\rightarrow$MU dataset combination. (a) HSI. (b) LiDAR image. (c) Ground truth. (d) MFT. (e) MsFE. (f) CFEN. (g) DKDN. (h) SDEN. (i) LLURN. (j) FDGN. (k) TFTN. (l) ADNet. (m) ISDGS. (n) EHSN. (o) LDGN. (p) FVMGN.} \label{HM}
	\end{figure*}
	
	\begin{table*}[t]\centering
		\caption{Classification results of different methods on the TR$\rightarrow$HU dataset combination. \label{Coms5}}
		\resizebox{\linewidth}{11.5mm}{
			\begin{tabular}{cccccccccccccc}
				\toprule
				Class No. & MFT & MsFE-IFN & CMFAEN & DKDMN & SDENet & LLURNet & FDGNet & TFTNet & ADNet & ISDGS & EHSNet & LDGNet & FVMGN \\ \midrule
				1 & $42.76_{\pm{29.9}}$ & $12.68_{\pm{15.8}}$ & $7.340_{\pm{4.17}}$ & $44.55_{\pm{22.7}}$ & $92.31_{\pm{5.92}}$ & $93.55_{\pm{2.91}}$ & $94.54_{\pm{3.24}}$ & $\mathbf{99.07}_{\pm{0.95}}$ & $95.16_{\pm{3.20}}$ & $\underline{98.97}_{\pm{1.14}}$ & $40.70_{\pm{30.5}}$ & $79.07_{\pm{16.1}}$ & $90.04_{\pm{6.46}}$ \\
				2 & $4.310_{\pm{12.9}}$ & $3.840_{\pm{5.31}}$ & $---$ & $26.83_{\pm{18.1}}$ & $45.68_{\pm{11.8}}$ & $50.15_{\pm{9.06}}$ & $49.48_{\pm{8.68}}$ & $\underline{80.18}_{\pm{4.97}}$ & $67.40_{\pm{4.15}}$ & $74.60_{\pm{4.12}}$ & $51.52_{\pm{31.6}}$ & $62.30_{\pm{20.0}}$ & $\mathbf{98.72}_{\pm{1.85}}$ \\
				3 & $36.20_{\pm{18.7}}$ & $15.02_{\pm{10.3}}$ & $50.88_{\pm{5.04}}$ & $9.990_{\pm{9.24}}$ & $32.42_{\pm{6.37}}$ & $39.63_{\pm{4.09}}$ & $33.79_{\pm{5.36}}$ & $61.39_{\pm{4.72}}$ & $52.95_{\pm{5.75}}$ & $53.01_{\pm{7.63}}$ & $40.54_{\pm{25.8}}$ & $\underline{63.52}_{\pm{12.8}}$ & $\mathbf{82.64}_{\pm{6.30}}$ \\ \midrule
				OA & $28.94_{\pm{9.62}}$ & $11.11_{\pm{5.29}}$ & $23.36_{\pm{2.53}}$ & $25.06_{\pm{5.91}}$ & $53.84_{\pm{2.39}}$ & $58.51_{\pm{1.57}}$ & $56.17_{\pm{2.02}}$ & $\underline{77.89}_{\pm{1.82}}$ & $69.53_{\pm{3.44}}$ & $72.75_{\pm{2.93}}$ & $43.76_{\pm{8.11}}$ & $67.74_{\pm{8.77}}$ & $\mathbf{89.46}_{\pm{3.35}}$ \\
				AA & $27.76_{\pm{9.43}}$ & $10.51_{\pm{5.24}}$ & $19.41_{\pm{2.27}}$ & $27.12_{\pm{6.86}}$ & $56.80_{\pm{2.74}}$ & $61.11_{\pm{2.24}}$ & $59.27_{\pm{2.03}}$ & $\underline{80.21}_{\pm{1.70}}$ & $71.84_{\pm{3.16}}$ & $75.53_{\pm{2.34}}$ & $44.25_{\pm{8.73}}$ & $68.30_{\pm{9.61}}$ & $\mathbf{90.47}_{\pm{3.09}}$ \\
				Kappa & $6.450_{\pm{13.65}}$ & $32.19_{\pm{8.27}}$ & $20.27_{\pm{3.65}}$ & $9.390_{\pm{9.50}}$ & $31.64_{\pm{3.69}}$ & $38.26_{\pm{2.65}}$ & $35.32_{\pm{2.82}}$ & $\underline{67.10}_{\pm{2.62}}$ & $54.61_{\pm{4.88}}$ & $59.62_{\pm{4.12}}$ & $16.59_{\pm{12.0}}$ & $51.32_{\pm{13.5}}$ & $\mathbf{84.12}_{\pm{4.96}}$ \\ \bottomrule
		\end{tabular}}
	\end{table*}
	
	\begin{figure*}[t]
		\centering
		\rmfamily
		\includegraphics[width=0.05\linewidth]{HU13_HSI}
		\hfill
		\includegraphics[width=0.05\linewidth]{HU13_LiDAR}
		\hfill
		\includegraphics[width=0.05\linewidth]{HU13_gt}
		\hfill
		\includegraphics[width=0.05\linewidth]{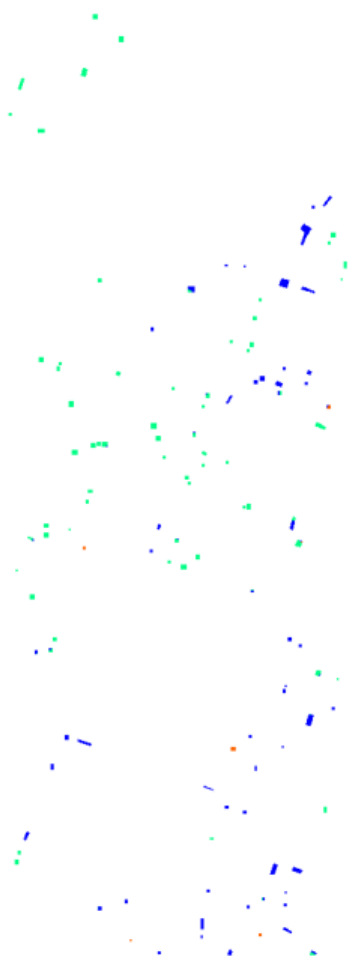}
		\hfill
		\includegraphics[width=0.05\linewidth]{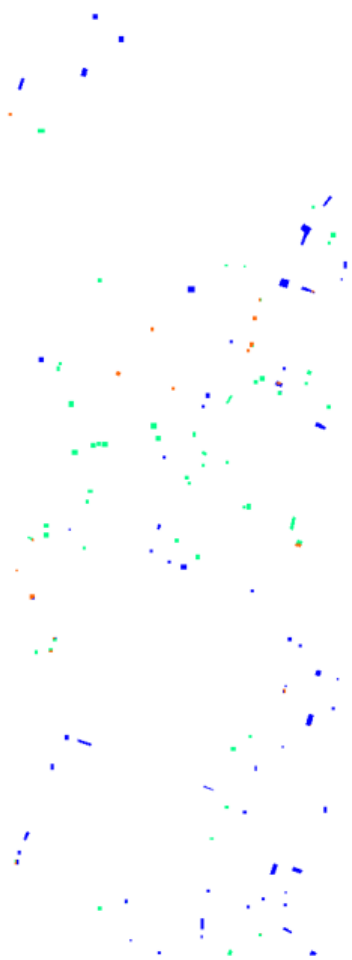}
		\hfill
		\includegraphics[width=0.05\linewidth]{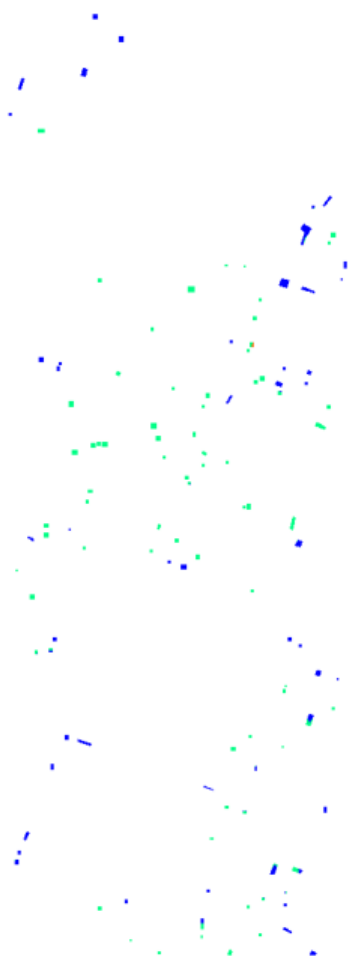}
		\hfill
		\includegraphics[width=0.05\linewidth]{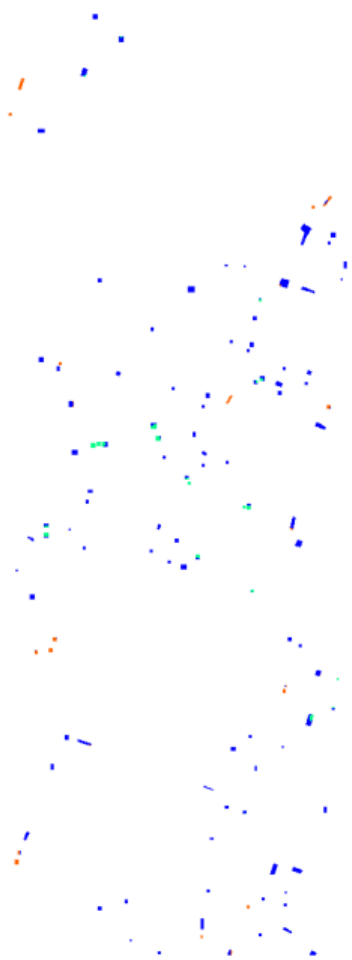}
		\hfill
		\includegraphics[width=0.05\linewidth]{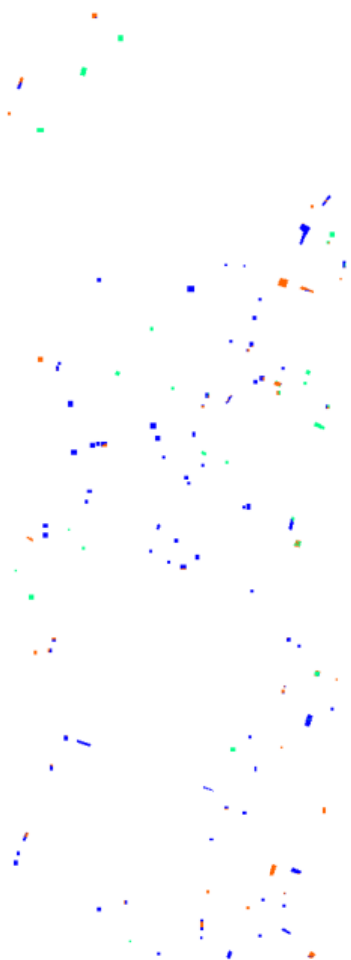}
		\hfill
		\includegraphics[width=0.05\linewidth]{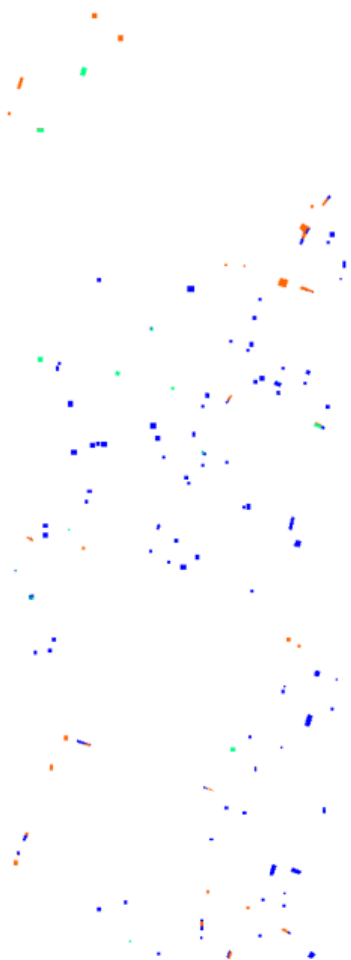}
		\hfill
		\includegraphics[width=0.05\linewidth]{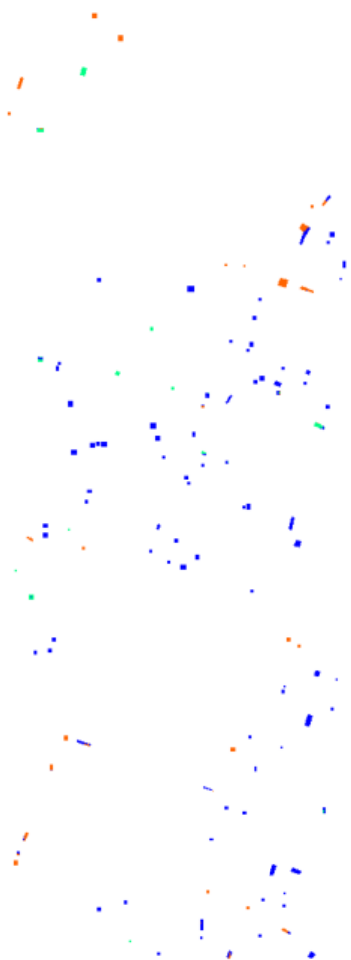}
		\hfill
		\includegraphics[width=0.05\linewidth]{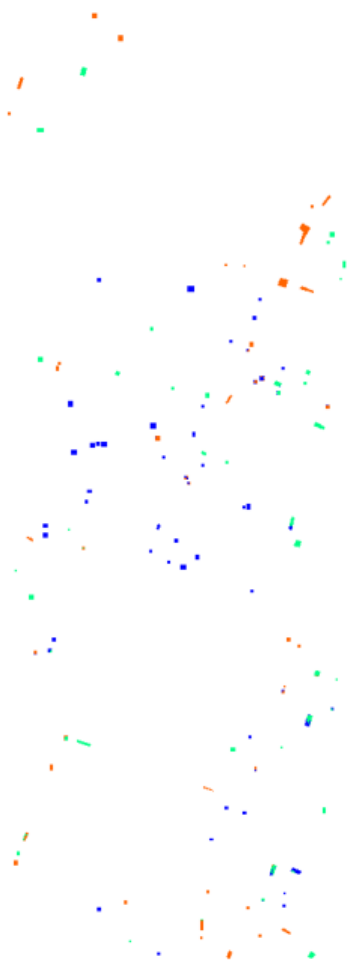}
		\hfill
		\includegraphics[width=0.05\linewidth]{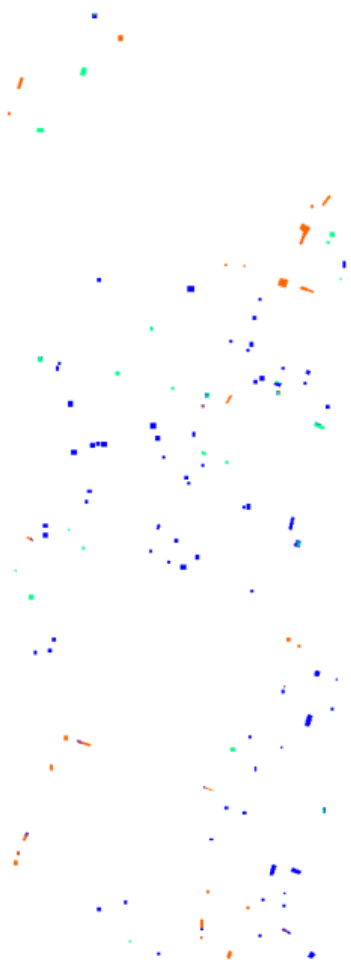}
		\hfill
		\includegraphics[width=0.05\linewidth]{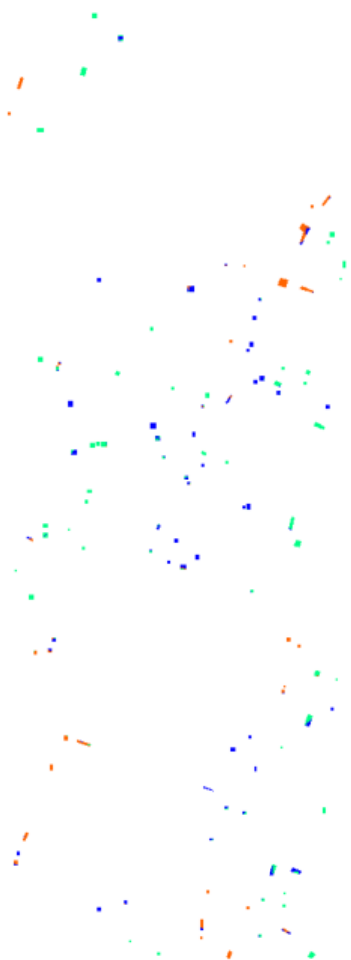}
		\hfill
		\includegraphics[width=0.05\linewidth]{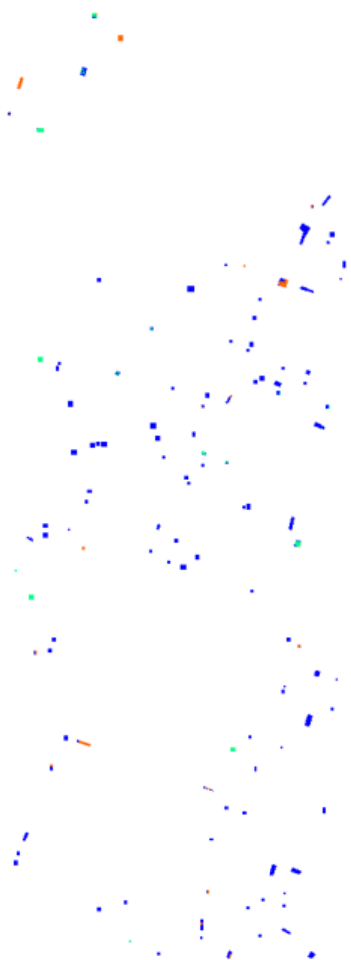}
		\hfill
		\includegraphics[width=0.05\linewidth]{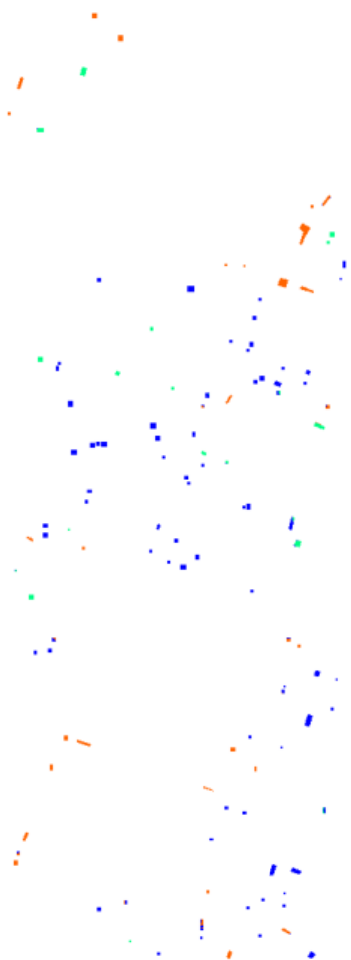}
		\hfill
		\includegraphics[width=0.05\linewidth]{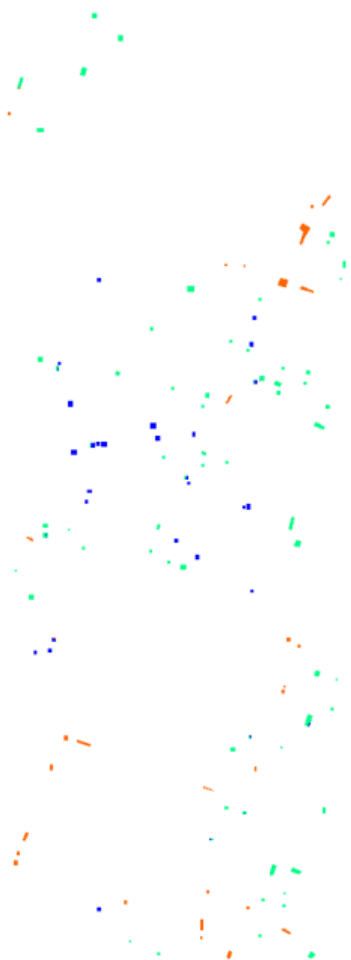}
		
		\makebox[0.05\linewidth][c]{(a)}
		\hfill
		\makebox[0.05\linewidth][c]{(b)}
		\hfill
		\makebox[0.05\linewidth][c]{(c)}
		\hfill
		\makebox[0.05\linewidth][c]{(d)}
		\hfill
		\makebox[0.05\linewidth][c]{(e)}
		\hfill
		\makebox[0.05\linewidth][c]{(f)}
		\hfill
		\makebox[0.05\linewidth][c]{(g)}
		\hfill
		\makebox[0.05\linewidth][c]{(h)}
		\hfill
		\makebox[0.05\linewidth][c]{(i)}
		\hfill
		\makebox[0.05\linewidth][c]{(j)}
		\hfill
		\makebox[0.05\linewidth][c]{(k)}
		\hfill
		\makebox[0.05\linewidth][c]{(l)}
		\hfill
		\makebox[0.05\linewidth][c]{(m)}
		\hfill
		\makebox[0.05\linewidth][c]{(n)}
		\hfill
		\makebox[0.05\linewidth][c]{(o)}
		\hfill
		\makebox[0.05\linewidth][c]{(p)}
		
		\caption{Classification maps (Blue/orange/green regions: Trees/Roads/Buildings) on the TR$\rightarrow$HU dataset combination. (a) HSI. (b) LiDAR image. (c) Ground truth. (d) MFT. (e) MsFE. (f) CMFAEN. (g) DKDMN. (h) SDENet. (i) LLURNet. (j) FDGNet. (k) TFTNet. (l) ADNet. (m) ISDGS. (n) EHSNet. (o) LDGNet. (p) FVMGN.} \label{TH}
	\end{figure*}
	
	\begin{table*}[t]\centering
		\caption{Classification results of different methods on the TR$\rightarrow$MU dataset combination. \label{Coms6}}
		\resizebox{\linewidth}{11.5mm}{
			\begin{tabular}{cccccccccccccc}
				\toprule
				Class No. & MFT & MsFE-IFN & CMFAEN & DKDMN & SDENet & LLURNet & FDGNet & TFTNet & ADNet & ISDGS & EHSNet & LDGNet & FVMGN \\ \midrule
				1 & $91.01_{\pm{4.88}}$ & $80.22_{\pm{15.2}}$ & $81.63_{\pm{5.15}}$ & $48.92_{\pm{22.6}}$ & $95.26_{\pm{7.39}}$ & $92.95_{\pm{3.88}}$ & $\mathbf{98.26}_{\pm{1.94}}$ & $82.23_{\pm{7.78}}$ & $\underline{96.95}_{\pm{5.06}}$ & $92.43_{\pm{6.33}}$ & $88.97_{\pm{9.10}}$ & $88.63_{\pm{4.47}}$ & $93.37_{\pm{4.03}}$ \\
				2 & $1.360_{\pm{2.70}}$ & $47.29_{\pm{38.4}}$ & $---$ & $\mathbf{91.17}_{\pm{8.87}}$ & $5.570_{\pm{12.6}}$ & $7.500_{\pm{10.78}}$ & $1.650_{\pm{2.70}}$ & $32.60_{\pm{22.2}}$ & $5.87_{\pm{17.60}}$ & $16.73_{\pm{19.8}}$ & $62.04_{\pm{41.1}}$ & $63.95_{\pm{25.4}}$ & $\underline{90.59}_{\pm{6.43}}$ \\
				3 & $44.58_{\pm{14.3}}$ & $48.17_{\pm{31.3}}$ & $55.67_{\pm{14.2}}$ & $7.090_{\pm{11.7}}$ & $31.73_{\pm{28.9}}$ & $\underline{57.19}_{\pm{19.9}}$ & $18.53_{\pm{14.3}}$ & $25.08_{\pm{17.3}}$ & $31.19_{\pm{39.7}}$ & $26.81_{\pm{15.8}}$ & $30.99_{\pm{28.2}}$ & $34.78_{\pm{22.9}}$ & $\mathbf{76.86}_{\pm{9.68}}$ \\ \midrule
				OA & $66.50_{\pm{3.14}}$ & $68.66_{\pm{6.38}}$ & $62.09_{\pm{3.22}}$ & $49.60_{\pm{13.0}}$ & $67.72_{\pm{2.87}}$ & $70.98_{\pm{2.97}}$ & $66.64_{\pm{1.97}}$ & $63.20_{\pm{2.27}}$ & $68.77_{\pm{4.89}}$ & $67.11_{\pm{3.03}}$ & $73.99_{\pm{6.77}}$ & $\underline{74.42}_{\pm{6.06}}$ & $\mathbf{90.01}_{\pm{2.27}}$ \\
				AA & $45.65_{\pm{4.81}}$ & $58.56_{\pm{8.52}}$ & $45.77_{\pm{4.33}}$ & $49.06_{\pm{5.81}}$ & $44.19_{\pm{7.90}}$ & $52.54_{\pm{7.08}}$ & $39.48_{\pm{4.49}}$ & $46.64_{\pm{4.83}}$ & $44.67_{\pm{11.8}}$ & $45.32_{\pm{7.29}}$ & $60.67_{\pm{14.8}}$ & $\underline{62.45}_{\pm{11.2}}$ & $\mathbf{86.94}_{\pm{3.73}}$ \\
				Kappa & $22.86_{\pm{8.28}}$ & $43.64_{\pm{11.4}}$ & $24.63_{\pm{6.53}}$ & $24.98_{\pm{10.9}}$ & $23.66_{\pm{13.5}}$ & $37.12_{\pm{11.4}}$ & $14.62_{\pm{11.0}}$ & $23.65_{\pm{7.17}}$ & $22.55_{\pm{22.3}}$ & $24.08_{\pm{13.8}}$ & $44.99_{\pm{23.9}}$ & $\underline{53.05}_{\pm{11.8}}$ & $\mathbf{81.27}_{\pm{3.94}}$ \\ \bottomrule
		\end{tabular}}
	\end{table*}
	
	\begin{figure*}[t]
		\centering
		\rmfamily
		\includegraphics[width=0.05\linewidth]{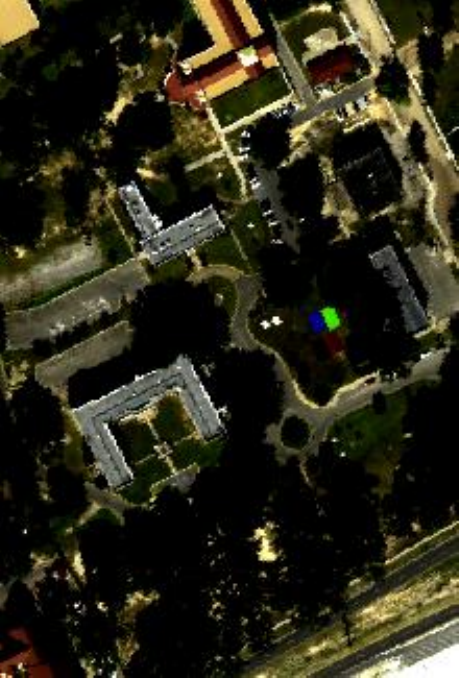}
		\hfill
		\includegraphics[width=0.05\linewidth]{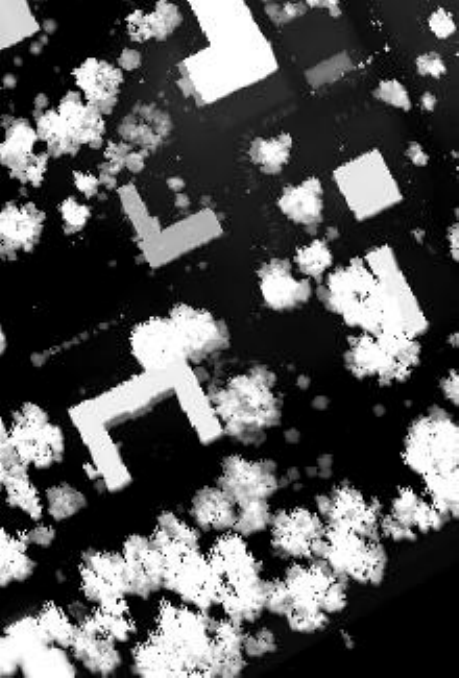}
		\hfill
		\includegraphics[width=0.05\linewidth]{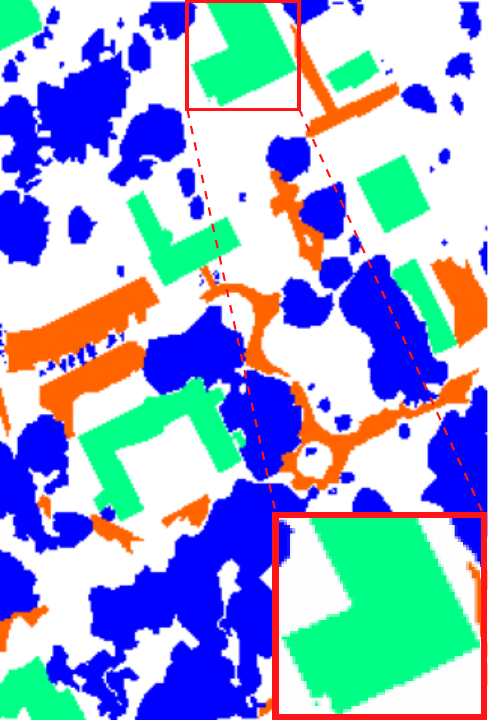}
		\hfill
		\includegraphics[width=0.05\linewidth]{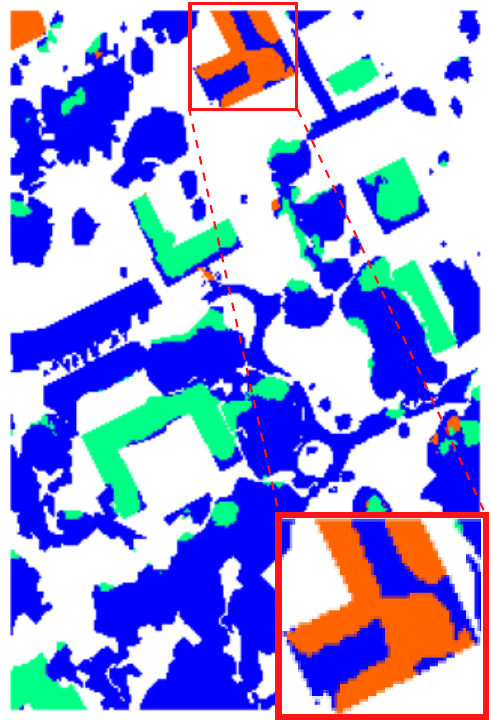}
		\hfill
		\includegraphics[width=0.05\linewidth]{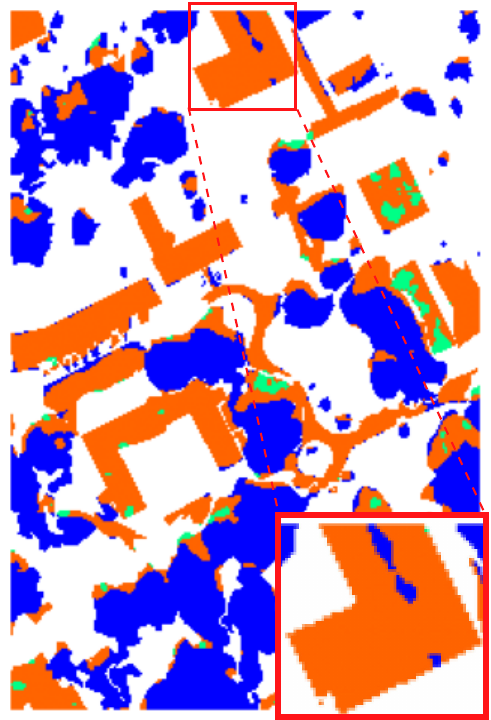}
		\hfill
		\includegraphics[width=0.05\linewidth]{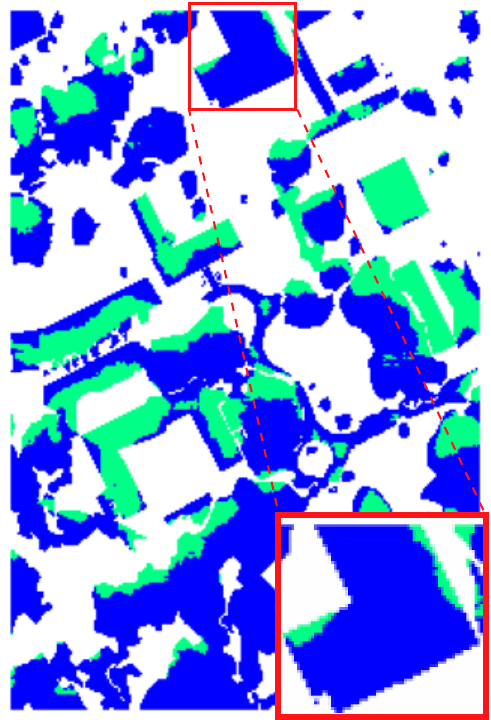}
		\hfill
		\includegraphics[width=0.05\linewidth]{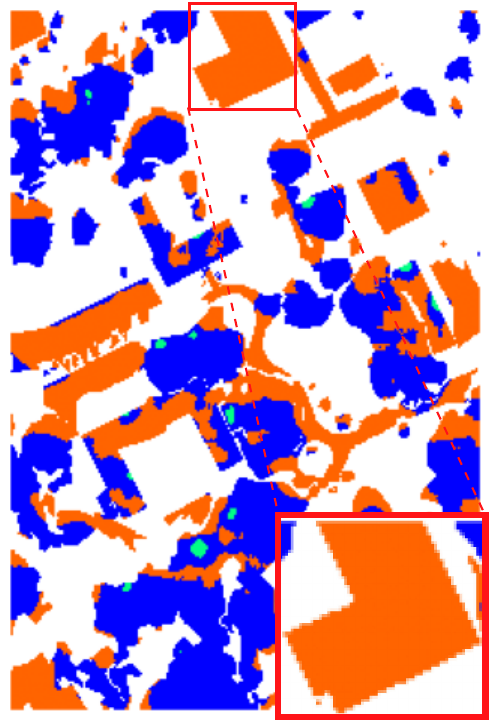}
		\hfill
		\includegraphics[width=0.05\linewidth]{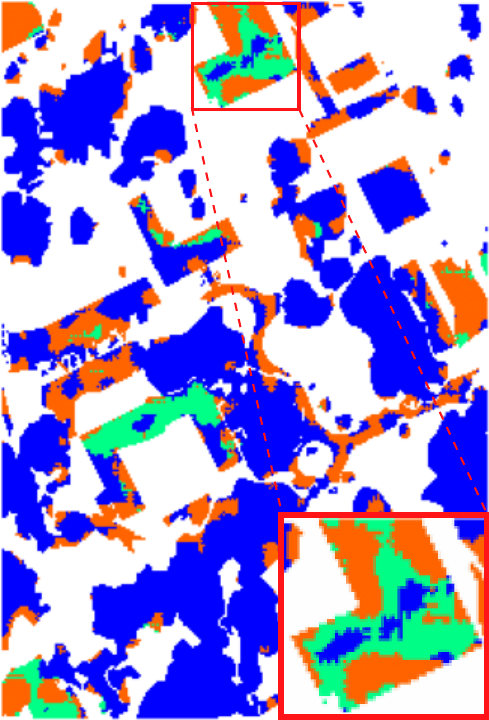}
		\hfill
		\includegraphics[width=0.05\linewidth]{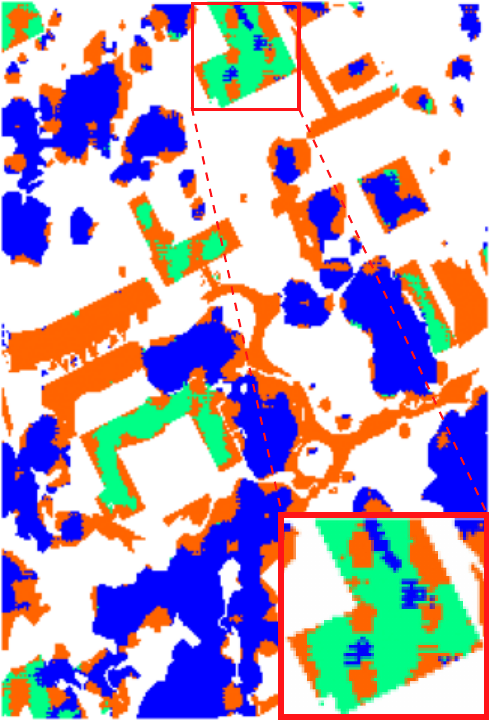}
		\hfill
		\includegraphics[width=0.05\linewidth]{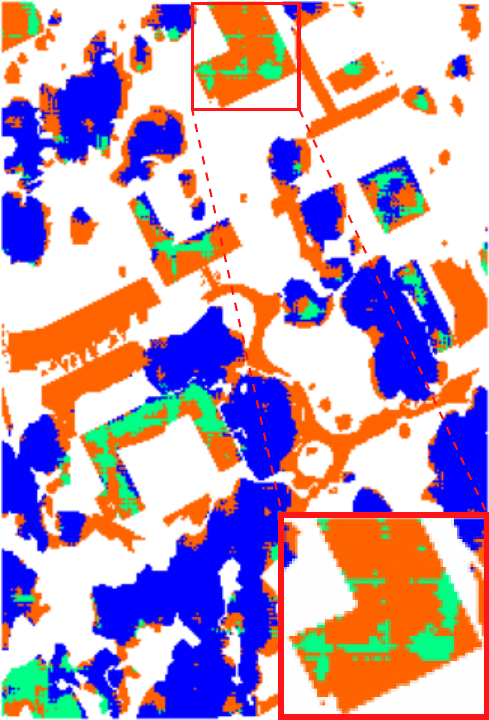}
		\hfill
		\includegraphics[width=0.05\linewidth]{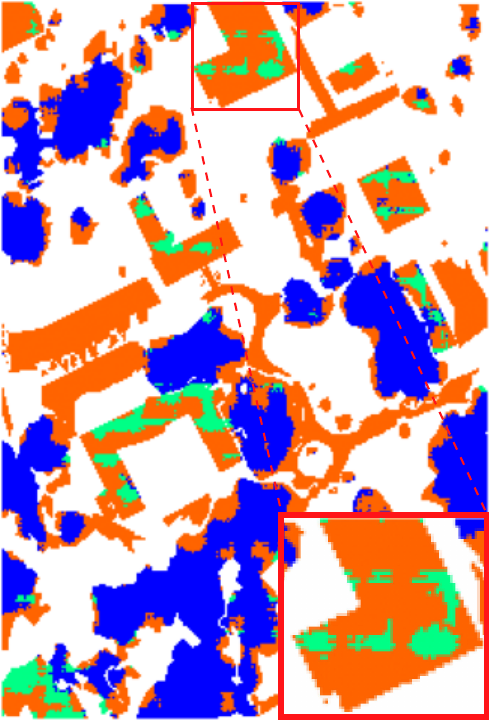}
		\hfill
		\includegraphics[width=0.05\linewidth]{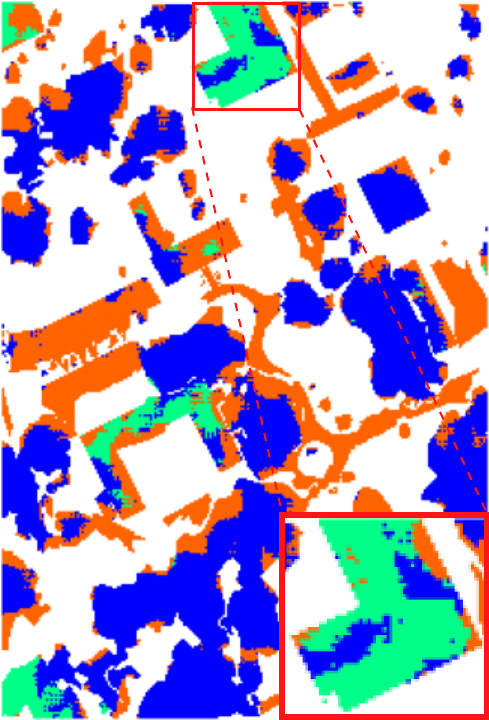}
		\hfill
		\includegraphics[width=0.05\linewidth]{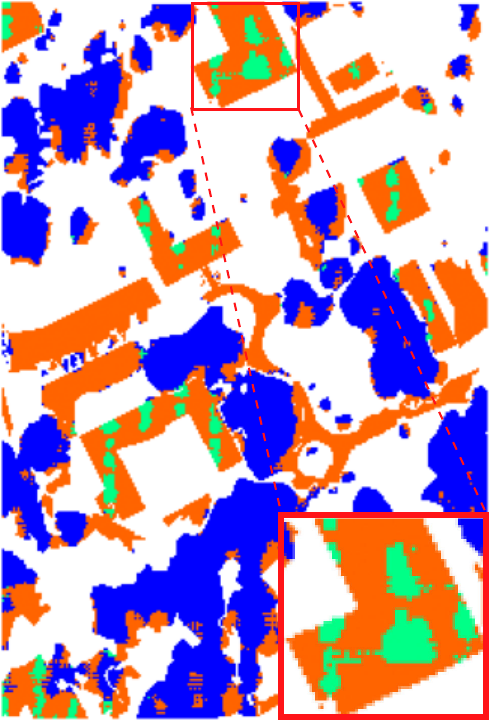}
		\hfill
		\includegraphics[width=0.05\linewidth]{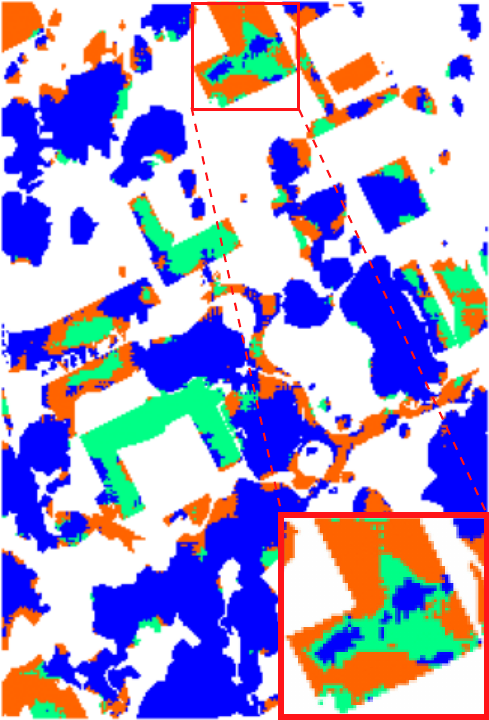}
		\hfill
		\includegraphics[width=0.05\linewidth]{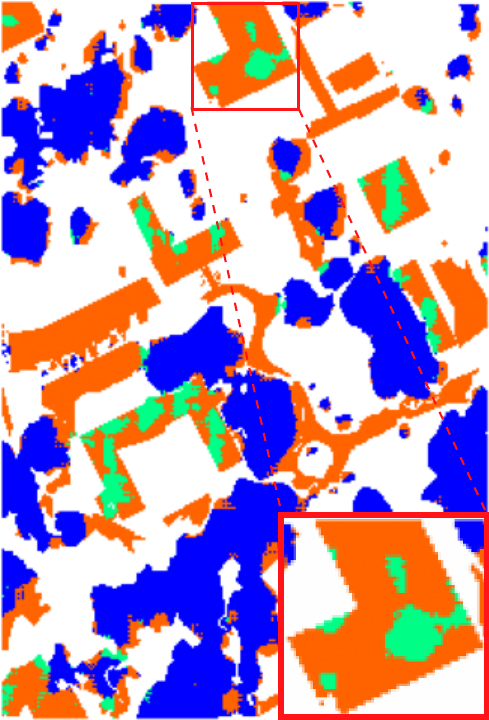}
		\hfill
		\includegraphics[width=0.05\linewidth]{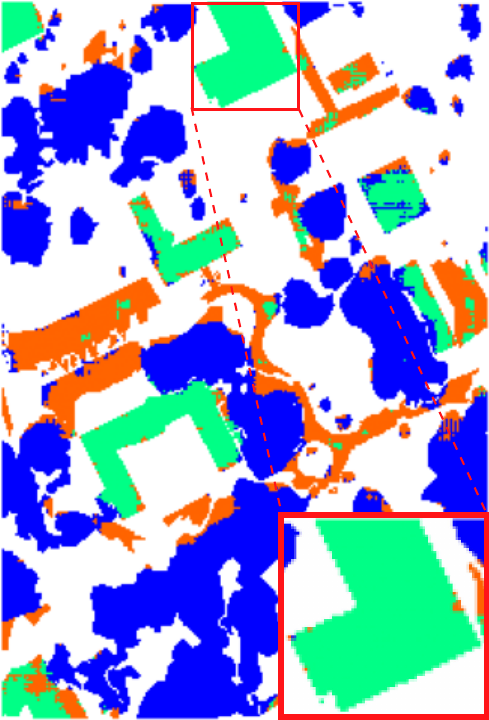}
		
		\makebox[0.05\linewidth][c]{(a)}
		\hfill
		\makebox[0.05\linewidth][c]{(b)}
		\hfill
		\makebox[0.05\linewidth][c]{(c)}
		\hfill
		\makebox[0.05\linewidth][c]{(d)}
		\hfill
		\makebox[0.05\linewidth][c]{(e)}
		\hfill
		\makebox[0.05\linewidth][c]{(f)}
		\hfill
		\makebox[0.05\linewidth][c]{(g)}
		\hfill
		\makebox[0.05\linewidth][c]{(h)}
		\hfill
		\makebox[0.05\linewidth][c]{(i)}
		\hfill
		\makebox[0.05\linewidth][c]{(j)}
		\hfill
		\makebox[0.05\linewidth][c]{(k)}
		\hfill
		\makebox[0.05\linewidth][c]{(l)}
		\hfill
		\makebox[0.05\linewidth][c]{(m)}
		\hfill
		\makebox[0.05\linewidth][c]{(n)}
		\hfill
		\makebox[0.05\linewidth][c]{(o)}
		\hfill
		\makebox[0.05\linewidth][c]{(p)}
		
		\caption{Classification maps (Blue/orange/green regions: Trees/Roads/Buildings) on the TR$\rightarrow$MU dataset combination. (a) HSI. (b) LiDAR image. (c) Ground truth. (d) MFT. (e) MsFE. (f) CFEN. (g) DKDN. (h) SDEN. (i) LLURN. (j) FDGN. (k) TFTN. (l) ADNet. (m) ISDGS. (n) EHSN. (o) LDGN. (p) FVMGN.} \label{TM}
	\end{figure*}
	
	\begin{table*}[t]\centering
		\caption{Computational complexity of different methods on six dataset combinations. \label{time}}
		\resizebox{\linewidth}{32mm}{
			\begin{tabular}{cccccccccccccccc}
				\toprule
				SD$\rightarrow$TD & Metric & MFT & MsFE-IFN & CMFAEN & DKDMN & SDENet & LLURNet & FDGNet & TFTNet & ADNet & ISDGS & EHSNet & LDGNet & FVMGN-NW & FVMGN   \\ \midrule
				\multirow{4}{*}{MU$\rightarrow$TR} & FLOPs (G) & $0.44$ & $66.23$ & $3.09$ & $6.26$ & $1.32$ & $0.33$ & $1.32$ & $2.67$ & $0.33$ & $1.32$ & $8.62$ & $7.14$ & $2.81$ & $2.84$   \\
				& Params (M) & $0.24$ & $\mathbf{0.02}$ & $0.19$ & $0.85$ & $0.44$ & $0.44$ & $0.44$ & $0.55$ & $0.44$ & $0.44$ & $0.34$ & $0.55$ & $0.65$ & $0.66$   \\
				& Training Time (s) & $345.16$ & $171.79$ & $475.24$ & $402.73$ & $6334.63$ & $1411.16$ & $3387.66$ & $6865.30$ & $1467.86$ & $1832.94$ & $15426.57$ & $19725.71$ & $5583.62$ & $5650.54$ \\
				& Test Time (s) & $6.06$ & $32.13$ & $16.23$ & $33.30$ & $6.52$ & $1.17$ & $3.02$ & $7.92$ & $2.37$ & $2.23$ & $47.94$ & $20.06$ & $69.84$ & $120.68$  \\ \midrule
				\multirow{4}{*}{MU$\rightarrow$HU} & FLOPs (G) & $0.44$ & $66.23$ & $3.09$ & $6.26$ & $1.32$ & $0.33$ & $1.32$ & $2.67$ & $0.33$ & $1.32$ & $8.62$ & $7.14$ & $2.81$ & $2.83$   \\
				& Params (M) & $0.24$ & $0.02$ & $0.19$ & $0.85$ & $0.44$ & $0.44$ & $0.44$  & $0.55$ & $0.44$ & $0.44$ & $0.34$  & $0.55$ & $0.65$ & $0.66$   \\
				& Training Time (s) & $340.47$ & $183.10$ & $526.25$ & $393.82$ & $3263.65$ & $1379.43$ & $3393.74$ & $8020.18$ & $1542.69$ & $2074.61$ & $8129.26$ & $24887.24$ & $605.17$ & $641.41$  \\
				& Test Time (s) & $22.86$ & $115.34$ & $63.23$ & $111.44$ & $0.91$ & $0.67$ & $0.92$ & $2.51$ & $1.43$ & $1.50$ & $1.27$ & $3.01$ & $4.36$ & $8.79$  \\ \midrule
				\multirow{4}{*}{TR$\rightarrow$MU} & FLOPs (G) & $0.44$ & $66.23$ & $3.09$ & $6.26$ & $1.32$ & $0.33$ & $1.32$ & $2.67$ & $0.33$ & $1.32$ & $8.62$ & $6.98$ & $2.81$ & $2.84$   \\
				& Params (M) & $0.24$ & $0.02$ & $0.19$ & $0.85$ & $0.44$ & $0.44$ & $0.44$ & $0.55$ & $0.44$ & $0.44$ & $0.34$ & $0.55$ & $0.65$ & $0.66$   \\
				& Training Time (s) & $138.24$ & $72.47$ & $206.19$ & $177.90$ & $6073.06$ & $3163.14$ & $7854.21$ & $14877.36$ & $835.30$ & $3799.24$ & $3392.07$ & $9296.51$ & $1207.13$ & $1277.57$ \\
				& Test Time (s) & $4.19$ & $21.22$ & $11.49$ & $24.16$ & $7.23$ & $3.02$ & $7.24$ & $18.01$ & $6.09$ & $5.21$ & $24.37$ & $58.51$ & $80.65$ & $144.57$  \\ \midrule
				\multirow{4}{*}{TR$\rightarrow$HU} & FLOPs (G) & $0.44$ & $66.23$ & $3.09$ & $6.26$ & $1.32$ & $0.33$ & $1.32$ & $2.67$ & $0.33$ & $1.32$ & $8.62$ & $6.98$ & $2.81$ & $2.84$   \\
				& Params (M) & $0.24$ & $0.02$ & $0.19$ & $0.85$ & $0.44$ & $0.44$ & $0.44$ & $0.55$ & $0.44$ & $0.44$ & $0.34$ & $0.55$ & $0.65$ & $0.66$  \\
				& Training Time (s) & $168.60$ & $79.06$ & $232.36$ & $172.51$ & $1093.65$ & $554.18$  & $1378.67$ & $4976.93$ & $649.74$ & $848.53$ & $3368.38$ & $9157.70$ & $1143.52$ & $1314.23$ \\
				& Test Time (s) & $25.06$ & $112.51$ & $64.08$ & $111.90$ & $0.90$ & $0.64$ & $0.87$ & $2.15$ & $1.41$ & $1.48$ & $1.27$ & $3.01$ & $5.62$ & $9.41$  \\ \midrule
				\multirow{4}{*}{HU$\rightarrow$MU} & FLOPs (G) & $0.44$ & $66.23$ & $3.09$ & $6.26$ & $1.32$ & $0.33$ & $1.32$ & $2.67$ & $0.33$ & $1.17$ & $8.62$ & $6.82$ & $2.81$ & $2.84$    \\
				& Params (M) & $0.24$ & $0.02$ & $0.19$ & $0.85$ & $0.44$ & $0.44$ & $0.44$ & $0.55$ & $0.44$ & $0.44$ & $0.34$ & $0.55$ & $0.65$ & $0.66$  \\
				& Training Time (s) & $23.27$ & $12.18$ & $35.48$ & $25.96$ & $801.25$ & $366.68$ & $839.26$ & $2416.54$ & $474.61$ & $145.02$ & $502.13$ & $1343.84$ & $301.27$ & $363.44$  \\
				& Test Time (s) & $4.48$ & $23.11$ & $13.10$ & $22.81$ & $6.21$ & $2.06$ & $6.26$ & $21.96$ & $5.13$ & $5.78$ & $24.67$ & $54.09$ & $76.92$ & $140.98$  \\ \midrule
				\multirow{4}{*}{HU$\rightarrow$TR} & FLOPs (G) & $0.44$ & $66.23$ & $1.94$ & $6.26$ & $1.32$ & $0.33$ & $1.32$ & $2.67$ & $0.33$ & $1.32$ & $8.62$ & $6.98$ & $2.81$ & $2.84$ \\
				& Params (M) & $0.24$ & $0.02$ & $0.37$ & $0.85$ & $0.44$ & $0.44$ & $0.44$ & $0.55$ & $0.44$ & $0.44$ & $0.34$  & $0.55$ & $0.65$ & $0.66$  \\
				& Training Time (s) & $24.57$ & $12.28$ & $9.47$ & $26.19$  & $808.52$ & $362.88$ & $846.25$ & $2503.76$ & $114.52$ & $265.25$ & $1105.90$ & $1366.92$ & $180.93$ & $221.12$  \\
				& Test Time (s) & $6.81$ & $32.59$ & $5.81$ & $31.71$ & $2.49$ & $1.05$ & $2.54$ & $9.88$ & $2.30$ & $2.20$ & $59.26$ & $24.21$ & $37.41$ & $63.32$  \\ \bottomrule
		\end{tabular}}
	\end{table*}

\end{document}